%% file: neurips_2026.tex
\documentclass{article}

\PassOptionsToPackage{numbers, compress}{natbib}
\usepackage[preprint]{neurips_2026}

\usepackage[utf8]{inputenc} % allow utf-8 input
\usepackage[T1]{fontenc}    % use 8-bit T1 fonts
\usepackage{hyperref}       % hyperlinks
\usepackage{url}            % simple URL typesetting
\usepackage{booktabs}       % professional-quality tables
\usepackage{amsfonts}       % blackboard math symbols
\usepackage{nicefrac}       % compact symbols for 1/2, etc.
\usepackage{microtype}      % microtypography
\usepackage{xcolor}         % colors
\usepackage{float}

\usepackage{microtype}
\usepackage{enumitem}
\usepackage{graphicx}
\usepackage{subfigure}
\usepackage{booktabs} % for professional tables
\usepackage{multirow}
\usepackage{pifont}
\newcommand{\xmark}{\ding{55}} % ✗
\usepackage{tcolorbox}
\usepackage{subcaption}
\usepackage{amsmath}
\usepackage{amssymb}
\newtcolorbox{infobox}[1]{
  colback=gray!5!white,
  colframe=gray!60!black,
  fonttitle=\bfseries,
  title={#1},
  breakable
}

\newtcolorbox{logicbox}[1]{
  colback=blue!5!white,
  colframe=blue!60!black,
  fonttitle=\bfseries,
  title={#1 \small{(Hidden Rules)}},
  breakable
}
\usepackage{hyperref}

\definecolor{mycellcolor2}{HTML}{e6f4f1}

\usepackage[export]{adjustbox}
\newcommand{\modellogo}[1]{%
    \raisebox{0.1em}{\includegraphics[height=0.9em, valign=c]{#1}\hspace{0.05em}}%
}
\newcommand{\modellogoo}[1]{%
    \raisebox{0.0em}{\includegraphics[height=0.9em, valign=c]{#1}\hspace{0.05em}}%
}
\usepackage{amsmath}
\usepackage{amssymb}
\usepackage{mathtools}
\usepackage{amsthm}

\usepackage[capitalize,noabbrev]{cleveref}

\theoremstyle{plain}

\theoremstyle{definition}

\theoremstyle{remark}

\usepackage{xspace}
\usepackage{listings}
\usepackage{xcolor}
\usepackage[table]{xcolor}
\usepackage[textsize=tiny]{todonotes}

\newcommand{\odyssey}{\textsc{OdysseyArena}\xspace}
\newcommand{\odysseylite}{\textsc{OdysseyArena-Lite}\xspace}
\newcommand{\odysseychallenge}{\textsc{OdysseyArena-Challenge}\xspace}

\definecolor{checkmarkgreen}{rgb}{0.0, 0.5, 0.0}
\definecolor{mycellcolor}{HTML}{FFFFFF}
\definecolor{darkgreen}{HTML}{1B5E20}
\definecolor{mycellcolor2}{HTML}{e6f4f1}
\definecolor{veronica-red}{RGB}{196,30,58}

\author{
  Fangzhi Xu\textsuperscript{1}\thanks{Equal contribution.}\\
  \And
  Hang Yan\textsuperscript{1}\footnotemark[1] \\
  \And
  Qiushi Sun\textsuperscript{2*\textdagger} \\
  \And
  Jinyang Wu\textsuperscript{3}\\
  \And
  Zixian Huang\textsuperscript{4}\\
  \And
  Muye Huang\textsuperscript{1}\\
  \And
  Jingyang Gong\textsuperscript{2}\\
  \And
  Zichen Ding\textsuperscript{4}\\
  \And
  Kanzhi Cheng\textsuperscript{5}\\
  \And 
  Yian Wang\textsuperscript{6}\\
  \And
  Xinyu Che\textsuperscript{1}\\
  \And
  Zeyi Sun\textsuperscript{4}\\
  \And
  Jian Zhang\textsuperscript{1}\\
  \And
  Zhangyue Yin\textsuperscript{7}\\
  \And
  Haoran Luo\textsuperscript{8}\\
  \And
  Ben Kao\textsuperscript{2}\\
  \And
  Qika Lin\textsuperscript{6}\thanks{Correspondence at qiushisun@connect.hku.hk and linqika@nus.edu.sg}\\
}
\title{\textsc{OdysseyArena}: Benchmarking \\ Large Language Models For Long-Horizon, Active and Inductive Interactions}

\begin{document}

\maketitle

\begingroup
\renewcommand\thefootnote{}
\footnotetext{\textsuperscript{1}Xi'an Jiaotong University, \textsuperscript{2}The University of Hong Kong, \textsuperscript{3}Tsinghua University, \textsuperscript{4}Shanghai AI Laboratory, \textsuperscript{5}Nanjing University, \textsuperscript{6}National University of Singapore, \textsuperscript{7}Fudan University,\textsuperscript{8}Nanyang Technological University}
\endgroup
\begin{abstract}
  The advancement of Large Language Models (LLMs) has catalyzed the development of autonomous agents capable of navigating complex environments.
However, existing evaluations primarily adopt a deductive paradigm, 
where agents execute tasks based on explicitly provided rules and static goals, often within limited planning horizons.
Crucially, 
this neglects the inductive necessity for agents to discover latent transition laws from experience autonomously,
which is the cornerstone for enabling agentic foresight and sustaining strategic coherence.
To bridge this gap, we introduce \odyssey, 
which re-centers agent evaluation on long-horizon, active, and inductive interactions.
We formalize and instantiate four primitives, 
translating abstract transition dynamics into concrete interactive environments.
Building upon this,
we establish \odysseylite for standardized benchmarking,
providing a set of 120 tasks to measure an agent’s inductive efficiency and long-horizon discovery.
Pushing further, we introduce \odysseychallenge to stress-test agent stability across extreme interaction horizons (\textit{e.g.}, $>$ 200 steps).
Extensive experiments on 15+ leading LLMs reveal that even frontier models exhibit a deficiency in inductive scenarios, identifying a critical bottleneck in the pursuit of autonomous discovery in complex environments.
\par\smallskip
{\small\noindent\textbf{Project homepage:} \url{https://os-copilot.github.io/OdysseyArena-Home/}\par}
\end{abstract}

\input{Sections/1Introduction}
\input{Sections/2Related}

\input{Sections/3Odyssey}

\input{Sections/4Odyssey-Bench}

\input{Sections/5Experiments}

\input{Sections/6Analysis}
\input{Sections/7Conclusion}

\bibliographystyle{plainnat}
\bibliography{neurips_2026}

%%%%%%%%%%%%%%%%%%%%%%%%%%%%%%%%%%%%%%%%%%%%%%%%%%%%%%%%%%%%

\appendix

\input{Sections/A_task_curation}

\input{Sections/B_more_analysis}

\input{Sections/C_experiment_settings}

\input{Sections/D_human_anno}

%%%%%%%%%%%%%%%%%%%%%%%%%%%%%%%%%%%%%%%%%%%%%%%%%%%%%%%%%%%%

\end{document}

%% file: Sections/1Introduction.tex
\section{Introduction}

% The rapid advancement of Large Language Models (LLMs) has catalyzed the development of autonomous agents capable of perceiving and acting within complex environments. These systems are increasingly integrated into specialized domains such as robotics, scientific discovery, and business automation. This expansion necessitates rigorous evaluation frameworks that reliably assess agent performance in dynamic and realistic settings. Although recent benchmarks have transitioned from static question answering toward sequential decision-making, they still fail to capture the fundamental capabilities required for truly autonomous operation. We identify three critical dimensions of agent intelligence that remain conspicuously absent from the current evaluation landscape.

% The emergence of Large Language Models (LLMs)
The emergence of Large Language Models (LLMs; \citep{comanici2025gemini,anthropic2024claude})
has sparked unprecedented interest in autonomous agents that can perceive environments, 
make decisions, and take actions to accomplish complex tasks.
These AI agents are increasingly deployed across diverse domains—from robotics~\citep{wang2024largelanguagemodelsrobotics} and game playing~\citep{Akata_2025} to scientific discovery~\citep{sun2025scienceboard} and business automation~\citep{xu2025theagentcompany}. 
As the capabilities of LLMs expand, so does the demand for evaluation benchmarks that can reliably assess agent performance in realistic, dynamic settings.
% \sqs{While recent benchmarks have evolved from static question answering toward interactive decision-making, 
% they predominantly assess a \textbf{deductive} mode of intelligence where agents operate under explicitly provided rules.}

% Recent agent benchmarks have moved beyond static question answering toward interactive environments that require sequential decision-making. 
% However, despite this shift, current dynamic evaluations still fail to capture several core capabilities that increasingly capable agents must exhibit in practice.

% This deductive paradigm overlooks a fundamental requirement of autonomous agency: 
% the capacity to actively explore and induce latent transition laws from raw interaction.
% , thereby limiting the evaluation of an agent's ability to navigate environments where rules are not pre-specified.
% We identify three such capability dimensions that remain largely missing from existing agent benchmarks.

\begin{figure}[t]
\begin{minipage}{0.48\textwidth}
specified success criteria, 
whereas the realistic deployment requires the agents to actively probe, react to feedback, and iteratively adjust their actions through self-improvement~\cite{yan2025tide}.
% Finally, \textbf{inductive reasoning from interaction} represents a critical evaluation gap in current benchmarks,
% as most protocols assess deductive compliance to provided instructions rather than the capacity to infer latent rules and transition dynamics from experience.
Finally, \textbf{inductive reasoning from interaction} represents a critical evaluation gap in current benchmarks, as most protocols assess deductive compliance to provided explicit instructions rather than the capacity to infer latent rules and transition dynamics from real-time interactive experience.
\end{minipage}
\hfill  % 自动填充空白，让左右分开
\begin{minipage}{0.48\textwidth}  

\vspace{-1.1em}
\centering
    \includegraphics[scale=0.42]{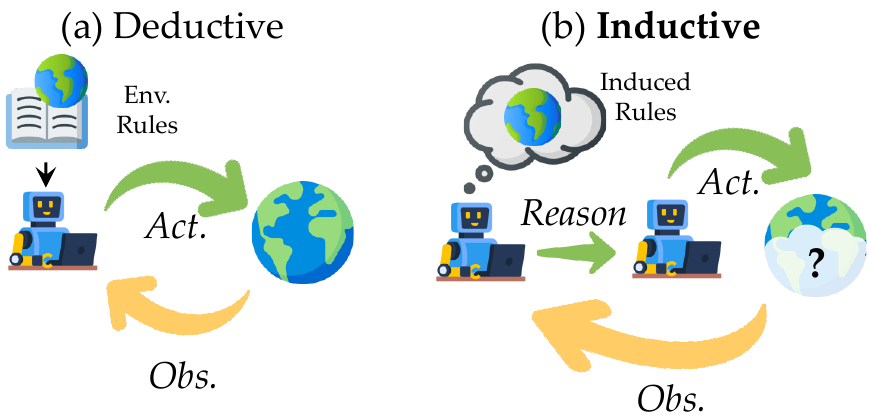}
        \vspace{-0.4em}
    \caption{Comparison between deductive and inductive settings in multi-turn agentic tasks.
    }
    \label{fig:intro}
    \vspace{-1em}
 
\end{minipage}
\end{figure}

While recent benchmarks have evolved from static question answering toward interactive decision-making~\citep{xie2024osworld, vodrahalli2024michelangelolongcontextevaluations, chung2025evaluatinglongcontextreasoningllmbased}, 
they predominantly assess a \textbf{deductive mode} of intelligence where agents rely on extensive prior knowledge to complete tasks.
% they predominantly assess a {deductive} mode of intelligence where agents operate under known environment priors and explicitly given rules.
% This deductive paradigm neglects the fundamental capacity to actively explore and induce latent transition laws from raw interaction,
% thereby limiting the evaluation of an agent's ability to navigate environments where rules are not pre-specified.
As illustrated in Figure~\ref{fig:intro},
such evaluation overlooks the essential \textbf{inductive mode}, in which agents are required to actively explore and induce the hidden rules underlying the environment.
% prevelant evaluation under deductive settings overlooks the essential capacity to actively explore and induce latent transition laws from raw interaction. 
This omission restricts the evaluation of an agent's proficiency in complex environments where rules are not pre-specified.

% \begin{figure}[t]
%     \centering
%     \includegraphics[scale=0.54]{Figures/inductive.pdf}
%         \vspace{-0.4em}
%     \caption{Comparison between deductive and inductive settings in multi-turn agentic tasks.
%     }
%     \label{fig:intro}
%     \vspace{-1em}
% \end{figure}

% qiushi ：太过理想化
We identify three critical capability dimensions that remain largely unaddressed in the current landscape. 
% First, \textbf{extremely long-horizon interaction} is often neglected by existing benchmarks that restrict episodes to fewer than 50 steps, which fails to capture the strategic coherence and error accumulation challenges inherent in sequences of thousands of steps. 
First, \textbf{extremely long-horizon interaction} is often neglected by existing widely used agent benchmarks that restrict episodes to fewer than 50 steps, which fails to capture the core strategic coherence and error accumulation challenges inherent in sequences of thousands of steps.
Second, \textbf{active exploration and trial-and-error} are frequently bypassed by environments that provide a fully 
% qiushi early exp?

% \textbf{(1) Extremely long-horizon interaction.}
% Most benchmarks restrict evaluation episodes to at most 50–100 steps, reflecting the limited planning horizons of early interactive agents. Yet real-world agents often operate over thousands of steps, where maintaining coherence and preventing error accumulation become central challenges. Whether LLM-based agents can sustain effective reasoning over such extreme horizons remains largely untested.

% \textbf{(2) Active exploration and trial-and-error.}
% Existing benchmarks typically assume that task goals and success criteria are fully specified in advance. 
% Under this setting, agents mainly demonstrate their ability to execute known strategies. In contrast, real environments frequently require agents to explore, fail, and adapt—actively probing the environment to discover how it works. This exploratory, trial-and-error behavior is largely absent from current evaluations.

% \textbf{(3) Inductive reasoning from interaction.}
% Most agent evaluations follow a deductive paradigm, where agents are given explicit rules or task descriptions and evaluated on compliance. However, many realistic environments are governed by implicit or undocumented dynamics. In such cases, agents must infer rules from experience rather than instructions—an inductive capability that existing benchmarks rarely assess.

In order to bridge this gap, 
we introduce \odyssey,
a suite of interactive environments that re-centers evaluation on long-horizon, active, and inductive reasoning, which entails inferring latent transition laws from empirical interactions. 
We formalize
% We abstract 
environments as generative state transition functions:
$(s_{t+1}, r_t) = \mathcal{T}(s_t, a_t),$
% \begin{equation}
% (s_{t+1}, r_t) = \mathcal{T}(s_t, a_t),
% \end{equation}
where $s_t \in \mathcal{S}$ is the latent state and $a_t \in \mathcal{A}$ is the agent action. 
The transition function $\mathcal{T}$ implicitly encodes the environment’s rules and regularities, which agents must actively induce from interaction in order to anticipate outcomes, plan effectively, and optimize behavior over long horizons. 
% This makes $\mathcal{T}$ the central object of agent evaluation.
To systematically study an agent’s inductive capacity,
we decompose $\mathcal{T}$ into a taxonomy of four representative structural primitives: discrete latent rules, continuous stochastic dynamics, multi-objective periodic patterns, 
and relational dependencies.

To facilitate empirical investigation into these abstract dynamics, we materialize these primitives into four diverse \odyssey environments. These environments are curated to be computationally efficient and lightweight while remaining functionally representative of real-world systems, providing a tractable and scalable testbed for the community.
% To concretely instantiate these four primitives, 
% we \textbf{materialize} these abstract primitives into four diverse \odyssey environments.
% They are curated to be computationally efficient and lightweight while remaining functionally representative of real-world systems, 
% providing a tractable testbed for the community.
% each highlighting a distinct form of structural complexity and requiring agents to induce hidden dynamics through interaction: 
Specifically,
{\textit{Turn On Lights} grounds discrete Boolean logic in interdependent bulb configurations, 
while \textit{AI Trading} presents continuous stochastic dynamics through multivariate stock-factor relationships. 
\textit{Energy Dispatch} requires agents to uncover periodic efficiency patterns under multi-objective constraints, 
and \textit{Repo System} engages agents in deducing the topological dependencies among software package versions in a virtual environment.}

% For practical evaluation,
% we further provide \odysseylite, a recommended\sqs{``lightweight'', reproducible, portable, reliable ... } benchmark configuration consisting of 30 tasks per environment (120 tasks in total across four environments), enabling fast and standardized assessment of agent performance. 
For practical evaluation,
we provide \odysseylite as a standardized benchmark of 120 curated tasks. 
This suite serves as a representative of a scalable task distribution, 
optimized for high evaluation throughput while preserving the core challenges of active and inductive discovery.
% Each task is designed with moderate interaction lengths to allow efficient evaluation while still requiring long-horizon reasoning, active exploration, and inductive inference. 
Each task maintains interaction horizons that are computationally tractable yet sufficiently non-trivial to necessitate long-horizon planning and the active induction of latent rules.
In addition, we release \odysseychallenge, a stress-test suite with 1,000+ steps per task and 10 tasks per environment, intended to probe the limits of agent persistence, reasoning stability, and the ability to maintain coherent strategies over extremely long horizons. Together, these two settings balance accessibility, efficiency, and scalability, supporting both rapid iteration on current agents and rigorous evaluation of next-generation capabilities.

% and wide range of model sizes. 

We evaluated over 15 trending LLMs,
spanning proprietary models and open-source models across different scales.
Overall, 
commercial models consistently outperform open-source alternatives, with Gemini 3 Pro Preview achieving the highest success rate of 44.17.
% and \textsc{GPT-5} following with 35.3. 
% \yh{Gemini-3-pro achieves 44.72 and GPT-5 achieves 35.28. Note that this data is only for a single test, not avg@4 (other data will be generated soon).}
% The results indicate that commercial models consistently maintain a performance lead, with \textsc{Gemini-3-Pro} and \textsc{GPT-5} achieving scores of 44.72 and 35.28 respectively.
Despite this, even the strongest commercial models broadly remain far below human-level performance across four environments in \odysseylite,
highlighting substantial gaps in long-horizon reasoning, active exploration, and inductive generalization.
Beyond these aggregate results, we conducted a detailed, fine-grained analysis of agent behavior across different environments and tasks, 
revealing patterns and failure modes that provide actionable insights for designing more capable and robust autonomous agents.

% In summary, this work makes the following contributions: 
Our primary contributions are as follows:

% \noindent (1) \textbf{Novel Perspectives For LLM Evaluation:} We introduce \textbf{\odyssey}, a benchmark suite covering long-horizon interaction, active exploration, and inductive reasoning. It features lightweight deployment and scalable difficulty control.
% \sqs{We introduce a novel evaluation paradigm that re-centers agentic intelligence on the capacity to autonomously induce latent world dynamics through long-horizon and active interaction.}

% \noindent (1) \textbf{Novel Perspectives For Agent Evaluation:} 
% We introduce \odyssey, 
% which initiates a novel evaluation paradigm that re-centers agentic intelligence on the capacity of long-horizon, active, and inductive reasoning.

\noindent (1) \textbf{Novel Perspectives For Agent Evaluation:} 
We propose a novel evaluation paradigm centered on the capacity for autonomous discovery. 
This shift refocuses agentic intelligence on the induction of latent world dynamics through long-horizon and active interaction.

% to autonomously induce latent world dynamics through long-horizon and active interaction.

% long-horizon, active, and inductive reasoning

% \noindent (2) \textbf{Diverse Environments:} Based on distinct world transition structures, we release four environments with two benchmark settings: \odysseylite (30 tasks per env) for standard evaluation, and \odysseychallenge (10 tasks, $>1,000$ steps) for stress-testing.
% qiushi：我建议这个部分，把环境along benchmark一起介绍出来
% xufangzhi：agree

% \noindent (2) \textbf{Diverse Environments:}
% We instantiate the proposed inductive paradigm into \odyssey environments, from which we establish \odysseylite as a standardized suite for efficient evaluation. This testbed is designed for high scalability and can be readily extended to more demanding interaction scenarios.

\noindent (2) \textbf{Reliable and Scalable Evaluation :}
% Reliable and Scalable Evaluation:
We instantiate the inductive paradigm evaluation into \odyssey, establishing \odysseylite as a standard suite for efficient evaluation.
% Our environments further provide a foundation that scales to probe the limits of agentic intelligence.

 % \sqs{@fangzhi, please check my comments}

\noindent (3) \textbf{Extensive Evaluations and Insights:}
Through an extensive evaluation of 15+ top-tier LLMs, we characterize the inductive bottleneck as a fundamental barrier to autonomous discovery while establishing rigorous benchmarking results for future research.

\vspace{-0.1in}

%% file: Sections/2Related.tex
\section{Related work}

\begin{table}[h]  % 这里把 figure 改成 table！
\begin{minipage}{0.48\textwidth}
% \vspace{-1em}
\vspace{-0.5em}

\paragraph{Interactive Benchmarks.} Interactive benchmarks for LLM agents have evolved from grounded language understanding in simplified grid-worlds~\citep{shridharalfworld, chevalierbabyai} 
to sophisticated digital~\citep{deng2023mind2web, zhouwebarena, xie2024osworld} and 
real-world systems~\citep{xu2025theagentcompany, yao2025taubench}.
Despite this progress, 
a critical bottleneck remains in temporal depth: most environments favor short horizons or trajectories \citep{rawlesandroidworld,sun2025genesis}), 
which fails to capture the error accumulation phenomenon and the decay of long-term planning consistency~\citep{liu2024agentbench}.
Furthermore, 
many existing protocols~\citep{wang2025odysseybench, patilberkeley} bypass exploratory requirements by providing gold instructions or detailed API docs.
\odyssey and derived benchmarks bridge these gaps by introducing long-horizon tasks that demand coherent internal states and robust
recovery strategies over extended interaction sequences.

\end{minipage}
\hfill
\begin{minipage}{0.48\textwidth}  
% \vspace{-1.7em}
\centering
\vspace{-2em}

\caption{Comparison of representative multi-turn agentic benchmarks. \emph{Ind.} indicates whether inductive reasoning is required. \emph{Horizon} denotes the number of steps required to complete a task, categorized as short ($<$50), long (50--100), or X-Long ($>$100). \emph{Deploy} describes the evaluation setup required to run the environment, with API-based deployment being the most lightweight.}
\label{tab:agent_benchmark_comparison}
\medskip
% 表格自动缩放到 0.48 宽度
\resizebox{\linewidth}{!}{%
\begin{tabular}{lccc}
\toprule
Benchmark 
& Ind.
& Horizon
& Deploy
\\
\midrule

BabyAI~\citeyearpar{chevalierbabyai}
& \color{checkmarkgreen}\checkmark
& Long
& Simulator
\\

ALFWorld~\citeyearpar{shridharalfworld}
& \color{checkmarkgreen}\checkmark
& Short
& Simulator
\\

GAIA~\citeyearpar{mialon2023gaia}
&\color{red}\xmark
& Long
& Offline
\\

WebArena~\citeyearpar{zhouwebarena}
& \color{red}\xmark
& Short
& Docker
\\

OSWorld~\citeyearpar{xie2024osworld}
& \color{red}\xmark
& Long
& Docker
\\

AndroidWorld~\citeyearpar{rawlesandroidworld}
& \color{red}\xmark
& Short
& Emulator
\\

BrowseComp~\citeyearpar{wei2025browsecomp}
& \color{red}\xmark
& Long 
& API
\\

\midrule
\textbf{\odyssey}
& \textbf{\color{checkmarkgreen}\checkmark}
& X-Long
& API
\\

\bottomrule
\end{tabular}
}
% \medskip

\end{minipage}
\end{table}  % 这里也要改成 table！

\paragraph{Inductive Reasoning.}
% Current agentic frameworks predominantly rely on deductive reasoning, where models apply pre-trained knowledge or provided rules to solve tasks~\citep{yao2022react, shinn2023reflexion}. 
% However, autonomous intelligence necessitates inductive reasoningthe capacity to infer latent transition dynamics from raw observations~\citep{lake2017building}. While static benchmarks like ARC~\citep{chollet2019measure} and Zebra-Logic~\citep{linzebralogic} evaluate symbolic induction, they remain passive and non-interactive, missing the "active discovery" loop essential for real-world agents. Interactive attempts like Mars~\citep{tang2024mars} evaluate induction through counter-commonsense games, yet performance is often confounded by pre-trained knowledge conflicts.
Current agentic frameworks, such as ReAct~\citep{yao2022react} and Reflexion~\citep{shinn2023reflexion}, 
primarily rely on deductive reasoning or test-time interactions~\citep{sun2024pushing,xu2025ph,xu2025genius} to apply internal knowledge or provided rules. 
However, 
the nature of intelligence necessitates inductive reasoning to infer latent rules and transition dynamics from raw observations~\citep{lake2017building}. 
While static benchmarks like ARC~\citep{chollet2019measure} and Zebra-Logic~\citep{linzebralogic} evaluate rule synthesis, 
they remain passive and fail to capture the active discovery loop essential for autonomous agents~\citep{linzebralogic}. 
% Although interactive environments like Mars~\citep{tang2024mars} introduce trial-and-error, they often struggle to decouple pure induction from pre-trained knowledge priors
% In contrast, \odyssey requires agents to induce latent structures through trial-and-error, 
% aligning with the concept of world models~\citep{ha2018recurrent} while posing higher-order symbolic challenges for LLMs.
While interactive environments like Mars~\citep{tang2024mars} facilitate exploratory interaction, they struggle to decouple pure induction from pre-trained knowledge priors.
In contrast with typical agent benchmarks shown in Table~\ref{tab:agent_benchmark_comparison},
this work necessitates inducing latent world structures through extremely long-horizon interactions,
aligning with the concept of world models~\citep{ha2018recurrent} while posing higher-order symbolic challenges for future agentic intelligence.

%% file: Sections/3Odyssey.tex
\vspace{-0.1in}
\section{\odyssey}

\subsection{Preliminaries}

Interactive environments can be characterized as generative processes where the environment's response to an action $a_t \in \mathcal{A}$ at a latent state $s_t \in \mathcal{S}$ is governed by a transition function $\mathcal{T}$:
\begin{equation}
(s_{t+1}, r_t) = \mathcal{T}(s_t, a_t).
\end{equation}
In this framework, $\mathcal{T}$ encapsulates the \textbf{unobservable regularities} and constraints that dictate the system's evolution. Unlike standard reinforcement learning paradigms that often focus on policy optimization within known or fixed MDPs, 
\odyssey emphasizes \textbf{latent structure induction}. To achieve long-horizon planning and optimal decision-making, agents must autonomously discover the functional form of $\mathcal{T}$ through strategic interaction.

% To systematically characterize the landscape of environment dynamics, we organize these transition functions into four \textbf{orthogonal structural primitives}, each posing a unique challenge to an agent's inductive bias:

\begin{figure*}[t]
    \centering
    \vspace{-0.1in}
    \includegraphics[width=0.93\linewidth]{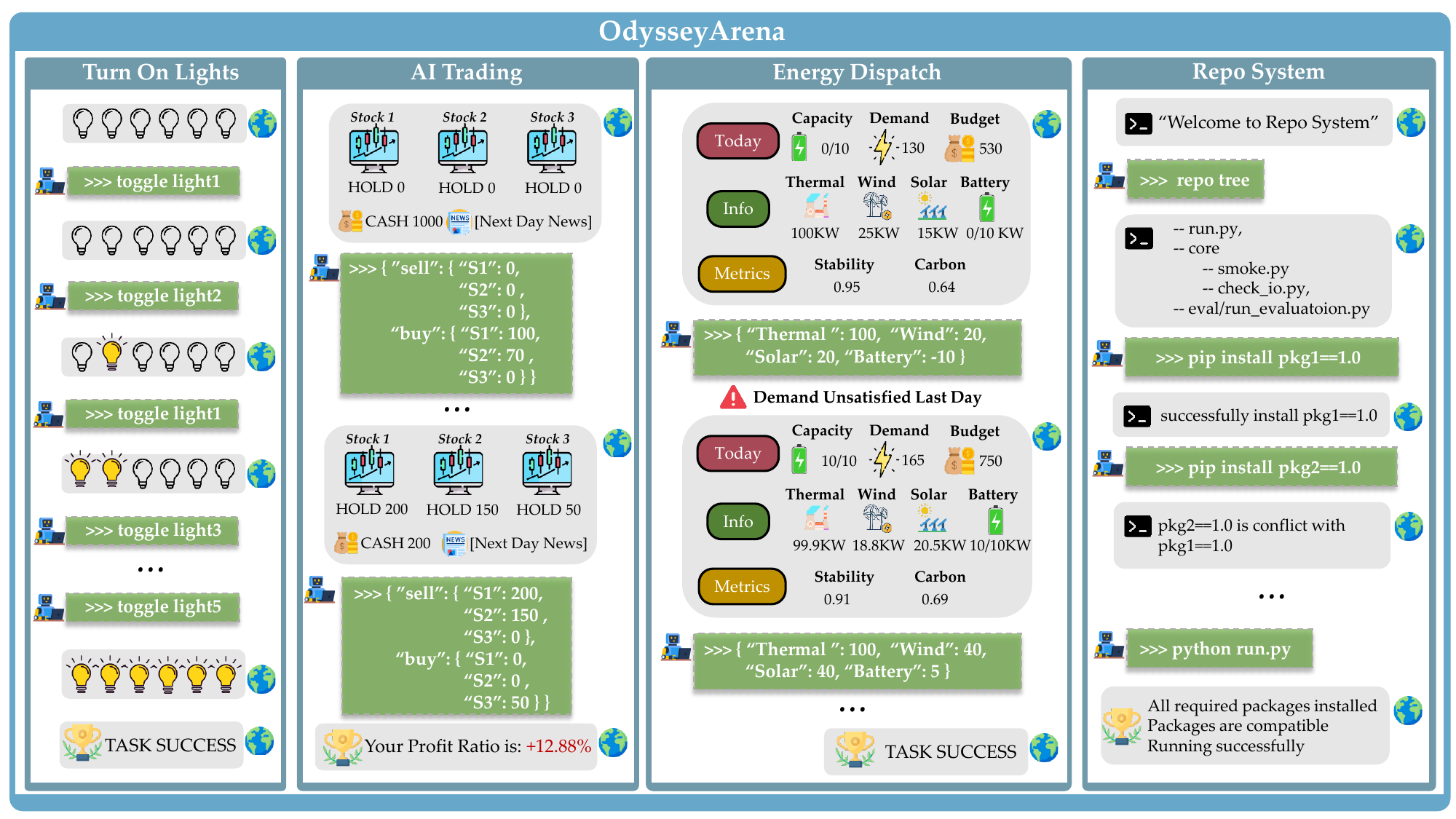}
    \caption{Demonstrations of four \odyssey environments: Turn On Lights, AI Trading, Energy Dispatch, and Repo System. For clarity, we omit the task prompts here and present only the interaction trajectories. Full prompts are provided in Appendix~\ref{appendix:exp_setting}.}
    \label{fig:illustration}
    \vspace{-0.1in}
\end{figure*}

To systematically characterize environment dynamics, we decompose the landscape of environment dynamics into a taxonomy of four \textbf{orthogonal structural primitives}, representing the fundamental mathematics that lead to complex real-world systems~\citep{clark1997dynamical,li2024state}.
By isolating these irreducible structures, we define a comprehensive set of world-modeling challenges that an autonomous agent must navigate:
\begin{itemize}[leftmargin=*,itemsep=0pt,topsep=2pt]
    % \item \textbf{Discrete Symbolic Rules:} Environments governed by logical predicates and causal dependencies, requiring symbolic hypothesis testing and intervention.
\item \textbf{Discrete Symbolic Rules:}  The transition is governed by Boolean logic over $N$ bits, where $s \in \{0, 1\}^N$. 
This requires the agent to perform symbolic hypothesis testing to uncover the latent causal dependencies and logical couplings.
    \item \textbf{Continuous Stochastic Dynamics:} 
    % The system evolves through a continuous state space $s \in \mathbb{R}^d$ subject to uncertainty $\epsilon$. Agents must apply statistical inference to disentangle underlying signals from exogenous noise.
    The system evolves through a continuous state space $s \in \mathbb{R}^d$ according to $s_{t+1} = f(s_t, a_t) + \epsilon$, where $\mathcal{T}$ incorporates a latent functional signal $f$ and  noise $\epsilon$. This necessitates statistical inference to disentangle underlying regularities from fluctuations.
    % Systems defined by continuous state spaces with aleatoric uncertainty, necessitating statistical inference and noise-robust modeling.
    \item \textbf{Periodic Temporal Patterns:} 
    % Processes characterized by long-range cyclic regularities, introducing complex temporal dependencies and multi-objective trade-offs.
    The transition function exhibits cyclic regularities defined by a period $P$, such that $\mathcal{T}(s, a, t) \approx \mathcal{T}(s, a, t+P)$. This necessitates identifying long-range temporal dependencies to optimize multi-objective trade-offs.
    \item \textbf{Relational Graph Structures:} 
    % Environments involving non-local interactions between entities, requiring combinatorial reasoning over graph-structured constraints.
    The environment is defined by a graph $G = (V, E)$, where transitions involve non-local interactions between entities. Success requires relational reasoning over the topological constraints that govern global state changes.
\end{itemize}
This taxonomy ensures a comprehensive evaluation, as each structure induces a distinct cognitive requirement—ranging from logical deduction to relational abstraction—that is irreducible to the others. 
% \odyssey integrates these primitives into a unified interface to benchmark agents' fundamental capacity for world-structure induction.
\textsc{OdysseyArena} integrates these primitives into a diverse suite of environments designed to assess an agent's fundamental capacity for world-structure induction, as illustrated in Figure~\ref{fig:illustration}.
We will introduce the respective environments in detail as follows.

% \begin{table*}[t]
% \small
% % \renewcommand{\arraystretch}{1.1}
% \centering
% \caption{Environment Statistics of \odyssey.}
% \label{tab:env_statistics}
% \begin{tabular}{lccccc}
% \toprule
% Environment & Action & Observation & Termination & Evaluation \\
% \midrule
% \textit{Turn On Lights} & Toggle Lights & Light States & Goal / Max Steps & Success Rate \\
% \textit{AI Trading} & Buy / Sell Positions & Prices \& Indicators & Fixed Horizon & Profit \\
% \textit{Energy Dispatch} & Set Generation Output & Demand \& Efficiency & Fixed Horizon & Success Rate \\
% \textit{Repo System} & Manage packages & Terminal Feedback & Goal / Max Steps & Success Rate \\
% \bottomrule
% \end{tabular}
% \end{table*}

\vspace{-0.1in}
\subsection{Env \uppercase\expandafter{\romannumeral 1}: Turn On Lights}

\paragraph{Overview.}
This environment instantiates the discrete symbolic rules primitive by simulating a network of $N$ interdependent lights.
The agent aims to reach a target configuration $s = \mathbf{1}$, 
representing the state where all lights are illuminated, 
through a sequence of toggling actions.
Dynamics are governed by latent Boolean couplings that remain fixed within one trial but vary across episodes,
where an intervention on a single light may trigger a deterministic cascade of state changes across the network.
Consequently, success requires active exploration to infer the underlying logic and the correct activation sequence.

% We simulate a network of $N$ interdependent lights with discrete symbolic rules.
% The agent must reach the target state $s=\mathbf{1}$ (all lights on) via toggling actions. Transitions are driven by latent Boolean couplings that are fixed per trial but vary across episodes: toggling one light can deterministically trigger cascades. Thus, success requires active exploration to infer the underlying logic and the correct activation sequence.

\vspace{-0.1in}
\paragraph{Hidden Rules.}
State transitions are governed by latent discrete rules that define how each action influences the configuration of lights. 
For each episode, these rules are instantiated by randomly combining Boolean operators to create a unique logical network, while ensuring that the resulting dependencies are solvable. 
As a result, an action targeting a particular light may deterministically affect multiple lights through indirect toggling or conditional activation. 
The dependencies remain fixed within an episode but vary across episodes, and are not directly observable to the agent. 
Consequently, the agent must infer the underlying logical structure through deliberate intervention and observation of state changes.

% Lights的hidden rules：每个灯泡是否可以控制（调亮/调灭）是根据依赖于其他灯泡的亮灭状态的，比如：
% B0：(B1 and B3) or (B2 and not B4)
% 表示在B1和B3同时亮，或者，B2亮且B4灭的时候，才可以控制B0这个灯泡。

% 底层在实现这个逻辑的时候，是采用了有向无环图生成。如果需要这部分的解释的话，明天pai可以根据python class翻译一下

\vspace{-0.1in}
\paragraph{Action and Observation Spaces.}
At each step, the agent takes a discrete action to activate/toggle a single light.
Although actions are defined over individual lights and remain fixed across episodes, their effects are not necessarily localized due to hidden logical couplings.
After each interaction, the agent observes the on/off status of all $N$ lights.
No information about the underlying logical dependencies or transition rules is revealed.
As a result, while the environment is fully observable with respect to the surface state, it is partially observable in terms of the latent transition dynamics induced by the hidden rule configuration.

% At each step, the agent takes a discrete action to activate/toggle one light. Actions are defined per light and fixed across episodes, but their effects may be non-local due to hidden logical couplings. After each step, the agent observes the on/off state of all $N$ lights; no transition rules or dependencies are revealed. Thus, the surface state is fully observable, while the latent transition dynamics remain hidden.

\vspace{-0.1in}
\paragraph{Task Evaluation.}
An episode is considered successful if all lights are turned on before the interaction budget is exhausted.
Performance is measured by the task success rate.
Solving the task efficiently requires the agent to identify and exploit the latent logical rules, rather than relying on myopic or brute-force exploration.

\vspace{-0.1in}
\subsection{Env \uppercase\expandafter{\romannumeral 2}: AI Trading}

\paragraph{Overview.}
This environment instantiates the continuous stochastic dynamics primitive, where the agent manages a multi-asset portfolio to maximize cumulative returns $\sum r_t$.
Market transitions are driven by latent factors obscured by stochastic fluctuations, 
with daily news serving as indirect hints for future price movements.
Unlike reactive tasks, success here depends on the agent's ability to induce the underlying market regularities, requiring it to disentangle the meaningful signal $f$ from noise to execute optimal multi-step, long-horizon trading strategies.

\vspace{-0.1in}
\paragraph{Hidden Rules.}
% Price transitions are governed by latent continuous-valued dynamics that relate asset returns to underlying market factors.
% Specifically, asset prices follow stochastic processes driven by unknown linear relationships with these factors, combined with exogenous noise.
% The parameters of these relationships remain fixed within an episode but vary across episodes, and are not directly observable.
% Consequently, the agent must infer the structure and strength of these dependencies from noisy observations and trading outcomes.

% Trading的底层规则是一个多元一次函数（矩阵），表达的是某一只股票Si的涨跌，依赖于多个因子的变化。可以参考prompt。

Price transitions are governed by the function $s_{t+1} = \mathbf{W} z_t + \epsilon$, where $s \in \mathbb{R}^d$ denotes asset returns. In this formulation, $\mathbf{W}$ represents a latent factor loading matrix that maps a vector of unobserved market factors $z_t$ to the price changes of $d$ assets. 
While $z_t$ and the noise $\epsilon$ fluctuate at each timestep, 
the structural relationships defined by $\mathbf{W}$ remain invariant within an episode but vary across tasks. 
Consequently, 
agents must treat the sequence of interactions as a noisy observation process to identify the functional form of $\mathbf{W}$ for effective long-horizon planning.

% \paragraph{Action Space.}
% At each timestep, the agent decides a set of buy and sell operations over multiple assets. 
% An action specifies the quantity of shares to buy or sell for each stock, subject to budget and inventory constraints.
% Sell orders are executed before buy orders within the same timestep, enabling portfolio reallocation in a single decision.
% This combinatorial and constrained action space requires the agent to jointly reason about asset selection, trade timing, and capital allocation.

% \paragraph{Observation Space.}
% The agent observes current and historical asset prices, along with a set of market indicators derived from latent factors.
% However, the true factor values and their relationships to asset returns are not revealed.
% This results in partial observability, where the agent must disentangle signal from noise to estimate the underlying dynamics.

\vspace{-0.1in}
\paragraph{Action and Observation Spaces}
At each step $t$, the agent perceives an observation $o_t \in \mathcal{O}$ comprising historical prices and news-derived indicators. Since $o_t$ serves as a noisy proxy for $z_t$, 
the agent must utilize the information within $o_t$ to estimate the underlying market state. Based on the perceived $o_t$, the agent issues a combinatorial action $a_t \in \mathcal{A}$ specifying buy and sell quantities across all assets. 
To enable portfolio reallocation within a single decision cycle, the environment enforces sequential execution, processing sell orders before buy orders, requiring the agent to jointly reason about asset selection, trade timing, and capital allocation under the transition logic $\mathcal{T}$.

% \sqs{At each step $t$, the agent perceives an observation $o_t \in \mathcal{O}$ comprising historical prices and news-derived indicators that serve as proxies for the latent factors $z_t$. Because the true relationships in $\mathbf{W}$ are not revealed, the agent must disentangle informative signals from noise to estimate the underlying market state. Based on these cues, the agent issues a combinatorial action $a_t \in \mathcal{A}$ specifying buy and sell quantities across all assets. To enable seamless portfolio reallocation within a single decision cycle, the environment processes sell orders prior to buy orders, subject to strict budget and inventory constraints. This constrained interface necessitates that the agent jointly reasons about asset selection, trade timing, and capital allocation under the transition logic of $\mathcal{T}$.} 

\vspace{-0.1in}
\paragraph{Task Evaluation.}
Performance is measured by cumulative return over the trading horizon, adjusted for transaction costs and risk constraints.
Successful performance requires learning and exploiting the latent stochastic dynamics over time, as opposed to reactive or single-step trading strategies.

\begin{figure*}[t]
    \centering
    \includegraphics[scale=0.66]{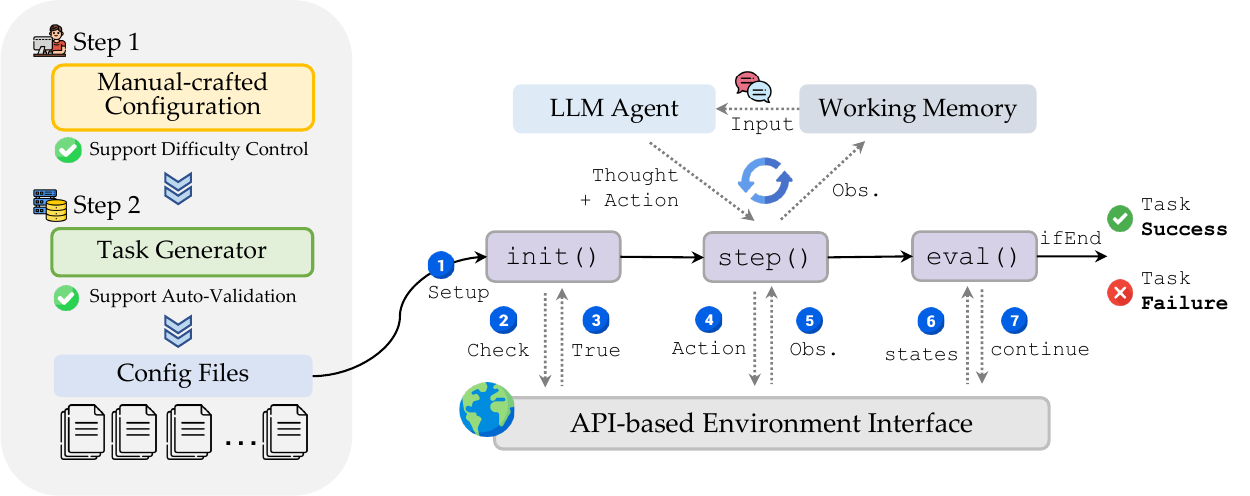}
    \caption{Overview of the benchmark architecture, illustrating the environment configuration initialization (left) and the interaction loop between the LLM agent and the environment step logic (right).
    }
    \label{fig:config}
    % \vspace{-1em}
    \vspace{-0.1in}
\end{figure*}

\subsection{Env \uppercase\expandafter{\romannumeral 3}: Energy Dispatch}

% \sqs{periodic temporal patterns}

\paragraph{Overview.}
This environment instantiates periodic temporal patterns by modeling a long-horizon energy dispatch problem in a dynamic power grid.
The agent acts as a dispatcher, allocating thermal, wind, solar, and battery resources to meet daily demand under budget constraints.
Unlike deductive optimization tasks, the environment enforces strict constraints: repeated budget or demand violations trigger immediate termination, simulating irreversible system failure.
Achieving a stable, low-carbon supply therefore requires anticipating latent efficiency cycles and planning over extended horizons to avoid systemic collapse.

\vspace{-0.12in}
\paragraph{Hidden Rules.}
% The actual power output of renewable sources is governed by hidden and time-varying efficiency factors.
% Specifically, wind and solar generation follow distinct periodic patterns, inducing fluctuations between rated and realized output.
% These temporal dynamics remain fixed within an episode but are not directly observable, requiring the agent to infer them from historical observations.
% As a result, effective control depends on learning long-term periodic structures and proactively allocating safety margins to hedge against uncertainty.

The core challenge lies in the discrepancy between the agent's planned dispatch and the actual power generation. 
The system evolves according to a transition logic where the realized output $P_{\text{real}}$ is determined by the agent's rated action $a_t$ modulated by a latent efficiency vector $E_t$, 
formally $P_{\text{real}} \approx a_t \odot E_t$. Here, $E_t$ represents the time-varying efficiency for each power source.
Crucially, the efficiency factors for wind and solar are governed by distinct, unobserved periodic functions $E_t \approx E_{t+T}$ with unique periods $T$. Since these fluctuations are not directly observable, the agent must infer the underlying periodic structures from historical output gaps.

% Energy的hidden rules是火电、风电、光电的每日效率值，是跟随周期变化的，且周期都互不相同。
% 比如某一天光电效率很高，所以可以通过多调配一点光电，满足预算和发电需求。但突然有一天光电波动很大，效率降低很多，这样的话，可能相同的钱就没法满足发电用量了。所以需要agent学会这种周期性的效率变化规律。

\vspace{-0.1in}
\paragraph{Action and Observation Space.}
% At each timestep, the agent specifies the rated output levels for thermal, wind, and solar generators, as well as the net charging or discharging command for a battery storage unit.
% All actions are subject to capacity limits and operational constraints, and jointly determine the next day's available supply.
% This continuous, multi-dimensional action space requires the agent to reason about resource allocation, inter-temporal trade-offs, and future system dynamics.

At each timestep,
the agent receives $o_t$ containing the current electricity demand $D_t$ and operating budget $B_t$. 
Based on these requirements, the agent issues a continuous action $a_t \in \mathbb{R}^4$ specifying the rated output for each generation type and the net charge/discharge command for the battery. 
The battery introduces inter-temporal dependencies, 
allowing the agent to buffer energy across time steps. Crucially, the environment enforces a strict safety protocol: repeated violations of demand satisfaction or budget limits trigger an early termination, simulating an irreversible grid collapse.

\vspace{-0.1in}
\paragraph{Task Evaluation.}
% Agent performance is evaluated using long-term metrics that capture grid stability and carbon intensity.
% The task is considered successful only if both stability and carbon metrics meet predefined thresholds over the full planning horizon, encouraging sustained, foresighted decision-making rather than myopic optimization.

Success is defined by a rigorous multi-objective criterion. First, the agent must ensure survival by maintaining system stability without triggering early termination over the full horizon $H$ (\textit{e.g.}, 120 days). Second, upon completion, the aggregate performance must satisfy predefined thresholds for both Carbon Intensity ($\mathcal{C} < \tau_c$) and Grid Stability ($\mathcal{S} > \tau_s$).

\subsection{Env \uppercase\expandafter{\romannumeral 4}: Repo System}

\paragraph{Overview.}
This environment models a realistic software repository management scenario, where the agent must configure a Python project to execute successfully.
The agent interacts with a partially observable dependency ecosystem, in which resolving local failures may introduce new global inconsistencies.
Success requires systematic diagnosis, relational reasoning over dependencies, and careful planning under non-monotonic side effects.

% 获胜条件：在规定步数内（比如我们设置的是120步），完成环境配置并成功运行脚本。

% \paragraph{Setup.}
% The environment simulates a hierarchical code repository containing multiple runnable scripts organized across subdirectories.
% The agent can inspect the repository structure, execute individual sub-scripts for debugging, and modify the software environment through package installation or removal.
% While sub-scripts can be used to localize errors, the episode is considered successful only when the full project entrypoint executes without failure.
% Episodes terminate either upon successful execution or when a maximum interaction budget is reached.

\vspace{-0.1in}
\paragraph{Hidden Rules.}
The transition logic $\mathcal{T}$ is defined by a latent dependency graph $G = (V, E)$. Here, nodes $V$ represent software packages with specific versions, and directed edges $E$ represent compatibility constraints (\textit{e.g.}, Pkg A v1.0 requires Pkg B $\ge$ v2.0). The system state $s_t \subseteq V$ represents the currently installed environment configuration. Crucially, the graph $G$ is hidden from the agent. A transition $s_{t+1} = \mathcal{T}(s_t, a_t)$ triggered by an installation command involves a rigorous resolution process: the system automatically installs required ancestors and uninstalls conflicting nodes to maintain local consistency, potentially altering parts of $s_t$ not targeted by $a_t$ (side effects).

\vspace{-0.1in}
\paragraph{Action and Observation Space.}
The agent perceives the environment through $o_t$, consisting of terminal outputs, file structures, and execution logs. Since $G$ is latent, $o_t$ serves as a sparse signal revealing only the graph ``broken edges'' (\textit{e.g.}, \texttt{ImportError}). 
Based on these error traces, the agent issues discrete symbolic actions $a_t \in \mathcal{A}_{\text{shell}}$ (\textit{e.g.}, \texttt{pip install}).
Since actions act as high-level graph mutation operators, the agent must sequence them to navigate the combinatorial state space, deducing the topology of $G$ to resolve conflicts without explicitly observing the full picture.

\vspace{-0.1in}
\paragraph{Task Evaluation.}
The task is evaluated by whether the agent achieves a globally consistent environment in which the full project executes successfully.
Partial or intermediate fixes (\textit{e.g.}, running individual sub-scripts) receive no reward if the global entry point fails, as local improvements may conflict with overall correctness.
This strictly binary protocol enforces global consistency, emphasizing long-horizon planning, relational reasoning over interdependent components, and robustness to delayed and indirect effects over myopic error patching.

%% file: Sections/4Odyssey-Bench.tex
% To bridge the gap between abstract primitives and concrete evaluation, we formalize each task as a deterministic instance of an \odyssey defined by its (1) structural configuration and (2) temporal trajectory. This ensures that while the underlying transition logic remains complex, every evaluation episode is fully reproducible and rigorously controlled.
% \sqs{@fangzhi, please provide me with the essence of curating benchmark cases.}

\vspace{-0.2cm}
\section{\odysseylite and \odysseychallenge}

% To enable reproducible online evaluation, 
% we first curate and introduce \odysseylite as a standardized task configuration for all environments. 
% Additionally, 
% we provide \odysseychallenge for stress-testing.

Building upon the infinite task space provided by \odyssey,
we derive two distinct benchmarking protocols to serve as standardized instantiations designed to evaluate agent performance across different difficulties:
% \odysseylite and \odysseychallenge....
% (1) \odysseylite is tailored for efficient and reproducible performance assessment, 
% providing a suite of tasks for rapid evaluation and analysis.
We primarily introduce \odysseylite as a suite tailored for efficient and reproducible performance assessment.
% Moreover, 
% (2) \odysseychallenge is conceived for rigorous stress-testing, 
% focusing on agent stability and inductive resilience over extreme interaction horizons.
Additionally, we provide \odysseychallenge for stress-testing agent stability and inductive resilience over extreme interaction horizons.
The benchmark construction process and sample details are described below.
% \sqs{trim challenge part}

\vspace{-0.2cm}
\subsection{Task configuration and curation}

% For each environment, we manually specify valid ranges for key environment parameters.
% Please refer to Appendix~\ref{appendix:task_curation} for details.
% Tasks are then randomly sampled within these predefined ranges, ensuring sufficient diversity while maintaining a bounded and controllable difficulty level.
% To further guarantee reproducibility and eliminate uncontrolled stochasticity, all sources of random noise in the environments are instantiated with fixed parameters.
% As a result, each task in the benchmark corresponds to a deterministic environment instance that can be consistently evaluated across different agents and runs.

To derive verifiable benchmarks from \odyssey,
we formalize each task as a deterministic instance sampled from a bounded parameter distribution (see Appendix~\ref{appendix:task_curation} for details).
This transition from abstract primitives to concrete evaluation is achieved by decomposing each instance into its structural configuration and temporal trajectory.

\vspace{-0.1in}
\paragraph{Structural Configuration.}
The configuration initializes latent rules $\mathcal{T}$ that remain invariant throughout an episode.
In \textit{Turn On Lights}, this corresponds to stochastically generating a DAG with specified density and logical operator ratios; in \textit{AI Trading} and \textit{Repo System}, it instantiates linear coefficient matrices and package dependency graphs, respectively.
To ensure solvability, constraint satisfaction algorithms are applied during curation to guarantee a valid solution path for every rule set.

\vspace{-0.1in}
\paragraph{Temporal Trajectory.}
To eliminate stochasticity, 
we pre-determine all time-varying factors in \odyssey as fixed sequences within the task metadata. For environments with dynamic states such as \textit{AI Trading} and \textit{Energy Dispatch}, 
components including daily factor fluctuations, efficiency curves, and resource budgets are pre-calculated. This ensures that the environment's evolution is not subject to runtime randomness, allowing for an identical and fair comparison across different agents and experimental trials.

% \paragraph{Task Sampling strategy / difficulty / diversity? \sqs{need discuss, delete diversity}.}

% We define valid ranges for key environment parameters i.e., the number of entities or the complexity of dependencies, and sample tasks within these boundaries. This approach  a controllable difficulty level while providing sufficient diversity to test the limits of inductive generalization. 
% Please refer to Appendix~\ref{appendix:task_curation} \sqs{paipai?} for a comprehensive list of the parameter distributions used in our sample curation pipeline.
% \sqs{need to discuss with Fangzhi, trying to connect this part with symbols}

% Task Sampling Strategy. 
% \vspace{-0.1in}

\vspace{-0.1in}
\paragraph{Task Sampling Strategy.}
We sample tasks by defining valid ranges for key environment parameters, such as the number of entities or the depth of logical dependencies. 
This sampling process is specifically calibrated to modulate task difficulty across two protocols. \odysseylite occupies a tractable parameter set for efficient evaluation, 
while \odysseychallenge targets the empirical limits of these ranges to stress-test agent stability over extreme horizons. 
This approach enables a characterization of the inductive bottleneck across varying environmental challenges. 
Detailed human anno processes are in Appendix~\ref{app:human_annotation}.

\subsection{Task statistics}
Under the \odysseylite setting, we construct a fixed evaluation suite by sampling 30 tasks for each environment.
The maximum allowed number of interaction steps is environment-specific, reflecting the inherent complexity of the underlying dynamics.
Specifically, the step limits are set to 200 for \textit{TurnOnLights}, and 120 for \textit{AI Trading}, \textit{Energy Dispatch}, and \textit{Repo Management}, respectively.
This configuration yields a balanced benchmark that supports reliable comparison across environments while remaining computationally efficient for online evaluation.

% While \odysseylite serves as our primary evaluation setting and the basis for all main experiments in this paper, 
% we additionally introduce \odysseychallenge as a more demanding variant designed for stress-testing advanced agents.
% In contrast to \odysseylite, \odysseychallenge substantially extends the required reasoning horizon, with tasks often exceeding 1{,}000 interaction steps.
% This setting targets failure modes that do not manifest under standard budgets, such as long-term credit assignment breakdown and compounding planning errors.
% Due to its significantly higher computational cost, \odysseychallenge is not used in our main evaluations, but is provided as an optional benchmark for future research.

While \odysseylite serves as our primary evaluation setting and the basis for all main experiments in this paper, 
we additionally introduce \odysseychallenge as a more demanding variant designed for stress-testing advanced agents, which substantially extends the required reasoning horizon, with tasks often exceeding 1{,}000 interaction steps.
This setting targets failure modes that do not manifest under standard budgets, such as long-term credit assignment breakdown and compounding planning errors.
Due to its significantly higher computational cost, \odysseychallenge is not used in our main evaluations, but is provided as an optional benchmark for future research.

% The details are presented in Table~\ref{tab:benchmark}.

% \begin{table}[t]
% \small
% % \renewcommand{\arraystretch}{1.1}
% \centering
% \caption{Statistics of \odysseylite and \odysseychallenge. The latter contains 15 samples for each environment.\xfz{merge this table in text}}
% \label{tab:benchmark}
% \begin{tabular}{lcc}
% \toprule
%  &\#Samples &Horizons \\
% \midrule
% \textbf{\odysseylite} &120 &120-200 \\
% \quad Turn On Lights &30 &200 \\
% \quad AI Trading &30 &120 \\
% \quad Energy Dispatch &30 &120 \\
% \quad Repo System &30 &120 \\
% \midrule
% \textbf{\odysseychallenge} &60 &$>$1,000 \\
% % \quad Turn On Lights &15 &$>$1,000   \\
% % \quad AI Trading &15 &$>$1,000   \\
% % \quad Energy Dispatch &15 &$>$1,000 \\
% % \quad Repo Management &15 &$>$1,000 \\
% \bottomrule
% \end{tabular}
% \end{table}

%% file: Sections/5Experiments.tex
\vspace{-0.2cm}
\section{Experiments}

\subsection{Experimental settings}
\label{sec:exp_settings}

\begin{table*}[t]
\centering
\caption{Performance comparison on four environments. We provide three different reasoning effort of gpt-oss-120b. For \textit{AI Trading} environment, we report the profit rate and \textbf{Best@4} is calculated based on the highest profit of each task. For other three environments, we report the success rate. \textcolor[HTML]{289BA2}{Colored Rows} represent proprietary models. The best results are in \textbf{bold}.}
\label{tab:main_results}
\resizebox{\textwidth}{!}{
\begin{tabular}{lcccccccccc}
\toprule
\multirow{2}{*}{Model} 
& \multicolumn{2}{c}{Turn On Lights} 
& \multicolumn{2}{c}{AI Trading} 
& \multicolumn{2}{c}{Energy Dispatch} 
& \multicolumn{2}{c}{Repo System}  \\
\cmidrule(lr){2-3} \cmidrule(lr){4-5} \cmidrule(lr){6-7} \cmidrule(lr){8-9}
& Avg@4 & Pass@4 
& Avg@4 & Best@4 
& Avg@4 & Pass@4 
& Avg@4 & Pass@4  \\
\midrule

% \multicolumn{9}{c}{\textbf{Human Evaluations}} \\
% \midrule
Human & 81.67 & 100.00 & +92.55\% & +197.23\% & 25.00 & 60.00 & 77.50 & 100.00  \\
\midrule

\rowcolor{mycellcolor2}
\modellogo{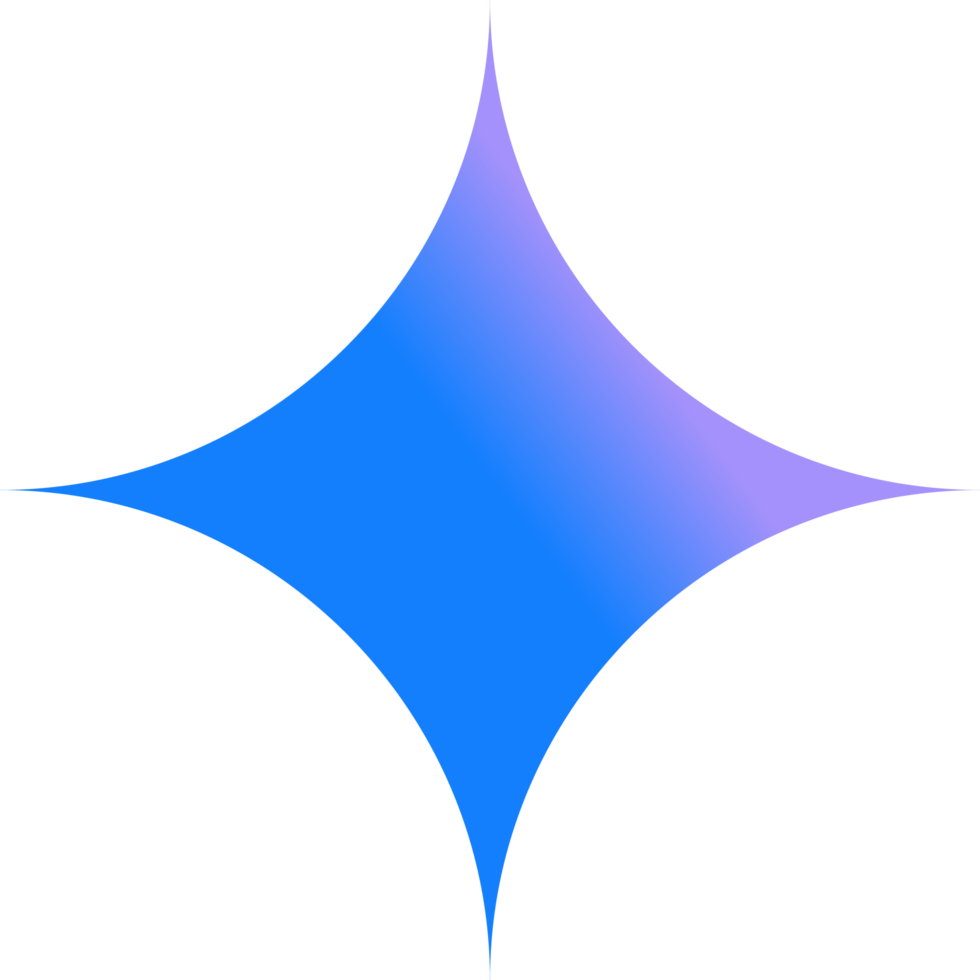}
Gemini 3 Pro Preview   
& \textbf{44.17} & \textbf{76.67} & \textbf{+67.71\%} & \textbf{+76.94\%} & \textbf{30.00} & 36.67 & \textbf{65.83} & 80.00  \\

\rowcolor{mycellcolor2}
\modellogo{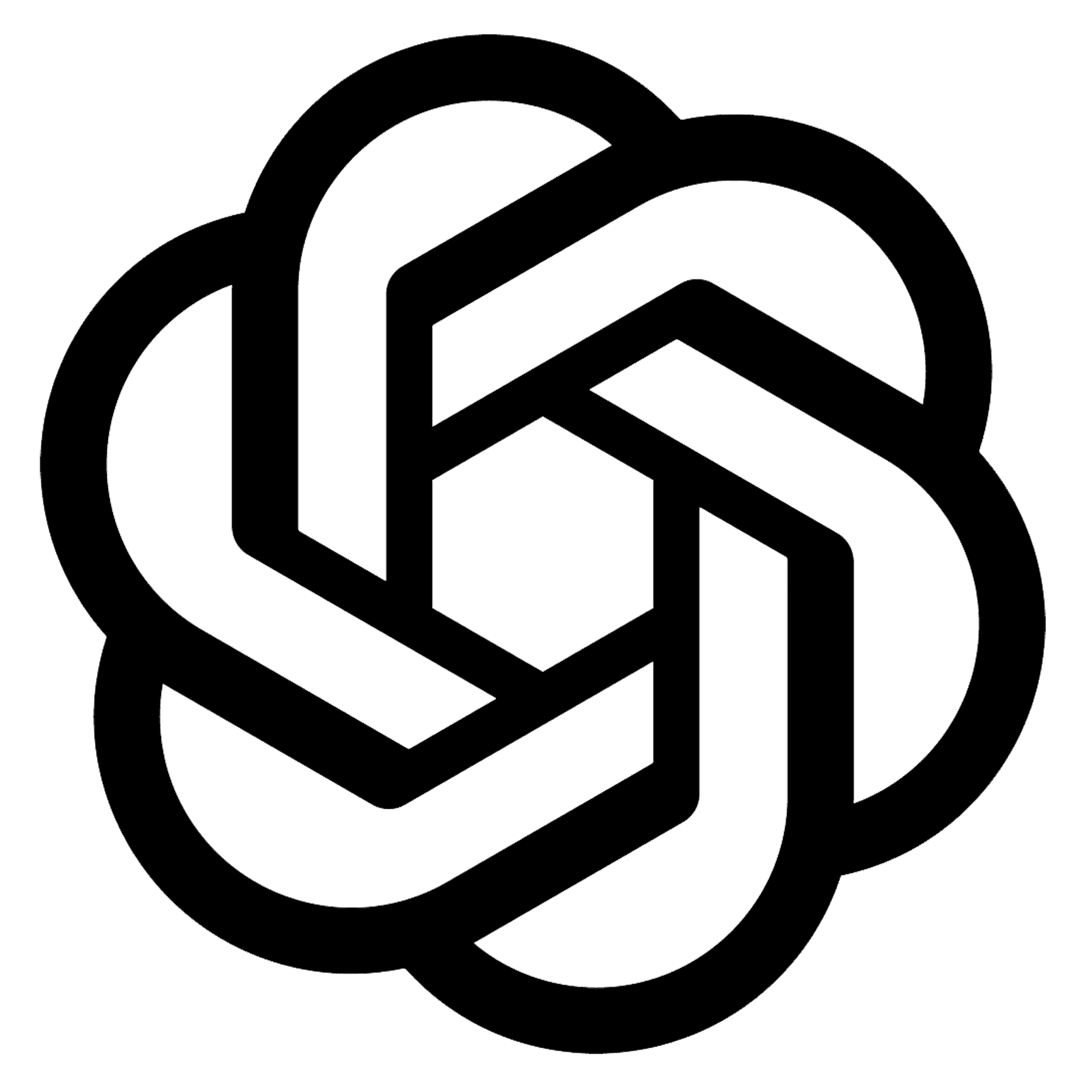}
GPT-5    
& 28.33 & 40.00 & +17.32\% & +20.47\% & 23.33 & \textbf{40.00} & 62.50 & \textbf{83.33} \\

\rowcolor{mycellcolor2}
\modellogo{Figures/gemini.png}
Gemini 2.5 Pro    
& 29.17 & 50.00 & +33.02\% & +40.12\% & 10.83 & 26.67 & 50.00 & 66.67    \\

\modellogoo{Figures/gpt.png}
gpt-oss-120b (high) 
& 27.50 & 40.00
& +23.27\% & +27.47\%
& 0.00 &  0.00
& 18.33 &  33.33 \\

\modellogo{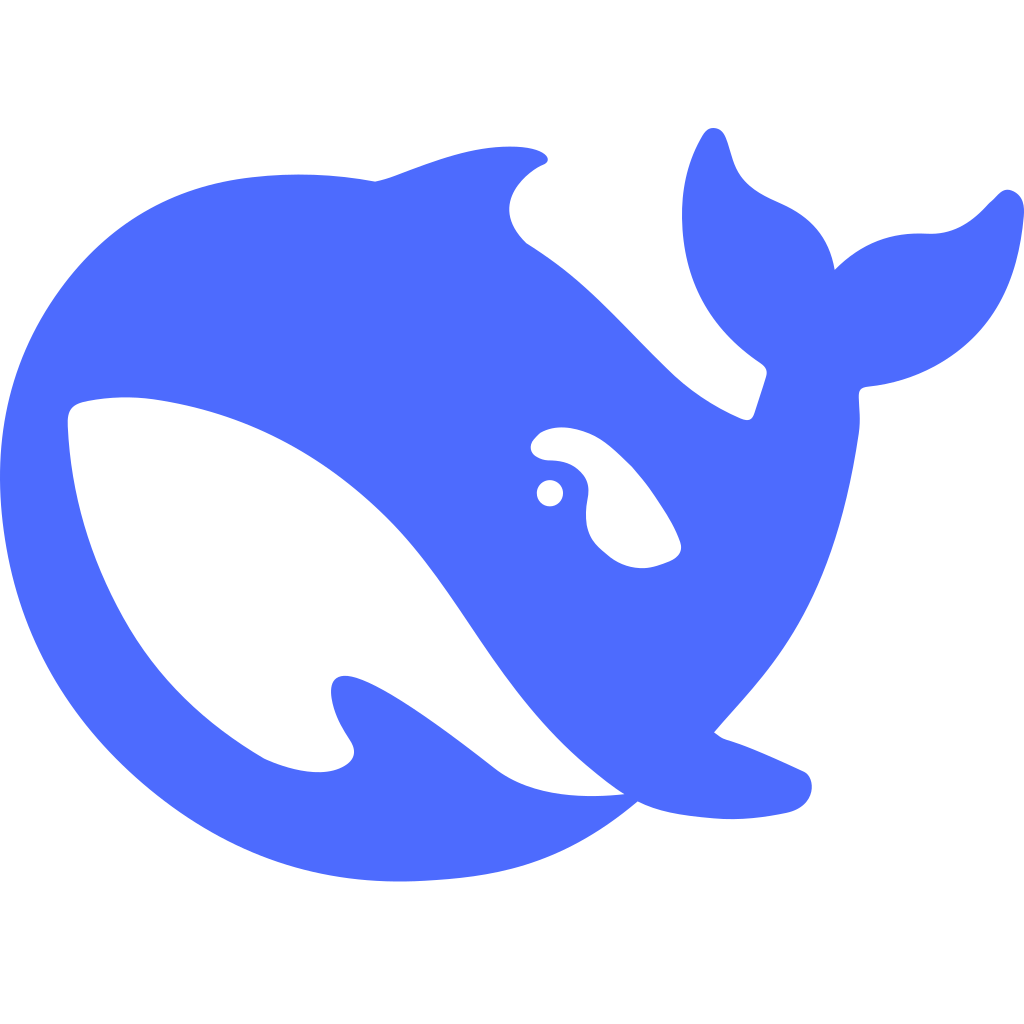}
DeepSeek-V3.2 
& 18.33 & 36.67 & +8.62\% & +12.88\% & 0.00 & 0.00 & 48.33 & 76.67  \\

\rowcolor{mycellcolor2}
\modellogo{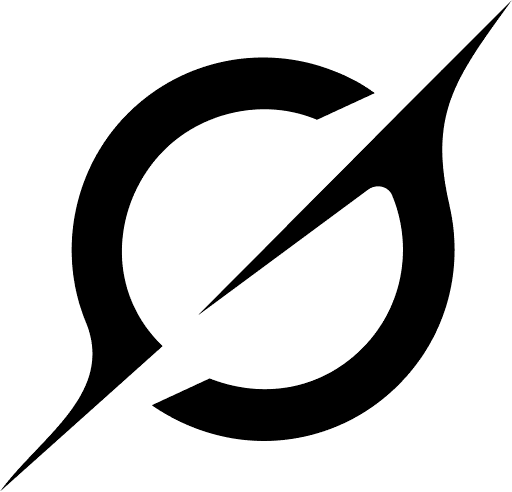}
Grok 4 Fast
& 14.17 & 40.00 & +5.70\% & +11.52\% & 0.00 & 0.00 & 38.33 & 60.00 \\

\modellogo{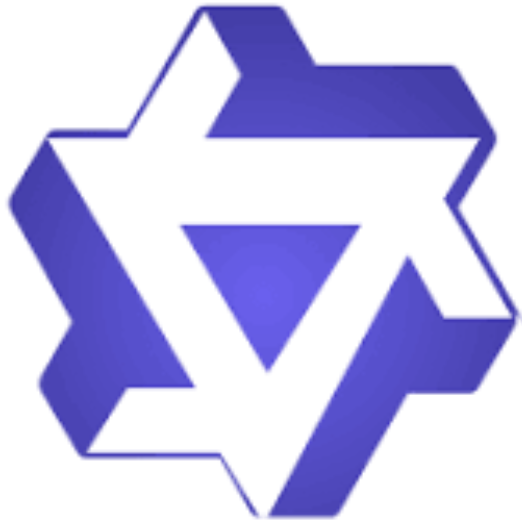}
Qwen3-235B-A22B-Instruct 
& 15.00 & 43.33 & +11.26\% & +17.67\% & 0.00 & 0.00 & 15.83 & 36.67  \\

\modellogo{Figures/qwen.png}
Qwen3-30B-A3B-Instruct 
& 11.67 & 26.67 & +4.76\% & +8.94\% & 0.00 & 0.00 & 26.67 & 50.00  \\

\modellogoo{Figures/gpt.png}
gpt-oss-120b (medium)
& 16.67 & 40.00 & +3.21\% & +7.09\% & 0.00 & 0.00 & 2.50 & 6.67  \\

\modellogo{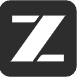}
GLM-4-32B-0414  
& 14.17 & 33.33 & +3.14\% & +7.24\% & 0.00 & 0.00 & 9.17 & 30.00  \\

\modellogoo{Figures/gpt.png}
gpt-oss-120b (low)
& 7.50 & 13.33 & +2.02\% & +5.70\% & 0.00 & 0.00 & 9.17 & 26.67  \\

% \multicolumn{9}{c}{\textbf{Proprietary Models}} \\
\modellogo{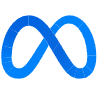}
Llama 3.3 70B Instruct 
& 6.67 & 16.67 & +0.77\% & +2.01\% & 0.00 & 0.00 & 19.17 & 40.00  \\

\modellogo{Figures/qwen.png}
Qwen3-4B-Instruct
& 0.00 & 0.00 & +1.67\% & +6.95\% & 0.00 & 0.00 & 13.33 & 26.67  \\

\modellogo{Figures/llama.png}
Llama 3.1 8B Instruct 
& 6.67 & 20.00 & +0.55\% & +3.07\% & 0.00 & 0.00 & 0.00 & 0.00  \\

\modellogo{Figures/glm.png}
GLM-4-9B-Chat  
& 0.00 & 0.00 & -0.18\% & +0.41\% & 0.00 & 0.00 & 0.00 & 0.00  \\

\bottomrule
\end{tabular}
}

\end{table*}

We evaluate over 15 trending LLMs on \odysseylite, 
encompassing proprietary frontiers such as Gemini 3 Pro Preview, Gemini 2.5 Pro~\citep{comanici2025gemini}, GPT-5, and Grok 4 Fast, alongside open-source series including DeepSeek-V3.2~\citep{liu2025deepseek}, gpt-oss-120b~\citep{openai2025gptoss}, Qwen3 series~\citep{yang2025qwen3}, Llama 3 series~\citep{grattafiori2024llama}, and GLM-4 series~\citep{glm2024chatglm}. Each test case is executed four times to report both \emph{Avg.@4} and \emph{Pass@4} success rates. Notably, for \textit{Repo System}, we use \emph{Best@4} to represent the best profit of 4 generated trajectories.
To manage the extreme interaction horizons, our standard prompts retain only the history of actions and environment feedback, omitting intermediate reasoning traces from previous steps to ensure context efficiency. Additional details regarding model specifications, reasoning effort, and prompt templates are provided in Appendix~\ref{appendix:baseline} and \ref{appendix:prompts}.

\begin{figure}[t]
\begin{minipage}{0.48\textwidth}
    \centering
    \includegraphics[width=\linewidth]{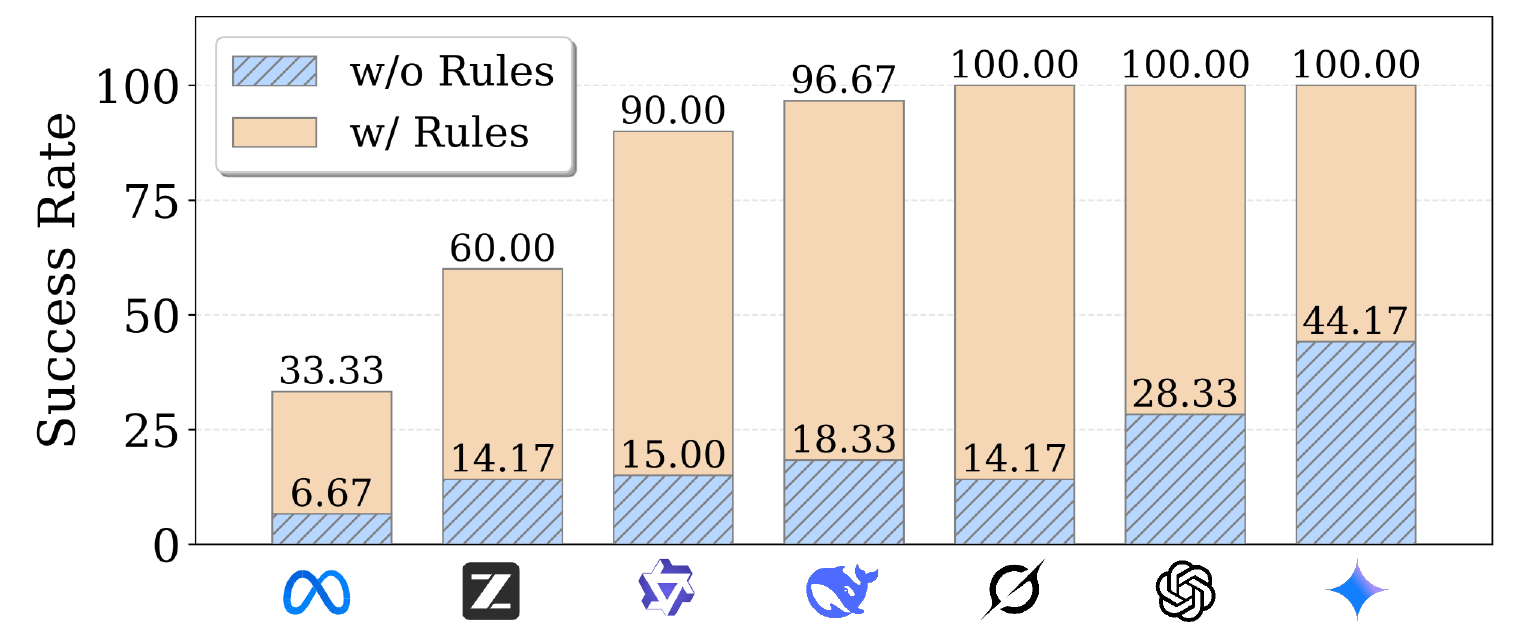}
    \caption{Success rate comparison of w/ and w/o rules in \textit{Turn On Lights}. We select Llama 3.3 70B Instruct, GLM-4-32B-0414, Qwen3-235B-A22B-Instruct, DeepSeek-V3.2, Grok 4 Fast, GPT-5, Gemini 3 Pro Preview for illustration.}
    \label{fig:rules_lights}
\end{minipage}
\hfill % 填充左右之间的空白
\begin{minipage}{0.48\textwidth}
    \captionof{table}{Failure mode of 47 failed trajectories from Gemini 3 Pro Preview, in which planning is not the bottleneck. Notably, each trajectory may contain multiple failure modes.}
    \label{tab:error_types}
    \centering
    % \vspace{-0.1em}
    \renewcommand{\arraystretch}{1.2}
    % 核心修改：用 \resizebox 包裹 tabular
    \resizebox{\linewidth}{!}{ % \linewidth 表示填满当前 minipage，! 表示高度自动按比例
      \begin{tabular}{l c c}
        \toprule
        \textbf{Failure Mode} & \textbf{Count} & \textbf{Percentage} \\
        \midrule
        Exploration Limitations & 38 & 80.85\% \\
        Memory Constrains & 27 & 57.45\% \\
        Behavior Stagnation & 13 & 27.66\% \\
        Planning Inability & 2 & 4.26\% \\
        \midrule
        \textbf{Total} & \textbf{47} & \textbf{100.00\%} \\
        \bottomrule
      \end{tabular}
    }
    
  \end{minipage}
    \vspace{-1em}
\end{figure}

\subsection{Main results.} 

\paragraph{The General Performance Gap.} Table~\ref{tab:main_results} illustrates a performance disparity between SOTA LLMs and human.
% While human participants demonstrate robust success through active exploration and strategic xxxxx, 
% LLMs exhibit a marked deficiency in tasks necessitating autonomous rule induction. 
Unlike human participants who successfully resolve tasks by isolating causal variables and distilling latent rules through instinctive trial-and-error,
LLMs exhibit a marked deficiency in autonomous rule induction.
This gap highlights a fundamental failure in current agents to internalize latent world dynamics from experience,
leading to significant performance degradation as interaction horizons extend.

% \vspace{-0.1in}

\vspace{-1em}
\paragraph{Proprietary Model and the Scaling Limit.}
Frontier proprietary models, led by {Gemini 3 Pro Preview}, 
consistently provides the current performance ceiling and substantially outperform other counterparts across most environments. 
% Other leading competitors, such as GPT-5, form a secondary tier that maintains a lead over open-source alternatives, yet they remain far below human proficiency.
Despite this advantage, the ubiquitous failure in \textit{Energy Dispatch} underscores a critical architectural limitation shared across the spectrum: an inability to synthesize periodic patterns over extended observation windows ($\sim$20 steps). This suggests that while increased scale enhances deductive compliance, 
it remains insufficient to overcome the inductive bottleneck for robust world-structure modeling.

\vspace{-0.1in}
\paragraph{Deductive Proficiency vs. Inductive Deficiency.}
% To isolate the root of failure, 
To disentangle the root of failure, 
we evaluate agents with explicit access to latent transition rules.
Figure~\ref{fig:rules_lights} shows that frontier models achieve near-perfect success when the underlying
logic is provided, yet falter significantly without it. 
This contrast identifies a fundamental asymmetry: 
LLMs excel at deductive reasoning but lack the inductive capacity to autonomously synthesize environment mechanics from experience. These findings confirm that the primary bottleneck in \odyssey is the discovery of world dynamics rather than the complexity of the task logic itself.

% \paragraph{General Performance.}
% Table~\ref{tab:main_results} reveals a substantial performance gap between leading LLMs and human performance, 
% particularly in tasks necessitating autonomous rule induction.
% While proprietary flagships can generally outperform their open-source counterparts on most tasks, 
% the ubiquitous failure in \textit{Energy Dispatch} underscores a critical architectural limitation: the inability to synthesize periodic patterns over extended observation windows ($\sim$20 steps).
% These results identify a fundamental inductive bottleneck, where agents struggle to internalize long-horizon temporal regularities despite demonstrating robust deductive compliance in standard instruction-following tasks.

% % \xfz{Main results too short !!!!}
% % \sqs{fxxk, 别催我，在弄了}

% \paragraph{Deductive Proficiency vs. Inductive Deficiency.}
% % To isolate the root of failure, 
% To disentangle the root of failure, 
% we evaluate agents with explicit access to latent transition rules. Figure~\ref{fig:rules_lights} shows that frontier models achieve near-perfect success when the underlying logic is provided,
% yet falter significantly without it. 
% This contrast identifies a fundamental asymmetry: 
% LLMs excel at deductive reasoning but lack the inductive capacity to autonomously synthesize environment mechanics from experience. These findings confirm that the primary bottleneck in \odyssey is the discovery of world dynamics rather than the complexity of the task logic itself.

%% file: Sections/6Analysis.tex
\vspace{-0.3cm}
\section{Analysis}
In this section, we first introduce the frequent failure modes of SOTA LLMs in Section~\ref{sec:failure_modes}, followed by detailed analyzes of how these phenomena degrade model performance in Section~\ref{sec:performance_saturation}, Section~\ref{sec:action_loops} and Appendix~\ref{app:memory_usage}. We include more analytical experiments, such as the difference between successful and unsuccessful trajectories in Appendix\ref{app:comparison_success_unsuccess}, step usage to finish a task in Appendix~\ref{app:step_distribution}, token efficiency for task completion in Appendix~\ref{app:token_efficiency}, comparison of in context learning in Appendix~\ref{app:in_context} and the results of \odysseychallenge in Appendix~\ref{app:odyssey_challenge_results}.

\begin{figure}[t]
\begin{minipage}{0.48\textwidth}
    \centering
    % \vspace{-1.5em}
    \includegraphics[width=\linewidth]{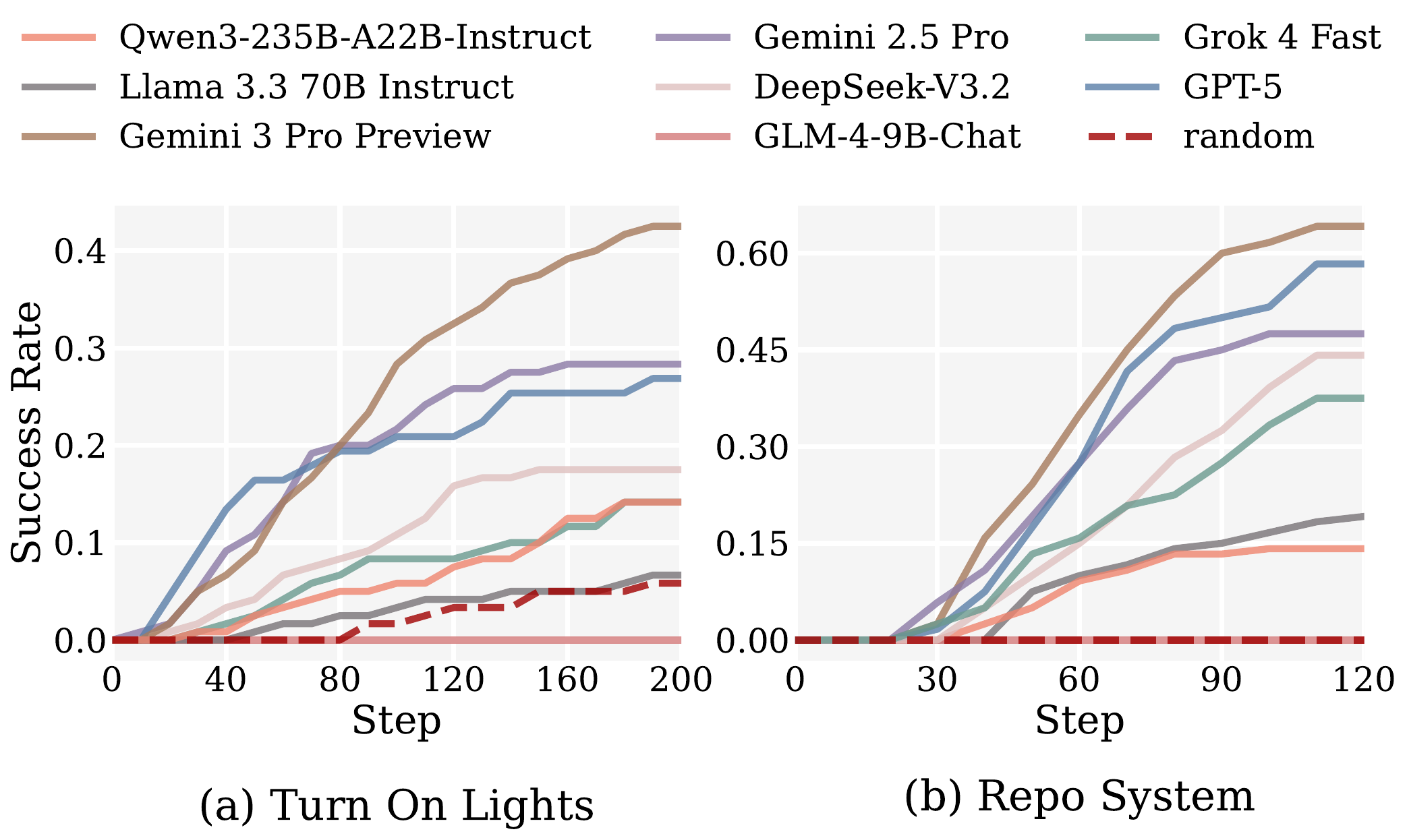}
    \vspace{-1.5em}
    \caption{Success Rate against step in two environments. Most of the success rate curve saturate with extended interaction steps. More results are in Appendix~\ref{appendix:evolution}.}
    \vspace{-0.9em}
    \label{fig:combined_progress}
\end{minipage}
\hfill
\begin{minipage}{0.48\textwidth}
\centering
    % \vspace{-2em}
    \includegraphics[width=\linewidth]{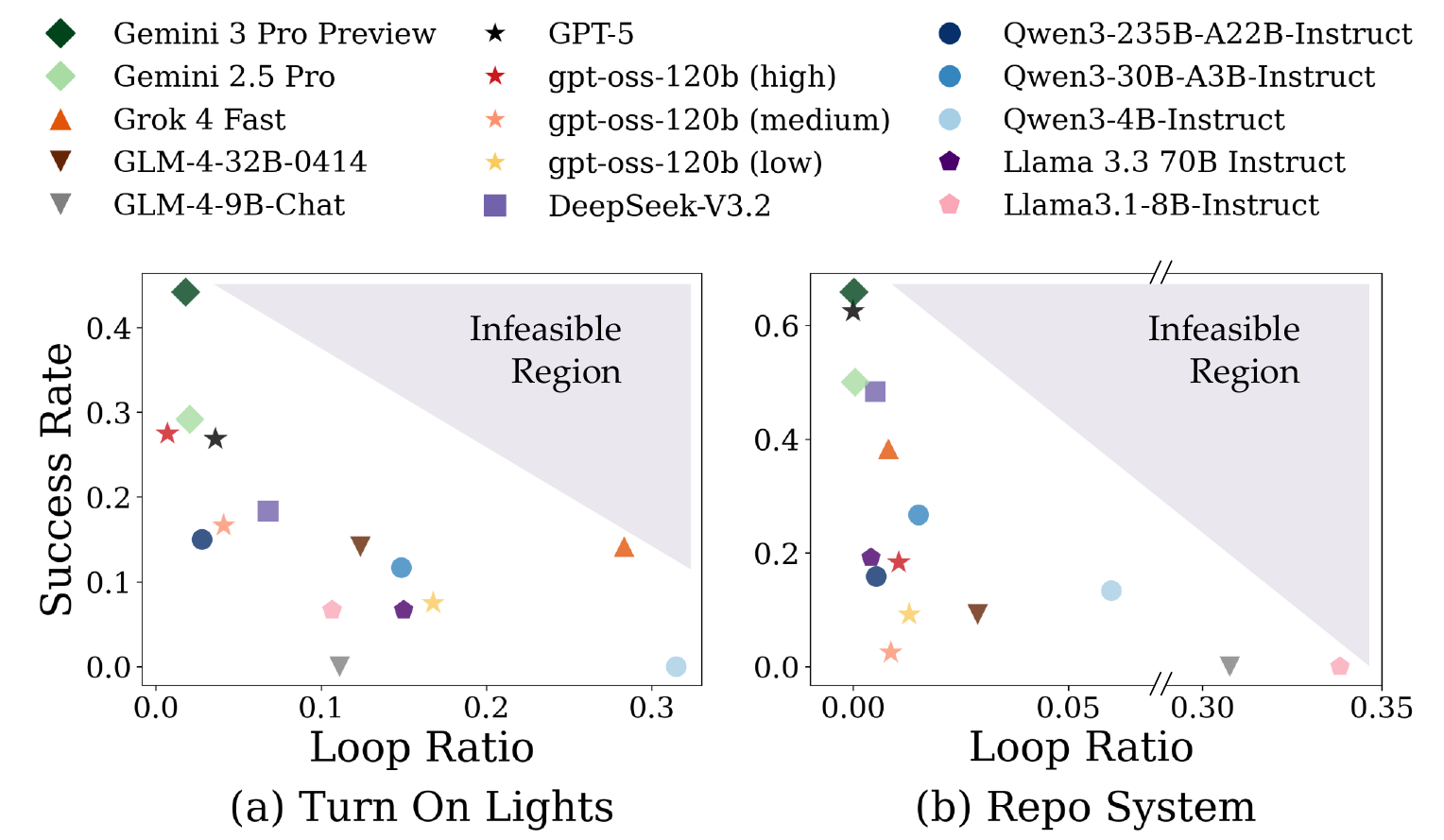}
    \vspace{-1.2em}
    \caption{Model performance is significantly related to loop ratio. Infeasible region indicates that a high Loop Ratio results in an inability to solve long-horizon inductive reasoning tasks.
    }
    \vspace{-1.4em}
    \label{fig:loop_sr}
\end{minipage}
\end{figure}

% \begin{figure}[t]
%     \centering
%     \includegraphics[width=\linewidth]{Figures/completion_matrix.pdf}
%     \vspace{-1.5em}
%     \caption{Task success status (based on \textit{pass@4}) of different tasks in \textit{Turn On Lights}. Each row represents: (a) Human, (b) Gemini3 Pro Preview, (c) GPT-5, (d) Gemini 2.5 Pro, (e) gpt-oss-120b (high), (f) DeepSeek-V3.2, (g) Grok 4 Fast, (h) Qwen3-235B-A22B-Instruct, (i) gpt-oss-120b (medium), (j) Qwen3-30B-A3B-Instruct, (k) GLM-4-32B-0414, (l) gpt-oss-120b (low), (m) Llama 3.3 70B Instruct, (n) Qwen3-4B-Instruct, (o) Llama 3.1 8B Instruct, (p) GLM-4-9B-Chat. \textcolor{green!25!black}{Dark green} cells indicate tasks solved by Human. \textcolor{green!60!black}{Green} cells indicate tasks solved by LLM agents. \textcolor{gray}{Gray} cells indicate unsolved tasks. for each subset (Easy, Medium and Hard), we report the average success rate across all LLMs.}
%     \label{fig:completion}
% \end{figure}

% \begin{figure}[t]
%     \centering
%     \includegraphics[width=\linewidth]{Figures/loop_sr.pdf}
%     \vspace{-1.5em}
%     \caption{Model performance is significantly related to loop ratio. Infeasible region indicates that a high Loop Ratio results in an inability to solve long-horizon inductive reasoning tasks.
%     }
%     \vspace{-1.4em}
%     \label{fig:loop_sr}
% \end{figure}

\subsection{Failure modes of current LLM agents}
\label{sec:failure_modes}

We manually analyze 47 failure trajectories from Gemini 3 Pro Preview, revealing four error categories in Table~\ref{tab:error_types}, which stem primarily from limitations in exploration, memory, and action adaptation.

The most prevalent issue is exploration limitation (38/47), where agents prematurely converge to local optima and default to suboptimal strategies without systematic hypothesis testing, which is detailed in Section~\ref{sec:performance_saturation}. 
Closely related are memory constraints (27/47), where agents successfully gather relevant information but fail to retrieve or prioritize key evidence over long interaction histories, which is detailed in Appendix~\ref{app:memory_usage}. 
We also observe behavior stagnation (13/47), where agents repeatedly execute invalid actions despite explicit negative feedback, especially in discrete-action environments, which is detailed in Section~\ref{sec:action_loops}. Notably, planning inability is rare (2/47), suggesting that forming high-level plans is not the primary bottleneck. 
% Moreover, we report the difference between successful and unsuccessful trajectories in Appendix\ref{app:comparison_success_unsuccess}.

\subsection{Exploration limitation: performance saturation in long-horizon}
\label{sec:performance_saturation}
As illustrated in Figure~\ref{fig:combined_progress}, the success rate curves reveal that the interaction budget beyond an initial exploratory phase yields negligible marginal gains for most models.
This plateau suggests a fundamental inductive bottleneck in long-horizon scenarios, 
where extended interaction fails to rectify the absence of a coherent internal world model. 
Furthermore, weaker models frequently underperform the random baseline, underscoring an inherent inability to extract latent regularities from environmental feedback. 
These findings indicate that the primary barrier is not interaction volume but the underlying capacity for inductive discovery.

\subsection{Behavior stagnation: action loops and inductive stagnation}
\label{sec:action_loops}
We reveal a prevalent failure mode characterized by persistent ``action loops,''
where models repeat invalid operations despite receiving negative environmental feedback.
As shown in Figure~\ref{fig:loop_sr}, a higher loop ratio directly correlates with diminished success rates, 
identifying a critical inability to capture hidden rules during interaction. 
This repetitive behavior signifies inductive stagnation, as agents fail to synthesize latent world laws from unsuccessful trials to refine their long-term strategy. 
Consequently, these cycles highlight the failure to transform trial-and-error into active discovery, underscoring the gap between deductive compliance and inductive world-modeling.
Details are in Appendix~\ref{appendix:loop_implementation}.

% We observe that SOTA models struggle to solve hard tasks that require reasoning over extended interaction steps, indicating a significant deficiency in handling long-horizon tasks. \yh{this conclusion seems the same as that in Sec 5.2} Notably, 

% \subsection{Failure Mode: Action Loop}
% We analyze the loop phenomenon in Figure~\ref{fig:loop_sr}, calculated as the ratio of steps in which the model repeats the previous invalid action. We can observe that a higher loop ratio introduces a lower success rate, indicating that these models may repeat the same failure, failing to inductively summarize the hidden rules. More loop ratio implementation details can be found in Appendix~\ref{appendix:loop_implementation}

%% file: Sections/7Conclusion.tex
\section{Conclusion}
This work introduces a paradigm shift in agentic evaluation by transitioning from deductive instruction-following to long-horizon, active and inductive modeling. We formalize abstract environment dynamics into four structural primitives and instantiate them via \odyssey, 
and then establish a standardized framework for the reproducible evaluation.
% of an agent's ability to discover latent transition laws.
The observed low-performance plateau across multiple flagships
% The observed performance saturation among multiple flagships 
reveals a fundamental inductive bottleneck that scaling alone cannot satisfy the necessity of moving beyond deductive compliance.
Future research toward more agentic intelligence should prioritize architectures capable of distilling latent transition laws from raw experience, bridging the gap between passive rule-following and active discovery in complex, dynamic worlds.

%% file: Sections/A_task_curation.tex
\section{Task curation}
\label{appendix:task_curation}
\subsection{{Turn On Lights} curation}
This task instantiates the discrete symbolic rules primitive by simulating a network of $N$ interdependent lights. The agent aims to reach the target configuration $s = \mathbf{1}$ (where all lights are illuminated) from an initial state $s_0 = \mathbf{0}$. The environment's dynamics are governed by a transition function $(s_{t+1}, r_t) = \mathcal{T}(s_t, a_t)$ that encapsulates unobservable regularities in the form of latent boolean logic. Success necessitates the systematic discovery of these hidden logical dependencies through strategic interaction and hypothesis testing.

\subsubsection{Structural configuration}
\paragraph{Dependency Structure Construction.}
Each light $L_i \in \mathcal{L} = \{L_0, \dots, L_{N-1}\}$ is associated with an activation condition $\phi_i$, a propositional formula over other light states. We define a strict partial order $\prec$ in $\mathcal{L}$ such that if $L_j$ appears in $\phi_i$, then $L_j \prec L_i$. This ensures that each light's condition only references lights with smaller indices:
\begin{equation}
    \phi_i = f(L_{j_1}, L_{j_2}, \ldots, L_{j_k}), \quad \text{where } j_1, \ldots, j_k < i.
\end{equation}

The minimal element $L_0$ is defined with a constant activation condition ($\phi_0 \equiv \text{True}$) and its role as the entry point of the causal chain is hidden. Specifically, we apply a random mapping $\sigma: \mathcal{L} \rightarrow \mathcal{L}$. Consequently, the agent cannot rely on numerical preference or ordering of IDs to infer the dependency graph; instead, it must perform systematic interactions to inductively reason and subsequently bootstrap its knowledge of the transition function $\mathcal{T}$.

% The minimal element $L_0$ is always toggleable ($\phi_0 \equiv \text{True}$), though its identity is obscured by a random permutation $\sigma: \mathcal{L} \rightarrow \mathcal{L}$ applied after generation. For subsequent lights, we construct $\phi_i$ using Boolean operators $\{\land, \lor, \neg\}$. For example: $\phi_3 = (L_0 \land \neg L_1) \lor L_2$.

\paragraph{Toggle Mechanism and State Transitions.}
Let $s \in \{0, 1\}^N$ denote the state vector. When an agent performs an action $a_t$ to toggle light $L_i$, the transition function $\mathcal{T}$ updates the state as follows:
\begin{equation}
s_{t+1, i} = 
\begin{cases} 
\neg s_{t, i} & \text{if } \phi_i(s_t) = \text{True} \\
s_{t, i} & \text{if } \phi_i(s_t) = \text{False}
\end{cases}
\end{equation}
If $\phi_i(s_t) = \text{False}$, the agent receives only a generic failure message. The presence of negation ($\neg$) introduces non-monotonic dynamics: turning a light \textit{on} may satisfy one condition while violating another, necessitating complex inductive reasoning to navigate the state space.

\subsubsection{Resolvability and diversity}
\paragraph{Guaranteed Resolvability.}
Every generated instance is verified to have at least one valid solution through an exhaustive search. An instance is accepted only if: (1) there exists a path to the goal state $s = \mathbf{1}$, and (2) the minimum steps required to finish the task is larger than a predefined threshold. The partial order structure provides a constructive guarantee for latent structure induction: since $L_0$ is always accessible and each subsequent $\phi_i$ depends only on its predecessors, a solution path always exists.

\paragraph{Task Diversity.}
To ensure a robust evaluation, we leverage the following parameters to generate a diverse suite of environment instances:

\begin{itemize}[leftmargin=*,itemsep=2pt]
    \item State Space Scaling ($N$): By modulating the number of lights $N$, we control the exponential growth of the state space $s \in \{0, 1\}^N$. This allows for the creation of tasks ranging from localized logic puzzles to complex systems with $2^N$ possible configurations.
    
    \item Logical Combination:
    % Diversity in environment dynamics is driven by the randomized construction of $\phi_i$.
    By varying the density of the predecessors $L_j \prec L_i$ and the specific mixture of $\{\land, \lor, \neg\}$ operators, each instance presents unique transition regularities.

\end{itemize}

\subsection{{AI Trading} curation}
This task instantiates the continuous stochastic dynamics primitive by simulating a multi-stock management scenario. Agents must autonomously infer the latent dependency matrix $\mathbf{W}$ to maximize cumulative reward $\sum r_t$. 
The challenge lies in performing statistical inference to unravel the underlying market signal $f$ from stochastic fluctuations $\epsilon$ over a fixed trading horizon. 
% \yh{remember to change s}

\subsubsection{Structural configuration}
\paragraph{Dependency Matrix Construction.}
The relationship between latent market factors and asset returns is modeled via a transition matrix $\mathbf{W} \in \mathbb{R}^{d \times K}$, where $d$ denotes the number of stocks and $K$ the number of unobserved market factors. At each time step $t$, the environment generates a factor change vector $z_t \in \mathbb{R}^K$. The resulting stock returns $s_{t+1}$ are governed by:
\begin{equation}
    s_{t+1} = \mathbf{W} z_t + \epsilon, \quad \epsilon \sim \mathcal{N}(0, \sigma^2)
\end{equation}
where $s \in \mathbb{R}^d$ represents the vector of price returns. The price of the stock $i$ at time $t+1$ is updated to $p_{t+1, i} = p_{t, i} + s_{t+1, i}$. The matrix $\mathbf{W}$ remains invariant within an episode to represent the unobservable regularities of the specific market task.

\paragraph{Trading Mechanism and State Transitions.}
The action $a_t$ consists of a set of buy/sell operations. Specifically, the agent's action $a_t$ specifies the buy/sell and quantity for each stock at every step. The transition function $\mathcal{T}$ then updates the agent's portfolio, which consists of available cash and a vector of stock holdings. The reward $r_t$ is determined by the sum of cash and the current market value of all held stocks. The agent must inductively reason about the hidden matrix $\mathbf{W}$ by observing how price changes correlate with market factors over time, enabling them to execute strategic, long-horizon trading decisions.

\subsubsection{Temporal trajectory}
To ensure reproducible and fair comparisons across different agents, the environment's stochastic elements are pre-determined within the task metadata. The complete timeline of the changes of $z$ and $\epsilon$ is generated at the beginning of each task and stored in the configuration. This approach ensures that while the market factors fluctuate, the environment's evolution remains deterministic for any given task instance. Consequently, the optimal trading strategy is theoretically computable with complete knowledge of the dependency matrix $\mathbf{W}$ and the pre-generated factor timeline, providing a consistent benchmark for evaluating agent performance.

\subsubsection{Resolvability and diversity.}
\paragraph{Guaranteed Resolvability.} There is no failure in this environment, and every trajectory finishes each task with different profit. So all the tasks are resolvable.

\textbf{Task Diversity.} We generate diverse tasks by modulating the following parameters:

\begin{itemize}[leftmargin=*,itemsep=2pt]
    \item Dimensionality Scalability ($d, K$): By scaling the number of stocks $d$ and market factors $K$, we control the complexity of the inference problem. A larger $d$ expands the action space, while a higher $K$ increases the difficulty of disentangling the latent factors from the observations.
    
    \item Dependency Sparsity: The density of non-zero entries in $\mathbf{W}$ determines the complexity of the structural relationships. Sparse matrices require the agent to identify a few factor-stock couplings, whereas dense matrices test the agent's ability to model global market correlations.
    
    \item Signal-to-Noise Ratio: By adjusting the variance of $\epsilon$, we modulate the information scarcity. This forces agents to distinguish between persistent structural signals defined by $\mathbf{W}$ and stochastic fluctuations.
    
    % \item Temporal Horizon: The episode length $T$ dictates the volume of observations available for induction. Longer horizons provide more data for refining the estimate of $\mathbf{W}$ but demand sustained accuracy in long-range planning.
\end{itemize}

\subsection{Energy Dispatch curation}

This task simulates an energy grid dispatch scenario where agents must allocate power generation resources to satisfy electricity demand while maintaining grid stability, budget constraints, and carbon emission targets. Given 4 generation sources (thermal, wind, solar, and battery) with respective capacities, the agent specifies daily rated power allocations $a_t \in \mathbb{R}^K$ over a horizon $H$. The challenge lies in adapting to time-varying renewable efficiency while balancing multiple competing objectives.

\subsubsection{Structural configuration}
To evaluate the agent's capacity for world-structure induction, we model the discrepancy between planned and realized power through a latent efficiency vector $E_t$ and real generated energy  
$P_{\text{real}} \approx a_t \odot E_t$.

\paragraph{Multi-Objective Constraints and State Transitions.}
Each day, the environment evaluates the following constraint satisfaction criteria:
\begin{itemize}[leftmargin=*]
    \item Demand: Total realized supply $\sum P_{\text{real}, t}$ must meet demand $D_t$.
    \item Budget: Total cost must not exceed budget $B_t$.
    \item Carbon: The ratio of cumulative  generated thermal energy $\frac{\sum_{\tau=1}^H P_{\text{thermal}, \tau}}{\sum_{\tau=1}^H (P_{\text{thermal}, \tau}, P_{\text{wind}, \tau}, P_{\text{solar}, \tau})}$ must remain below target $\tau_c$.
    \item Stability: Grid stability penalizes large allocation changes between consecutive days. Additionally, a violation of either demand or budget could significantly influence the grid stability.
\end{itemize}
Consecutive violations (3 days in \odysseylite) of demand or budget constraints trigger early termination, simulating an irreversible grid collapse.

% \paragraph{Dependency Structure Construction.}
% We model the relationship between rated power allocations and actual generation through efficiency curves. At each time step $t$, the realized power output $P_{\text{real}, t}$ follows:
% \begin{equation}
%     P_{\text{real}, t} = a_t \odot E_t
% \end{equation}
% where $E_t \in \mathbb{R}^K$ denotes the efficiency vector. Thermal generation maintains near-constant efficiency ($E_{\text{thermal}, t} \approx 1.0$), while renewable sources exhibit periodic fluctuations reflecting weather patterns:
% \begin{equation}
%     E_{\text{wind/solar}, t} = E_{\text{base}}(t \mod T) + \delta_{\text{cycle}} + \epsilon_t
% \end{equation}
% where $T \in [15, 25]$ in \odysseylite is the latent period length, $E_{\text{base}}$ represents a piecewise-constant base pattern within one period, $\delta_{\text{cycle}}$ is a per-cycle offset, and $\epsilon_t \sim \mathcal{N}(0, \sigma^2)$ captures daily noise. These efficiency curves are hidden from agents, who must inductively infer the periodic patterns by comparing observed $P_{\text{real}, t}$ against their previous actions $a_t$.

\subsubsection{Temporal trajectory}

For renewable sources wind and solar, $E_t$ is governed by a five-level hierarchical generative process that simulates multi-scale temporal dependencies:

\begin{equation}
    E_{\text{wind/solar}, t} = \text{Clip} \left( E_{\text{base}}(t \mod T) + \delta_{\lfloor t/T \rfloor} + \epsilon_t \right)
\end{equation}

where $T \in [15, 25]$ denotes the hidden period length randomly sampled. The generative logic is structured as follows:

\begin{itemize}[leftmargin=*]
    \item Base Pattern ($E_{\text{base}}$):
    % This represents the "signature" of a single period. 
    It is constructed as a piecewise-linear sequence where each segment (2--5 days) is assigned a random efficiency baseline to simulate the consistency of weather. To introduce intra-period complexity, we add  stochastic \textit{spikes} (with a probability of 5\%) representing extreme weather.
    \item Cyclic Variation ($\delta_{\lfloor t/T \rfloor}$): To prevent the agent from relying on the memorization of a fixed efficiency curve, each full cycle incorporates a unique random offset $\delta$. This mimics the changes between different months or seasons, requiring the agent to continuously recalibrate its internal model.
    \item Micro-Fluctuations ($\epsilon_t$): A high-frequency Gaussian noise $\epsilon_t \sim \mathcal{N}(0, 0.01^2)$ is added to the daily output, simulating real-world stochastic.
    \item Value Clipping: Finally, the efficiency is clipped to domain-specific ranges ($[0.6, 1.05]$ for wind and $[0.65, 1.1]$ for solar) to ensure physical realism and prevent extreme outliers.
\end{itemize}

Additionally, the efficiency of thermal is near to 1 with minimum fluctuation and the battery efficiency is a constant 1 without any fluctuation. A positive action $a_{\text{battery}, t} > 0$ denotes discharging stored energy to the grid, while $a_{\text{battery}, t} < 0$ denotes charging from excess generation.

The efficiency curves are unobservable to the agent. Consequently, the agent must inductively reason for the stable periodic signal from transient fluctuations by comparing its historical rated actions $a_t$ with the real power outputs $P_{\text{real}, t}$.

To ensure reproducibility and fair comparison, all time-varying factors ($D_t, B_t, E_t$) are pre-determined as fixed sequences within the task metadata. This ensures that the environment's evolution is fully deterministic given the agent's sequence of actions, allowing for identical experimental trials across different models.

\subsubsection{Resolvability and diversity}

\paragraph{Guaranteed Resolvability.}
Each instance is guaranteed to be feasible through careful parameter design. Total capacity exceeds peak demand, and budgets are set as $B_t = 4.2 \times D_t$ to provide adequate financial space. Efficiency values and objective thresholds are tuned to ensure that a foresighted dispatch policy can successfully complete the horizon $H$ without triggering early termination, while leaving substantial space for diverse strategies.

\paragraph{Task Diversity.}
Our generation framework provides fine-grained control over task difficulty through several orthogonal parameters:
\begin{itemize}[leftmargin=*]
    \item Temporal Dynamics: Varying period lengths $T$ for wind and solar creates complex interference patterns that the agent must disentangle.
    \item Constraint Tightness: Adjusting targets $\tau_c$ and $\tau_s$ modulates the precision required to balance competing carbon and stability goals.
    % \item \textbf{Information Asymmetry}: Agents only observe historical $P_{\text{real}, t}$ and current $D_t, B_t$. They must internalize the latent efficiency $E_t$ to anticipate future generation capacity, mirroring real-world energy forecasting challenges.
\end{itemize}

\subsection{{Repo System} curation}
This task instantiates the relational graph structures primitive by simulating a software repo dependency resolution scenario. The agent must discover a valid configuration of packages that satisfies all latent constraints in a dependency graph $G = (V, E)$.

\subsubsection{Structural configuration}
\paragraph{Dependency Graph Construction.}
The environment is defined by a latent graph $G = (V, E)$, where each node $v \in V$ represents a specific version of a software package. Edges $E$ represent directed compatibility constraints; for instance, an edge $(v_i, v_j)$ may indicate that version $v_i$ of Package A requires version $v_j$ of Package B. To ensure a structured challenge, we generate a random topological ordering over packages to prevent circular dependencies, while allowing \textit{version-level} constraints to create complex requirements.

To simulate ubiquitous dependencies on foundational libraries (e.g., NumPy, PyTorch), we designate $1$-$2$ packages as base libraries at the root of the topological order. A proportion of other packages depend on these base libraries with varying version constraints. There are two types of constraints: (1) the base library and the dependent library should be the same version. For example, the version of base library A is 1.1, then the version of dependent package B should be 1.1 as well. (2) the base library and the dependent library should be the same main version. For example, the version of base library A is 1.1, then the version of dependent package B should be 1.X.

\paragraph{Transition Logic and Side Effects.}
The state represents the set of currently installed package versions. When an agent issues an installation action $a_t$, the transition function $s_{t+1} = \mathcal{T}(s_t, a_t)$ simulates a rigorous resolution process. Unlike simple state updates, $\mathcal{T}$ may induce side effects, such as automatically upgrading, downgrading, or uninstalling conflicting packages to satisfy the constraints in $G$.
We implement four types of resolution behaviors: (1) Ensure: automatically install missing dependencies; (2) Force-high/low: coercing dependencies to extreme compatible versions; and (3) Pin: locking a package to a specific version. These behaviors introduce non-monotonic dynamics, where the sequence of actions $a_1, a_2$ may result in a different final state $s$ than $a_2, a_1$, forcing the agent to reason about the order of interventions.

\subsubsection{Resolvability and diversity}
\paragraph{Guaranteed Resolvability.}
Every instance is verified to have at least one valid goal state through a solution-first generation strategy. We first sample a ground-truth configuration and then construct the edges $E$ such that they provably contain this solution. All dependencies and  constraints are then generated to provably include the ground-truth packages version.

\begin{table*}[t]
\centering
\caption{Action Space of and description for \odyssey. All parameters are wrapped in \texttt{<>}.}
\label{tab:action_space}
\begin{tabular}{l p{5.5cm} p{\dimexpr\textwidth-9.5cm\relax}} 
\toprule
\textbf{Environment} & \textbf{Action} & \textbf{Description} \\
\midrule
\multirow{3}{*}{Turn On Lights}  & \multirow{3}{*}{\texttt{Toggle <light>}}  & The parameter \texttt{light} is the specific index of the light you want to toggle. This action is successfully executed only if its latent logical condition is satisfied. \\
\midrule
\multirow{6}{*}{AI Trading}  & 

\texttt{\{} \newline
\hspace*{1em} \texttt{"Buy":<stocks, shares>,} \newline
\hspace*{1em} \texttt{"Sell":<stocks, shares>} \newline
\texttt{\}}
& Purchase specific shares of stock and sell specific shares of stock. Multiple stocks can be acquired or sold simultaneously. The buy action is successful only if the cash is enough. If the selling shares exceed the current holdings, the environment defaults to selling the entire existing position. \\
\midrule
\multirow{8}{*}{Energy Dispatch}  & 
\texttt{\{} \newline
\hspace*{1em} \texttt{"Thermal":<p$_1$>,} \newline
\hspace*{1em} \texttt{"Wind":<p$_2$>,} \newline
\hspace*{1em} \texttt{"Solar":<p$_3$>,} \newline
\hspace*{1em} \texttt{"Battery":<p$_4$>} \newline
\texttt{\}}
& Dispatch \texttt{p}$_1$, \texttt{p}$_2$, \texttt{p}$_3$ units of thermal, wind, and solar power respectively. For \texttt{p$_4$}, a negative value represents charging and positive represents discharging. Notably, charging is truncated upon reaching the battery's max capacity, and the discharging is limited to the available capacity if it exceeds the current state of charge. The cost of the four energies should be limited within the total budget.\\
\midrule
\multirow{9}{*}{Repo System} 
& \multirow{2}{*}{\texttt{repo tree/ls}} & Inspect the repository's directory structure to identify script paths. \\
\cmidrule{2-3}
& \multirow{2}{*}{\texttt{pip install <package>}} & Install the \texttt{package}. Supports specific version (\texttt{==}) and range constraints ($>$, $>=$, $<$, $<=$). \\
\cmidrule{2-3}
& \texttt{pip uninstall <package>} & Remove the specified package from the configuration. \\
\cmidrule{2-3}
& \texttt{pip list} & List all installed packages with version identifiers. \\
\cmidrule{2-3}
& \multirow{2}{*}{\texttt{python <script>}}  & Execute a specified \texttt{script} to verify configuration or trigger entry points. \\
\bottomrule
\end{tabular}

% \medskip

\end{table*}

\subsection{Action space for \odyssey}
For the four environments in \odyssey: \textit{Turn On Lights}, \textit{AI Trading}, \textit{Energy Dispatch}, and \textit{Repo System}, we list detailed action space and their description in Table~\ref{tab:action_space}.

Notably, the agent can only execute one action in \textit{Turn On Lights} and \textit{Repo System} environments. 

For \textit{AI Trading} environments, the agent should first sell and then buy stocks within one step.

For \textit{Energy Dispatch} environment, the agent can plan for thermal, wind, solar, and battery together. However, charging and discharging cannot be executed simultaneously  in one step.

\subsection{License}
\label{sec:license}
We agree to release our datasets under CC-BY 4.0 license.

\subsection{Limitation}
\label{app:limitation}
To ensure reproducibility and fair comparison, all stochastic elements in each environment are pre-determined as fixed sequences within the task metadata. While this guarantees deterministic evaluation across agents, it removes the non-stationary dynamics inherent in real interactive systems, where environment responses may evolve or co-adapt with the agent over time.

%% file: Sections/B_more_analysis.tex
\section{Details of analysis}

\begin{table*}[t]
\centering

\caption{Performance comparison between w/o rules and w/ rules settings.  We provide three different reasoning effort of gpt-oss-120b. For \textit{AI Trading} environment, we report the profit rate and pass@4 is calculated based on the highest profit of each task. For other three environments, we report the success rate. \textcolor[HTML]{289BA2}{Colored Rows} represent proprietary models. }
\label{tab:rule_comparison}

\resizebox{\textwidth}{!}{
\begin{tabular}{lcccccccc}
\toprule
\multirow{2}{*}{Model} 
& \multicolumn{2}{c}{Turn On Lights} 
& \multicolumn{2}{c}{AI Trading} 
& \multicolumn{2}{c}{Energy Dispatch} 
& \multicolumn{2}{c}{Repo Management}  \\
\cmidrule(lr){2-3} \cmidrule(lr){4-5} \cmidrule(lr){6-7} \cmidrule(lr){8-9}
& w/o rules & w/ rules
& w/o rules & w/ rules
& w/o rules & w/ rules 
& w/o rules & w/ rules  \\
\midrule

\rowcolor{mycellcolor2}
\modellogo{Figures/gemini.png}
Gemini 3 Pro Preview   
& 44.17 & 100.00 
& +67.71\% & +135.48\% 
& 30.00 & 16.67 
& 65.83 & 96.67  \\

\rowcolor{mycellcolor2}
\modellogo{Figures/gpt.png}
GPT-5    
& 28.33 & 100.00 
& +17.32\% & +132.02\% 
& 23.33 & 13.33 
& 62.50 & 96.67 \\

\rowcolor{mycellcolor2}
\modellogo{Figures/gemini.png}
Gemini 2.5 Pro    
& 29.17 & 100.00 
& +33.02\% & +126.24\% 
& 10.83 & 20.00 
& 50.00 & 96.67    \\

\modellogoo{Figures/gpt.png}
gpt-oss-120b (high) 
& 27.50 & 100.00
& +23.27\% & +141.63\%
& 0.00 &  0.00
& 18.33 & 50.00  \\

\modellogo{Figures/deepseek.png}
DeepSeek-V3.2 
& 18.33 & 96.67 
& +8.62\% & +118.88\% 
& 0.00 & 0.00 
& 48.33 & 56.67  \\

\rowcolor{mycellcolor2}
\modellogo{Figures/grok.png}
Grok 4 Fast
& 14.17 & 100.00 
& +5.70\% & +62.96\% 
& 0.00 & 0.00 
& 38.33 & 36.67 \\

\modellogo{Figures/qwen.png}
Qwen3-235B-A22B-Instruct 
& 15.00 & 90.00 
& +11.26\% & +123.66\% 
& 0.00 & 0.00 
& 15.83 & 33.33  \\

\modellogo{Figures/qwen.png}
Qwen3-30B-A3B-Instruct 
& 11.67 & 50.00 
& +4.76\% & +94.00\% 
& 0.00 & 0.00 
& 26.67 & 43.33  \\

\modellogoo{Figures/gpt.png}
gpt-oss-120b (medium)
& 16.67 & 100.00 
& +3.21\% & +100.27\% 
& 0.00 & 0.00 
& 2.50 & 30.00  \\

\modellogo{Figures/glm.png}
GLM-4-32B-0414  
& 14.17 & 60.00 
& +3.14\% & +18.50\% 
& 0.00 & 0.00 
& 9.17 & 6.67  \\

\modellogoo{Figures/gpt.png}
gpt-oss-120b (low)
& 7.50 & 36.67 
& +2.02\% & +26.40\% 
& 0.00 & 0.00 
& 9.17 & 26.67  \\

\modellogo{Figures/llama.png}
Llama 3.3 70B Instruct 
& 6.67 & 33.33 
& +0.77\% & -0.93\% 
& 0.00 & 0.00 
& 19.17 & 50.00  \\

\modellogo{Figures/qwen.png}
Qwen3-4B-Instruct
& 0.00 & 10.00 
& +1.67\% & +60.52\% 
& 0.00 & 0.00 
& 13.33 & 16.67  \\

\modellogo{Figures/llama.png}
Llama 3.1 8B Instruct 
& 6.67 & 10.00 
& +0.55\% & +0.18\% 
& 0.00 & 0.00 
& 0.00 & 3.33  \\

\modellogo{Figures/glm.png}
GLM-4-9B-Chat  
& 0.00 & 0.00 
& -0.18\% & +0.34\% 
& 0.00 & 0.00 
& 0.00 & 0.00  \\

\bottomrule
\end{tabular}
}

\end{table*}

\subsection{Detailed comparison between w/ and w/o rules}
\label{appendix:rule_comparison}
To additionally demonstrate the performance between w/ and w/o rules for SOTA models, we provide detailed results in Table~\ref{tab:rule_comparison}. From the extended results, we argue the same conclusion that SOTA models are strong deductive reasoners rather than good inductive reasoners.

\begin{figure*}[h]
    \centering
    \includegraphics[width=\linewidth]{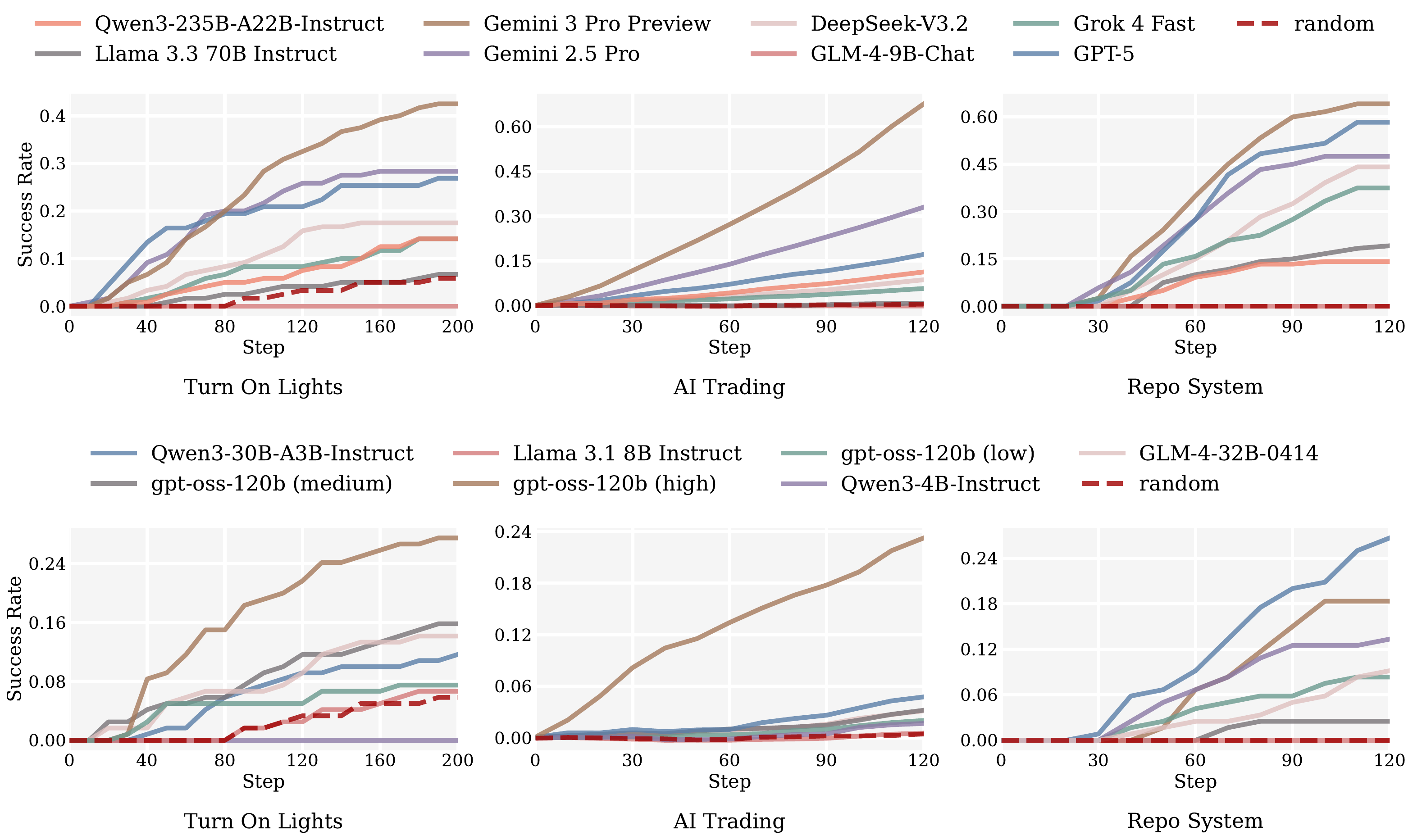}
    \caption{All data for Success Rate against Step. We do not plot for \textit{Energy Dispatch} environment due to its complex success conditions.}
    \label{fig:all_step}
\end{figure*}

\subsection{Comparison between successful and unsuccessful trajectories}
\label{app:comparison_success_unsuccess}
We manually analyze 16 T\textit{urn On Lights} problems where Gemini 3 Pro Preview exhibit partial success rather than absolute success or failure. We find that only two problems are due to random guesses. Moreover, this intra-problem variance mainly stems from exploration limitations (unstable exploration). For example, if activating light3 requires light2 (on) and light1 (off), and the agent activates light3 right after activating light2, it may wrongly conclude that only light2 is necessary, ignoring light1.

\subsection{Reasoning boosts inductive reasoning}
In Table~\ref{tab:main_results}, we further compare gpt-oss-120b across varying reasoning budgets (low, medium, and high). The results reveal that LLMs demonstrate better inductive performance with more reasoning budget. For example, in \textit{Turn On Lights}, gpt-oss-120b (high) achieves average success rate of 27.50\%, while gpt-oss-120b (medium) achieves 16.67\% and gpt-oss-120b (low) achieves 7.50\%. 

\subsection{More results and implementation details of evolution progress}
\label{appendix:evolution}
\paragraph{More Results of Evolution Progress.}We provide detailed results for 15 models and a random baseline in Figure~\ref{fig:all_step}. We can observe that: (1) In \textit{Turn On Lights} and \textit{Repo System} environments, most models demonstrate saturation as the interaction step increases, indicating that long-horizon remains a bottleneck for SOTA LLM models. (2) Some models, e.g. GLM-4-9B-Chat, perform equivalent or even worse than random baseline, indicating their inability of discovering hidden rules of the environment. (3) Models demonstrate different performance curves, suggesting that we should evaluate each model from multiple timestamps.

\paragraph{Implementation Details.}
For the \textit{Turn On Lights} and \textit{Repo Management} environments, we plot evolution curves with the interaction step index on the x-axis and the cumulative task success rate on the y-axis, defined as the proportion of successfully completed tasks from the beginning to the current step. This formulation captures the progression of task completion over time and reflects the agent’s inductive reasoning capability as the interaction horizon increases. At any given step, a higher y-axis value indicates a stronger inductive reasoning performance in the current step. 
For the \textit{AI Trading} environment, the y-axis represents the profit rate relative to the initial capital, which is then averaged across all tasks in this environment.
We do not report evolution curves for the \textit{Energy Dispatch} environment, as task success is determined by complex, multi-constraint conditions that do not admit a well-defined step-wise success metric.

Notably, as shown in the results above, model performance varies across different interaction steps, indicating that evaluation at a single step is insufficient. Therefore, it is necessary to assess model performance across multiple steps to obtain a more comprehensive and reliable understanding of their inductive reasoning capabilities.

\begin{figure*}[h]
    \centering
    \includegraphics[width=\linewidth]{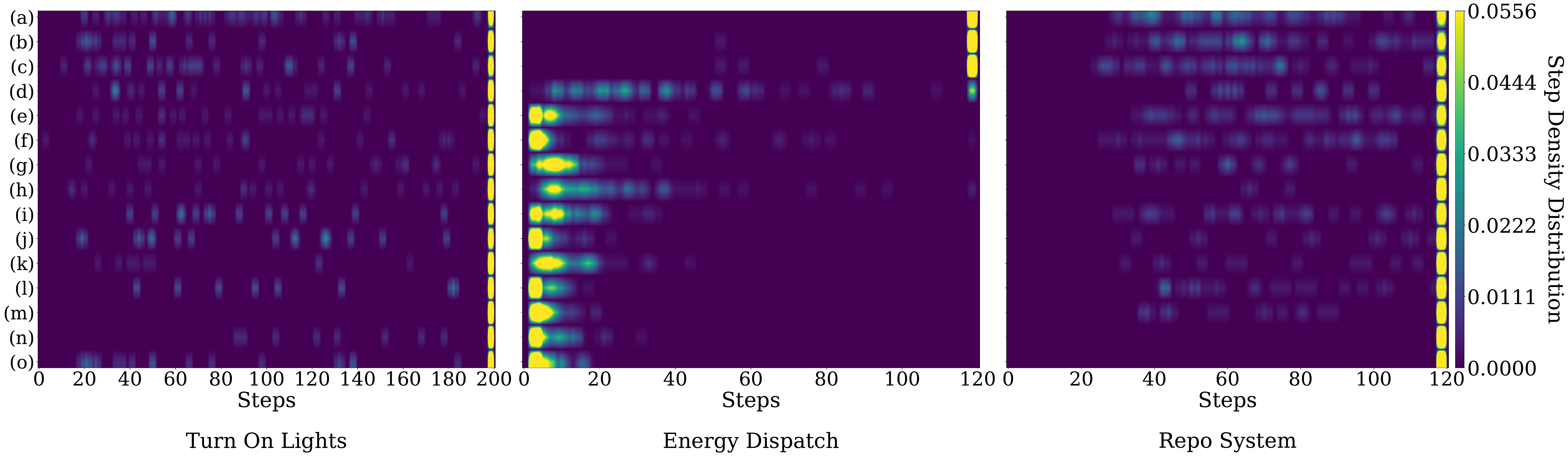}
    \caption{Step density distribution for LLM models: (a) Gemini 3 Pro Preview, (b) GPT-5, (c) Gemini 2.5 Pro, (d) gpt-oss-120b (high), (e) DeepSeek-V3.2, (f) Grok 4 Fast, (g) Qwen3-235B-A22B-Instruct, (h) gpt-oss-120b (medium), (i) Qwen3-30B-A3B-Instruct, (j) GLM-4-32B-0414, (k) gpt-oss-120b (low), (l) Llama 3.3 70B Instruct, (m) Qwen3-4B-Instruct (n) Llama 3.1 8B Instruct, (o) GLM-4-9B-Chat.}
    \label{fig:appendix_step_distribution}
\end{figure*}

\begin{figure*}[h]
    \centering
    \includegraphics[width=\linewidth]{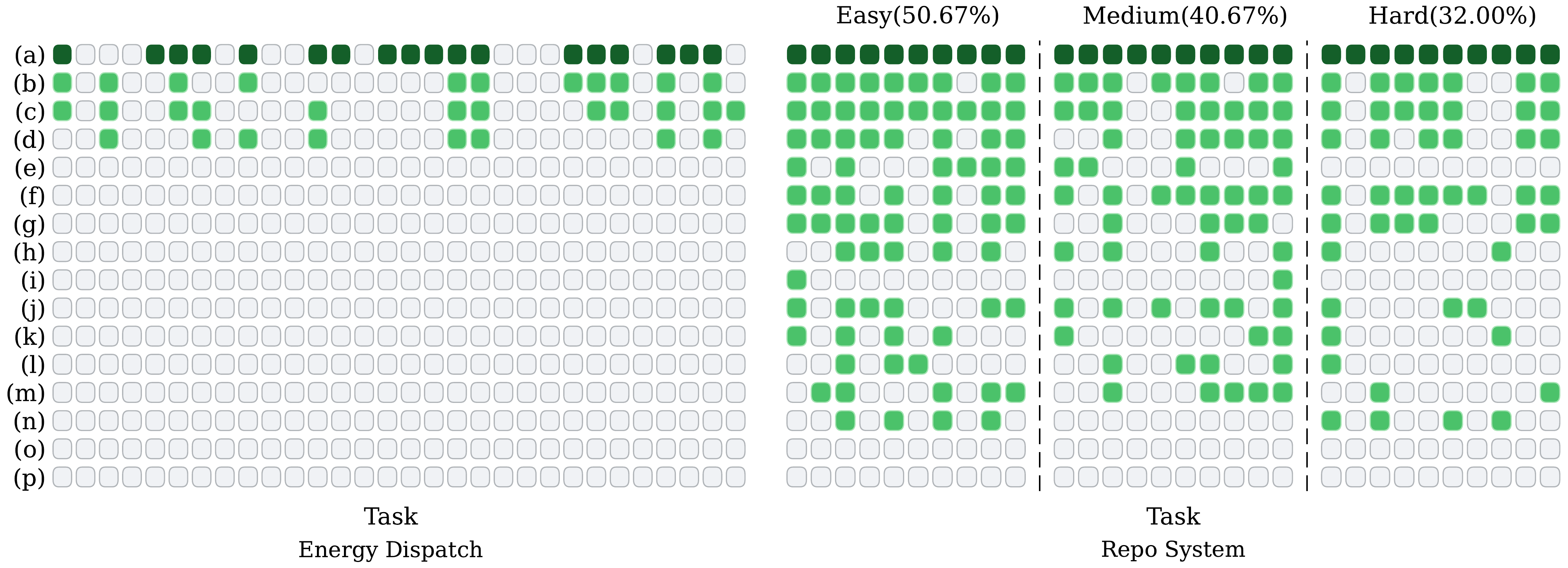}
    \caption{Task success status (based on \textit{pass@4}). Each row represents: (a) Human, (b) Gemini3 Pro Preview, (c) GPT-5, (d) Gemini 2.5 Pro, (e) gpt-oss-120b (high), (f) DeepSeek-V3.2, (g) Grok 4 Fast, (h) Qwen3-235B-A22B-Instruct, (i) gpt-oss-120b (medium), (j) Qwen3-30B-A3B-Instruct, (k) GLM-4-32B-0414, (l) gpt-oss-120b (low), (m) Llama 3.3 70B Instruct, (n) Qwen3-4B-Instruct, (o) Llama 3.1 8B Instruct, (p) GLM-4-9B-Chat. \textcolor{green!25!black}{Dark green} cells indicate tasks solved by Human. \textcolor{green!60!black}{Green} cells indicate tasks solved by LLM agents. \textcolor{gray}{Gray} cells indicate unsolved tasks. We report the average success rate in \textit{Repo System} across all LLMs for each subset (Easy, Medium and Hard).}
    \label{fig:appendix_task_success}
\end{figure*}

\subsection{More results and implementation details of 
task success status}
\label{appendix:task_success_status}

% \paragraph{More Results of Task Success Status.}
We provide detailed results of task success status for 14 models in \textit{Energy Dispatch} and \textit{Repo System} in Figure~\ref{fig:appendix_task_success}. We can observe that there is a significant gap between human and SOTA LLMs. Notably, most LLMs fail all tasks in \textit{Energy Dispatch}, indicating an inability of long-horizon inductive reasoning.

\subsection{Analyze of task success status}
We provide an overview of solved tasks and unsolved tasks in Figure~\ref{fig:appendix_task_success}. Additionally, for \textit{Repo System}, we simplify the difficulty representation using the number of required packages and separate all tasks into easy, medium and hard sets based on lights number . Human can solve all the 30 tasks, while the best LLM (Gemini 3 Pro Preview) can only solve 25 tasks. Notably, the we do not simplify or provide the difficulty level of \textit{Energy Dispatch} due to its complex multi-object goal.

% \paragraph{Implementation Details of Task Success Status.}
% The difficulty of \textit{Turn On Lights} task is decided by multiple conditions and we simplify it using the number of lights in Figure~\ref{fig:completion}. For \textit{Repo System}, we also simplify the difficulty representation using the number of required packages. Notably, the we do not simplify or provide the difficulty level of \textit{Energy Dispatch} due to its complex multi-object goal.

% 导言区确保包含：
% \usepackage{graphicx, booktabs, multirow}

\subsection{Memory constrain: the impact of memory usage}
\label{app:memory_usage}
We conduct a comparison experiment in Repo System environment, where different length of interaction history is provided to the agent, which is detailed in Table~\ref{tab:long_horizon_benchmark}. The following results demonstrate that the best performance occurs with about 100 turns of context window, indicating that current agents cannot inductively reasoning across overlong memory (about 200 turns). And merely about 50 turns cannot provide sufficient information for inductive reasoning.

\begin{table*}[htbp]
  \centering
  \caption{Performance in \textit{Repo System} with different memory usage.} % 标题移至此处
  \label{tab:long_horizon_benchmark}
  \resizebox{\textwidth}{!}{ % 强制缩放至页面宽度
  \begin{tabular}{lcccccccc}
    \toprule
    \multirow{2}{*}{Model} & \multicolumn{4}{c}{Avg@4} & \multicolumn{4}{c}{Pass@4} \\ % 一级表头
    \cmidrule(lr){2-5} \cmidrule(lr){6-9}
    & 200 turns & 150 turns & 100 turns & 50 turns & 200 turns & 150 turns & 100 turns & 50 turns \\ % 二级表头
    \midrule
    Gemini 3 Pro Preview & 65.83 & 70.00 & 76.67 & 70.83 & 80.00 & 83.33 & 86.67 & 83.33 \\
    gpt-oss-120b-high    & 18.33 & 19.17 & 23.33 & 11.67 & 33.33 & 33.33 & 36.67 & 20.00 \\
    \bottomrule
  \end{tabular}
  }
  
\end{table*}

\subsection{Loop implementation details}
\label{appendix:loop_implementation}
Given a trajectory dataset, a global measure of interaction volume is obtained by aggregating the total number of executed actions across all trajectories. 
For \textit{Turn On Lights} and \textit{Repo System} environments, we additionally analyses the Loop Ratio for each model.
To characterize the inability of inductive reasoning, we focus on repetitive execution of previously taken action that yields no effective task progress. 

Concretely, we define the state as the on/off of each light for \textit{Turn On Lights} environment and the version of each package for \textit{Repo System} environment.  Within each trajectory, we then detect whether an agent replays the same pair (state, action)  immediately after completing it and the two interactions make no task progress. We then sum the action counts of all such immediately repeated pairs and normalize this quantity by the total number of actions. This normalized proportion defines the Loop Ratio, which measures how much of the agent's interaction is spent on consecutive, unproductive repetition.

Under this formulation, a smaller Loop Ratio indicates that the agent is more capable to find hidden rules through inductive reasoning.

\subsection{Eliciting inductive reasoning is NOT trivial}
\label{app:in_context}
We conduct experiments with in-context examples in \textit{Repo System} environment. We additionally provide one-shot example consisting of 12 actions for completion. Results in Table~\ref{tab:in_context_fix}    demonstrate that merely providing in-context examples cannot strengthen inductive reasoning ability of agents. More techniques such as memory management or systematically exploration encouragement should be involved in future agents.

\subsection{Step distribution analysis}
\label{app:step_distribution}
We plot the distribution of the total steps number required to complete the task for each trajectory, including both successful and failed trajectories. 
In the \textit{Turn On Lights} and \textit{Repo System} environments, step usage serves as a proxy for inductive reasoning efficiency, where fewer steps indicate stronger inductive reasoning capability. In the \textit{Energy Dispatch} environment, step usage corresponds to the number of days during which energy is supplied without violating the three-day consistency constraint. A higher step usage represents a better inductive reasoning capability. Notably, every trajectory in \textit{AI Trading} environment has 120 steps in \odysseylite, so we do not plot a step distribution figure for this environment. The results are in Figure~\ref{fig:appendix_step_distribution}.

We can observe a sharp concentration around max step limit (200 steps for \textit{Turn On Lights} and 120 steps for \textit{AI Trading}), indicating that most SOTA models demonstrate limited inductive reasoning capability in this environment and can not solve the task within pre-defined steps limit.

Additionally, in \textit{Energy Dispatch} environment, most models exhibit step distributions that are sharply concentrated at relatively small values, indicating an inability to satisfy the demand and budget constraints for three consecutive days. 
Conversely, for Gemini 3 Pro Preview, Gemini 2.5 Pro, and GPT-5, the step distribution is sharply concentrated around 120 steps, suggesting that these models can satisfy the demand and budget constraints over the evaluated horizon. Notably, task success is additionally governed by the carbon and stability constraints and sustaining 120 days does not necessarily imply success.
% 导言区确保包含：
% \usepackage{graphicx, booktabs, multirow}

\begin{table}[t]
  \centering
  \caption{In-Context Learning Performance Comparison in \textit{Repo System}}
  \label{tab:in_context_fix}
  \medskip
  \resizebox{\linewidth}{!}{
  \begin{tabular}{lcccc}
    \toprule
    Model & Original Avg@4 & In-Context Avg@4 & Original Pass@4 & In-Context Pass@4 \\
    \midrule
    Gemini 3 Pro Preview & 65.83 & 66.67 & 80.00 & 81.67 \\
    gpt-oss-120b-high    & 18.33 & 17.50 & 33.33 & 30.00 \\
    \bottomrule
  \end{tabular}
  }
  \medskip
  
\end{table}

\begin{figure*}[h]
    \centering
    \includegraphics[width=\linewidth]{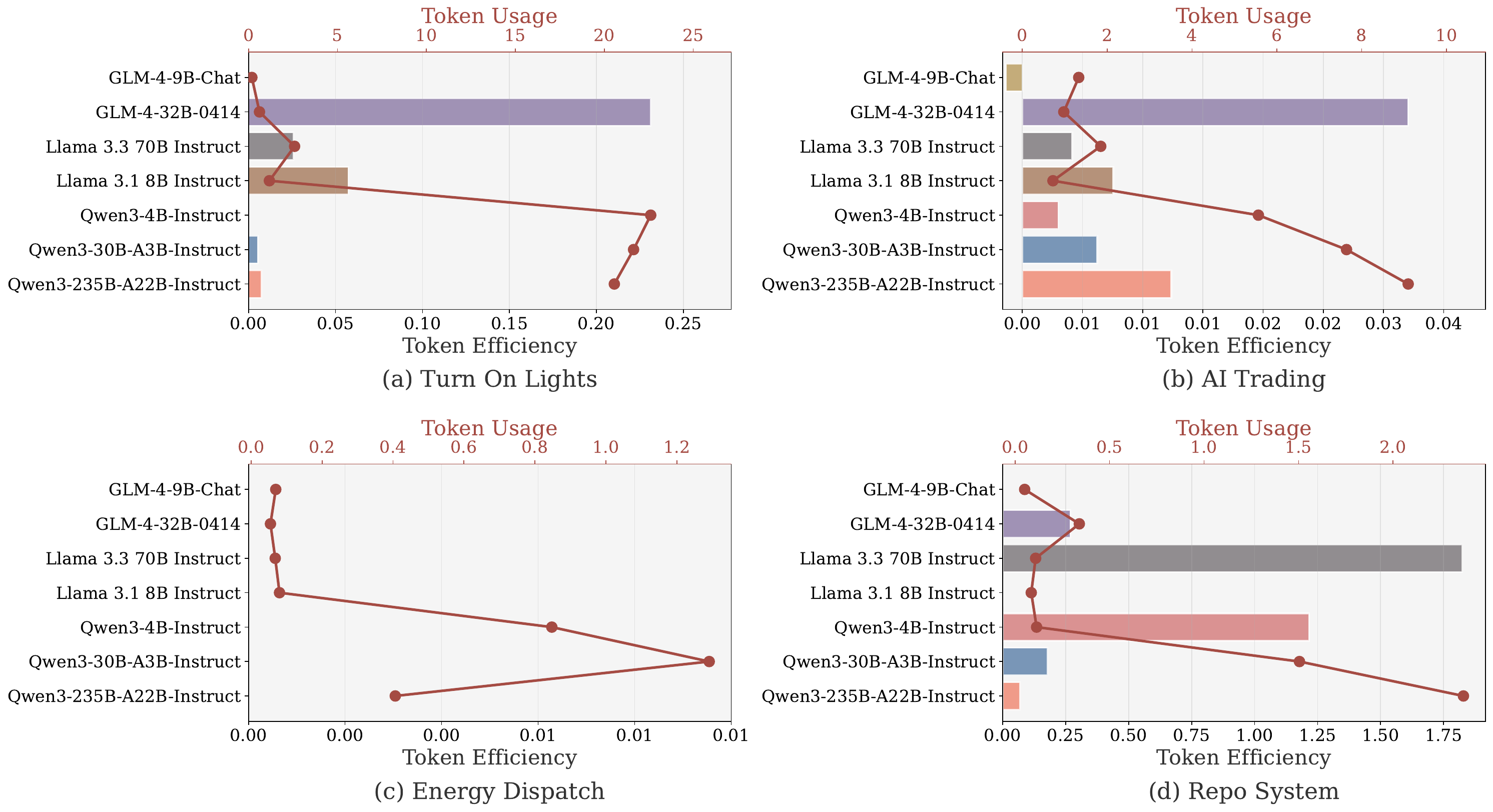}
    \caption{The line chart represents token usage and the bar chart represents token efficiency. Token Usage is measured in units of $10^6$, while Token Efficiency is reported in units of $10^{-6}$.}
    \label{fig:token_efficiency}
\end{figure*}

\subsection{Token efficiency analysis}
\label{app:token_efficiency}
We analyze token efficiency, defined as the contribution of each token to the final success rate (or trading profit), and calculate it as the success rate (or trading profit) divided by the total token usage. We provide detailed analysis of GLM series, Llama series and Qwen series in Figure~\ref{fig:token_efficiency}.

Results in Figure~\ref{fig:token_efficiency} demonstrate that GLM-4-32B-0414 is the most token-efficient in \textit{Turn On Lights} and \textit{AI Trading} environment and Llama 3.3 70B Instruct is the most token-efficient in \textit{Repo System} environment. 

Interestingly, model GLM-4-9B-Chat generates negative profit in \textit{AI Trading} environment, so its token efficiency is negative.
Moreover, models from the three series fail all the tasks in \textit{Energy Dispatch} environment.

Notably, although Qwen achieves best success rate (or trading profit) among the three series, their consistent high token usage makes their token efficiency extremely low.

\subsection{Different strategies for AI trading}
In this subsection, we provide more methods for \textit{AI Trading} environment and analyze their profit. Notably, the strategies in this subsection do not require LLMs and is implemented through only python scripts.

\paragraph{Optimal Strategy.}
We introduce an optimal strategy that operates under the assumption of perfect information regarding future market dynamics. Specifically, with full access to the price trajectory of all stocks, this method adopts a greedy strategy: at each time step, it allocates the entire portfolio to the stock with the highest single-day return ratio. In scenarios where all stocks exhibit a downward trend, keep all assets as cash.

\textbf{Conservative Strategy.}
The fundamental principle of this strategy is to delay trading until sufficient historical data is collected. Specifically, wait and do nothing for at least $\text{num\_factors} + 2$ days. This threshold ensures that the linear system of equations is overdetermined or has a unique solution before any parameter estimation occurs, thereby avoiding instability in the early stages.

We first estimate the dependency matrix $M$, which maps factor changes  to price changes.
We then constructs the price and factor matrices to solve for $M$ using linear algebra solvers and buy/sell stocks based on the hidden rules we find. 
% However, its primary drawback is the opportunity cost incurred by waiting during the initial data accumulation phase.

\paragraph{Progressive Strategy.}
This strategy adopts an aggressive mechanism, starting trading from the third day rather wait for more information.

We employ an incremental learning framework based on the least square method. At any given time step, we utilize the entire history of available price and factor changes accumulated up to that point to estimate the dependency matrix. Unlike approaches that rely on fixed data windows or static thresholds, this strategy dynamically updates its understanding of the market. As new data become available each day, the linear regression model is re-calculated using the full dataset, allowing the estimated parameters to evolve continuously.

This approach fundamentally represents a strategic trade-off between estimation precision and market timing. In the initial stages, the model operates with higher risk because the dataset is sparse, which may lead to significant estimation errors and suboptimal trading decisions. The estimator naturally converges towards the true parameters as more information is gathered. 
% \begin{table}[t]
% \centering
% \caption{Comparison of different strategies, human and models. For LLM models, we provide the best results from both proprietary models and open-source models. $\Delta$ represents the delta of average profit ratio against optimal strategy. \textcolor[HTML]{289BA2}{Colored Rows} represent proprietary models.}
% \label{tab:profit_comparison}
% \resizebox{\linewidth}{!}{
% \begin{tabular}{lcc}
% \toprule

% Method & Avg. Profit Ratio & $\Delta$ \\
% \midrule

% Optimal Strategy & +211.13\% & -- \\
% Conservative Strategy & +192.23\% & -18.90\% \\
% Progressive Strategy & +197.33\% & -13.80\% \\
% Correlation Strategy & +181.51\% & -29.62\% \\
% Rolling Window Strategy & +197.31\% & -13.82\% \\
% Ridge Regression Strategy & +192.63\% & -18.50\% \\
% \midrule

% Human Annotation & 92.55\% & -118.58\% \\

% \midrule

% \rowcolor{mycellcolor2}
% \modellogo{Figures/gemini.png}
% Gemini 3 Pro Preview & +67.71\% & -143.42\% \\
% \modellogo{Figures/qwen.png}
% Qwen3-235B-A22B-Instruct & +11.26\% & -199.87\% \\
% \bottomrule
% \end{tabular}
% }
% \end{table}
\paragraph{Correlation Strategy.}
While progressive strategy attempt to solve for the entire dependency matrix simultaneously by considering the joint influence of all factors, the correlation strategy operates on the assumption that these factors are statistically independent. 

For every individual pair of stocks and factors, we utilize the full historical dataset available to compute this simple regression coefficient. By iterating through all combinations, the model updates the dependency matrix element by element in each step. 

\paragraph{Rolling Window Strategy.}
The Rolling Window strategy posits that recent market behaviors are more predictive of immediate future trends. Consequently, the strategy limits its estimate scope to a fixed-size trailing window (specifically set to 15 days), ensuring that the model remains sensitive to structural shifts in the market environment.

On any given trading day $t$, we construct the dependency matrix using only  data pairs from the most recent $w$ days through the least squares method. Old data points that fall outside this window are systematically ignored in the calculation.

\paragraph{Ridge Regression Strategy.}
This strategy modifies the core learning logic of the progressive strategy by introducing a regularization mechanism. 

Instead of simply minimizing the prediction error through the least squares method, we apply a strict penalty to the magnitude of the regression coefficients. By artificially shrinking the coefficients, the algorithm effectively suppresses the noise that arises when factors are too similar, preventing any single factor from exerting an unrealistically large influence on the trading decision.

% \begin{table}[t]
% \centering
% \caption{Performance of three LLMs in \textit{Repo System} of \odysseychallenge and \odysseylite. \textcolor[HTML]{289BA2}{Colored Row} represents proprietary models. We additionally provide the performance gap.}
% \label{tab:challenge}
% \resizebox{\linewidth}{!}{
% \begin{tabular}{lccc} % 修改列数为4列 (lccc)
% \toprule
% Model & Lete & Challenge & $\Delta$ \\ % 表头扁平化，增加 Delta
% \midrule
% \rowcolor{mycellcolor2}
% \modellogo{Figures/gemini.png}
% Gemini3 Pro Preview & 65.83 & 10.00 & -55.83 \\ % 增加示例数据
% \modellogo{Figures/qwen.png}
% Qwen3-235B-A22B-Instruct & 15.83 & 0.00 & -15.83 \\
% \modellogo{Figures/qwen.png}
% Qwen3-30B-A3B-Instruct & 26.67 & 0.00 & -26.67 \\
% \bottomrule
% \end{tabular}
% }
% \end{table}

% 导言区需确保包含：
% \usepackage{graphicx, booktabs, xcolor, multirow}
% 假设已定义：\modellogo, \odysseychallenge, \odysseylite, mycellcolor2

\begin{table*}[t]
  \centering
  
  % 左侧表格 (Minipage 1)
  \begin{minipage}{0.48\textwidth}
    \centering
    \caption{Comparison of different strategies, human and models. For LLM models, we provide the best results from both proprietary models and open-source models. $\Delta$ represents the delta of average profit ratio against optimal strategy. \textcolor[HTML]{289BA2}{Colored Rows} represent proprietary models.}
    \label{tab:profit_comparison}
    \resizebox{\linewidth}{!}{
    \begin{tabular}{lcc}
    \toprule
    Method & Avg. Profit Ratio & $\Delta$ \\
    \midrule
    Optimal Strategy & +211.13\% & -- \\
    Conservative Strategy & +192.23\% & -18.90\% \\
    Progressive Strategy & +197.33\% & -13.80\% \\
    Correlation Strategy & +181.51\% & -29.62\% \\
    Rolling Window Strategy & +197.31\% & -13.82\% \\
    Ridge Regression Strategy & +192.63\% & -18.50\% \\
    \midrule
    Human Annotation & 92.55\% & -118.58\% \\
    \midrule
    \rowcolor{mycellcolor2}
    \modellogo{Figures/gemini.png}
    Gemini 3 Pro Preview & +67.71\% & -143.42\% \\
    \modellogo{Figures/qwen.png}
    Qwen3-235B-A22B-Instruct & +11.26\% & -199.87\% \\
    \bottomrule
    \end{tabular}
    } % 结束 resizebox
    
  \end{minipage}
  \hfill % 填充中间空白
  % 右侧表格 (Minipage 2)
  \begin{minipage}{0.48\textwidth}
    \centering
    \caption{Performance of three LLMs in \textit{Repo System} of \odysseychallenge and \odysseylite. \textcolor[HTML]{289BA2}{Colored Row} represents proprietary models. We additionally provide the performance gap.}
    \label{tab:challenge}
    \resizebox{\linewidth}{!}{
    \begin{tabular}{lccc}
    \toprule
    Model & Lete & Challenge & $\Delta$ \\
    \midrule
    \rowcolor{mycellcolor2}
    \modellogo{Figures/gemini.png}
    Gemini3 Pro Preview & 65.83 & 10.00 & -55.83 \\
    \modellogo{Figures/qwen.png}
    Qwen3-235B-A22B-Instruct & 15.83 & 0.00 & -15.83 \\
    \modellogo{Figures/qwen.png}
    Qwen3-30B-A3B-Instruct & 26.67 & 0.00 & -26.67 \\
    \bottomrule
    \end{tabular}
    } % 结束 resizebox
    
  \end{minipage}
\end{table*}

\paragraph{Comparison of Strategies}
We provide results of the above mentioned strategies, human annotation and two SOTA models in Table~\ref{tab:profit_comparison}. We can observe that various strategies achieve different performance. Notably, SOTA models demonstrate a significant performance gap against both human annotation and the strategies proposed in this subsection, indicating a huge space for inductive reasoning ability optimization.

\subsection{\odysseychallenge results}
\label{app:odyssey_challenge_results}
For our proposed extremely long-horizon and complex dataset \odysseychallenge, we test \textit{Repo System} and report the results in Table~\ref{tab:challenge}, indicating that the long-horizon scenario is still the bottleneck of current LLM's inductive reasoning.

%% file: Sections/C_experiment_settings.tex
\section{Main results details}
\label{appendix:exp_setting}

\subsection{Baseline settings}
\label{appendix:baseline}
For Qwen3-4B-Instruct and Qwen3-235B-A22B-Instruct, we use the version Qwen3-4B-Instruct-2507 and Qwen3-235B-A22B-Instruct-2507. For Llama3.1-8B-Instruct and Llama3.3-70B-Instruct, we use the checkpoint updated in February, 2025. For GLM-4-9B-Chat, we use the checkpoint updated in January, 2025. For DeepSeek-V3.2, we use the non-thinking version. The reasoning effort of GPT-5 is medium. We additionally report different reasoning effort of gpt-oss-120b: low, medium, high.

Experiments of open-source models are conducted on NVIDIA H200 GPUs, Intel Platinum 8480+ Processor CPU with 56 cores and 1 TB memory. For all experiments, we set the temperature to 0.6 using vLLM engine~\citep{kwon2023efficient} 0.8.5.post1.

For generation, we use \textit{$<$/action$>$} and \textit{$<$/finish$>$} as stop tokens. We then parse the generated action within each $<$action$><$/action$>$ $<$finish$><$/finish$>$ pair.

For AI Trading environment, we keep the 50 most recent interactions as the memory. For Energy Dispatch environment, we keep the 40 most recent interactions as the memory. For other two environments, we keep all interactions as  memory.

\subsection{Evaluation prompts}
\label{appendix:prompts}

Each environment provides the model with three components: system prompt, history records, and current state observation.

\paragraph{Turn On Lights Environment}

The Lights environment provides the following information:
\begin{itemize}
    \item \textbf{System Prompt}: Concise goal description (light all bulbs), rule explanation, action format (integer index).
    \item \textbf{History}: Interaction history with explicit feedback: \texttt{Action: \{action\}, Feedback: \{feedback\}, State: \{obs\}}.
    \item \textbf{Current State}: Visual representation using symbols.
\end{itemize}

\paragraph{AI Trading Environment}

The Trade environment provides the following state information:
\begin{itemize}
    \item \textbf{System Prompt}: Trading objectives, market dynamics explanation, action format (JSON buy/sell combinations).
    \item \textbf{History}: Interaction history formatted as \texttt{market\_info + Action: + action}, containing stock prices, holdings, cash, total value, and news hints.
    \item \textbf{Current State}: Current day, stock prices and holdings, cash, total value, and next day's news (predictive price change hints).
\end{itemize}

\paragraph{Energy Dispatch Environment}

The Energy environment provides the following state information:
\begin{itemize}
    \item \textbf{System Prompt}: Environment description (four generation types), constraints (demand, budget, stability, carbon), dynamic target thresholds, action format.
    \item \textbf{History}: Interaction history formatted as \texttt{state + Action: + action}, containing daily status metrics, previous generation results, supply/demand balance, and financial status.
    \item \textbf{Current State}: Day number, status indicators (stability, carbon, battery), previous step results, current demand, and the next day's budget.
\end{itemize}

\paragraph{Repo System Environment}

The Repo environment provides the following state information:
\begin{itemize}
    \item \textbf{System Prompt}: Detailed environment description, debugging strategies, error interpretation guidelines, action format (command strings).
    \item \textbf{History}: Interaction history in structured format: \texttt{Feedback:\{result\}\textbackslash n\textbackslash n=== Step \{n\} ===\textbackslash n>>> Command: \{command\}}, separated by double newlines.
    \item \textbf{Current State}: Execution result of the last command (success message or error details).
\end{itemize}

%% file: Sections/D_human_anno.tex
\section{Human annotation details}
\label{app:human_annotation}

In this section, we provide detailed information regarding the human annotation process conducted to evaluate \odysseylite, in accordance with responsible NLP research practices.

\subsection{Instructions given to participants}
\begin{figure}[h]
    \centering
    \includegraphics[width=1\linewidth]{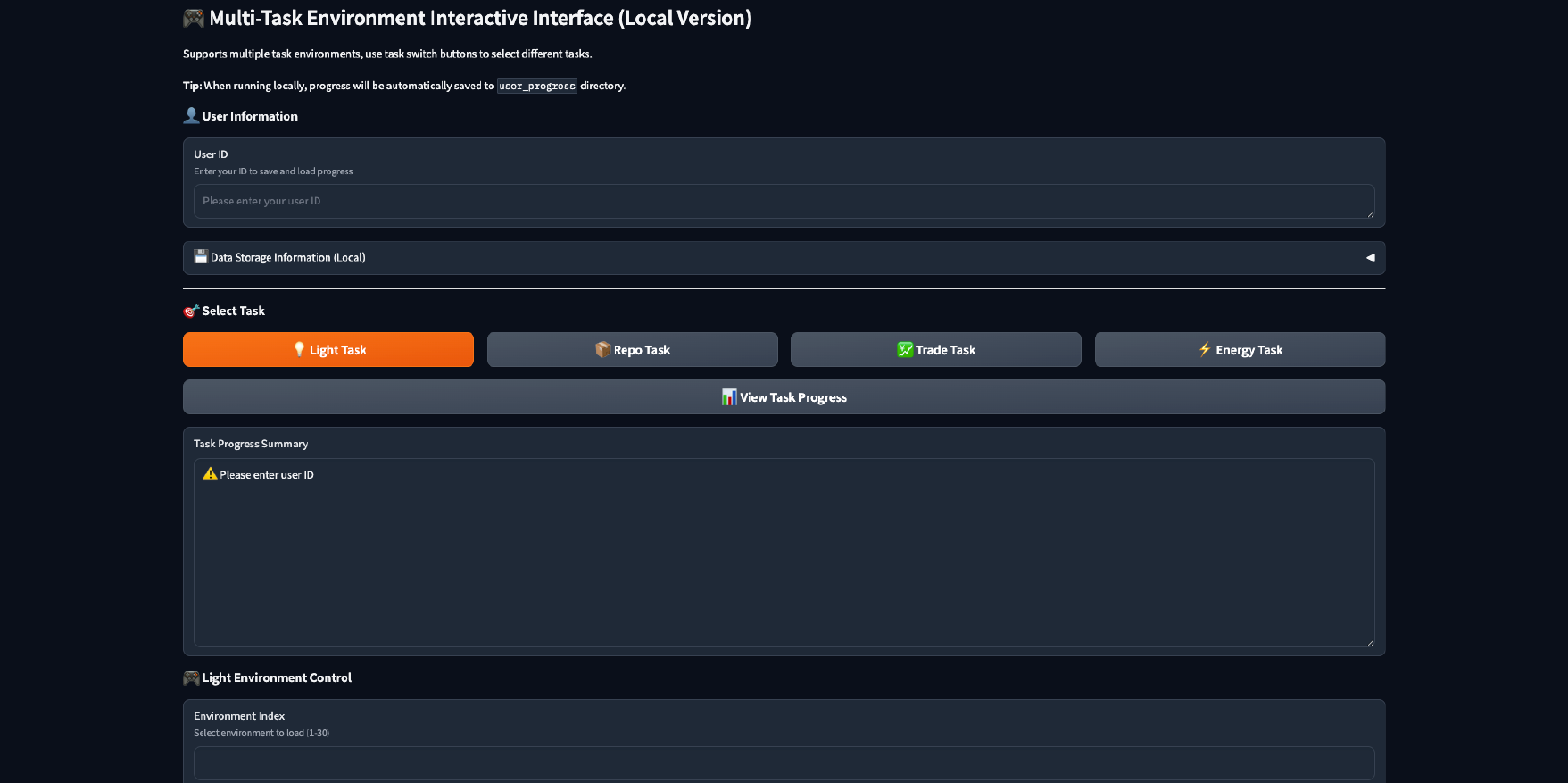}
    \caption{Screenshot of the user interface and instructions provided to the human annotators.}
    \label{fig:annotation_interface}
\end{figure}

%  Reporting instructions and screenshots
We provide annotators with the full text of instructions and a comprehensive guide prior to the task. To facilitate a clear understanding of the evaluation criteria, we designed an intuitive annotation interface. A screenshot of this interface is presented in Figure~\ref{fig:annotation_interface}. Moreover, we provide the user instructions for all environments.

We detail the System Prompts of the four above mentioned environments in the following part.
% \newpage

\begin{figure}[H]
\begin{tcolorbox}[title={Turn On Lights Environment System Prompt}]
You are an intelligent agent.

\vspace{1em}

\#\#\# Goal:

Your mission is to light on all the bulbs.
However, the accessibility of the bulbs is based on the current condition of other bulbs.
You need to learn the hidden rule behind the environment and complete the task.

\vspace{1em}

\#\#\# Action Space:

The action space is based on the index of bulbs. For example, you would like to light on / off the first bulb, you should \
output $<$action$>$0$<$/action$>$ to toggle the state of the bulb. 

\vspace{1em}

\#\#\# History Action and Feedback:

\{history$\_$trajectories\}

\vspace{1em}

\#\#\# Current State:

\{the\_state\_of\_each\_light\}

\vspace{1em}

Now think step by step and choose the next action to act in the environment.
You are encouraged to act actively to derive the environment dynamics.
Output ONLY one action in the format: $<$action$>$n$<$/action$>$
\end{tcolorbox}
\begin{tcolorbox}[title={AI Trading Environment System Prompt}]
\#\#\# Goal:

Your mission is to maximize your total portfolio value by buying and selling stocks.
The market prices are influenced by underlying variables F, and each day's news provides hints about future price changes.
You need to learn the hidden dynamics of the simulated market and make decisions accordingly.
Please note that the underlying meaning of variables may differ from the real stock.

\vspace{1em}

\#\#\# Action Space:

You can take actions in the form of buying or selling multiple stocks each day.
You can combine buy and sell in one action.
The environment will first execute all sell actions, then all buy actions.
You cannot spend more cash than you have or sell stocks you don't own.

**Action Format Examples:**

\quad - To buy 10 shares of S0 and 20 shares of S2, and sell 10 shares of S1:
$<$action$>$\{\{"buy": \{\{"S0": 10, "S2": 20\}\}, "sell": \{\{"S1": 10\}\}\}\}$<$/action$>$

\quad - To only buy:
$<$action$>$\{\{"buy": \{\{"S0": 5\}\}, "sell": \{\{\}\}\}\}$<$/action$>$

\quad - To do nothing:
$<$action$>$\{\{"buy": \{\{\}\}, "sell": \{\{\}\}\}\}$<$/action$>$

**Important:** 

\quad - Stock symbols and numbers should NOT have quotes

\quad - Use valid JSON format inside $<$action$>$$<$/action$>$ tags

\quad - If you cannot afford a purchase or don't own enough shares to sell, that part of the action will be ignored

\vspace{1em}

\#\#\# History Actions and Feedback:

\{history$\_$trajectories\}

\vspace{1em}

\#\#\# Current State:

\{cash\_and\_the\_price\_of\_each\_stock\}

\vspace{1em}

Think carefully step by step and decide your next action.
You are encouraged to act proactively, using the news to predict future price changes,
and to improve your strategy over time.

Provide your action in the format: $<$action$>$...$<$/action$>$

\end{tcolorbox}
\end{figure}

\begin{figure}[H]
\begin{tcolorbox}[title={Energy Dispatch Environment System Prompt (Part I)}]
You are an intelligent energy system operator managing a Dynamic Energy Grid.
Your goal is to achieve a safe, stable, and low-carbon electricity supply across a long planning horizon.
Each day, you adjust the composition of generation resources within strict physical and economic limits.
To perform well, you must learn and exploit hidden temporal patterns from the history.

\vspace{1em}

\# ENVIRONMENT OVERVIEW

This environment simulates a long-horizon national power grid with four generation types:
Thermal — highly reliable, carbon-intensive, lowest cost.
Wind — highly variable, driven by seasonal cycles.
Solar — variable, driven by seasonal cycles.
Battery (Storage) — A storage buffer that can charge or discharge based on the capacity. Its carbon footprint is determined by the source of energy used for charging.

Each day t, the system evolves according to underlying temporal dynamics.
The agent must design the next day's rated generation scheme while anticipating these dynamics.

\vspace{1em}

\#\# Demand \& Budget

The allocation scheme must strictly satisfy both demand and budget constraints.
current$\_$demand (MW) — electricity required today.
current$\_$budget — tomorrow's maximum allowable total generation cost.

\vspace{1em}

\#\# Generation Cost Model (Unit Prices)

Each generation type has a fixed unit cost per MW of rated output:

Thermal: cheapest (e.g., 3.0 cost/unit)

Wind: moderate cost (e.g., 5.0 cost/unit)

Solar: more expensive (e.g., 6.0 cost/unit)

Battery: operational cost (Charge/Discharge), very low (e.g., 0.1 cost/unit)

\vspace{1em}

\#\# Grid Stability

To maintain a stable grid, the agent must avoid large day-to-day changes in the rated outputs.
Sudden increases or decreases (ramping) reduce stability, which affects overall performance.

A good strategy adjusts gradually, anticipating future needs rather than reacting abruptly.

Violating the daily budget or failing to meet the demand would largely damage system stability.

\vspace{1em}

\#\# Carbon Intensity

Thermal generation emits carbon.
To maintain a clean and sustainable city, the agent should limit the proportion of thermal power while still meeting demand and respecting budget constraints.
This creates a non-trivial trade-off between cost, stability, and carbon impact.

\vspace{1em}

\#\# Season \& Efficiency

Actual generation is not equal to rated generation.
It depends on a time-varying efficiency term:

actual$\_$output $=$ rated$\_$output × efficiency (t)

Efficiency changes periodically over time. Solar and Wind share different periods. Agent is required to derive the hidden temporal rules from the history observation.
Because actual output fluctuates around rated output, the agent must leave safety margins and learn the temporal structure.

\vspace{1em}

\# Objective

The agent needs to simulate across a long planning horizon (120 Turns).
The task is successful only if the final metric **Stability $>$ \{target$\_$stability\}, Carbon $<$ \{target$\_$carbon\} **.
Notably, violation of daily cost or demand constraints for 3 consecutive steps would lead to termination.

\end{tcolorbox}
\end{figure}

\begin{figure}[H]
\begin{tcolorbox}[title={Energy Dispatch Environment System Prompt (Part II)}]

\# Action Space

Each day, the agent must decide the rated generation for the next day within the capacity limit:

Thermal (MW), Rated Power Command, [0,600], Must be non-negative.

Wind (MW), Rated Power Command, [0,350], must be non-negative.

Solar (MW), Rated Power Command, [0,250], must be non-negative.

Battery (MW), Net Flow Command, battery capacity=80, Bidirectional: Negative = Charge (Consumption), Positive = Discharge (Supply).

\vspace{1em}

**Action Format Example 1**:

$<$action$>$\{\{"thermal": 400.0, "wind": 10.0, "solar": 35.0, "battery": -15.0\}\}$<$/action$>$

Interpretation: The agent sets the Rated Power for Thermal, Wind, and Solar to 400 MW, 10 MW, and 35 MW, respectively. Additionally, the agent commands the battery to consume 15 MW from the grid for charging. This 15 MW consumption will be drawn from the total supply available from the three generation units.

\vspace{1em}

**Action Format Example 2**:

$<$action$>$\{\{"thermal": 350.0, "wind": 25.0, "solar": 15.0, "battery": 10.0\}\}$<$/action$>$

Interpretation: The agent sets the Rated Power for Thermal, Wind, and Solar to 350 MW, 25 MW, and 15 MW, respectively. Additionally, the agent commands the battery to supply 10 MW of power to the grid (discharging). This 10 MW is added to the total supply from the three generation units.

\vspace{1em}

\# History Action and Feedback:

\{history$\_$trajectories\}

\vspace{1em}

\# Current State:

\{last\_day\_info\_and\_today\_info\}

\vspace{1em}

**Important Note:** 

- Set Rated Capacity above Actual Demand to save room for the efficiency gap (Rated vs. Actual output) and forecast uncertainty.

- Keep daily cost within the budget and meet the daily demand, violation of either cost and supply for three consecutive steps would lead to immediate, irreversible grid collapse.

- Stability and Carbon are long-term average metric. After 120-turn, stability must be $>$ \{target$\_$stability\}, Carbon must be $<$ \{target$\_$carbon\}.

\vspace{1em}

Now think step by step and choose the next action to act in the environment.
You are encouraged to act actively to derive the environment dynamics.
Output ONLY one action within the tag of $<$action$>$$<$/action$>$.
\end{tcolorbox}
\end{figure}

\begin{figure}[H]
\begin{tcolorbox}[title={Repo Management Environment System Prompt (Part I)}]
You are an intelligent computer-using agent.

\vspace{1em}

\# Environment Overview

You are interacting with a simulated Python project setup environment.
This environment mimics real-world difficulties of configuring a repo:

\quad - Partial information (no full dependency graph)

\quad - Object-level runtime failures (module/symbol/kwarg), not explicit version instructions

\quad - Non-monotonic side-effects: installing one package may upgrade/downgrade other packages

\quad - Hidden rules that may only trigger in specific sub-modules or late-stage scripts

\vspace{1em}
\end{tcolorbox}
\end{figure}

\begin{figure}[H]
\begin{tcolorbox}[title={Repo System Environment System Prompt  (Part II)}]

\# Repo Hierarchy \& Debugging

The repo is hierarchical: it contains multiple runnable scripts under subdirectories.
You can debug incrementally by running sub-scripts (to locate which subsystem fails),
but the final goal is to make the entire project pass.

Use:

\quad - `repo tree` (or `repo ls`) to list available scripts in the repo.

\quad - `python $<$script$\_$path$>$` to run a specific sub-script and fix it step by step.

\quad - `python run.py` to run the whole project (a sequence of entry points). This is the only command that ends the episode with success.

\vspace{1em}

\# Goal

Your ultimate goal is to make:
`python run.py`
execute successfully.

\vspace{1em}

\# Action Space (ONE command per step)

- Install Python:

 \quad  - `pip install python==3.10`

- Install packages:

 \quad  - `pip install pkgX`
  
 \quad  - `pip install pkgX==1.2`  (note: if you output x.y.z, it will be interpreted as x.y)
  
 \quad  - `pip install pkgX$>=$1.1,$<$2.0`

- Uninstall packages:

  \quad - `pip uninstall pkgX`

- Inspect environment:

  \quad - `pip list`

- Inspect repo structure:

 \quad  - `repo tree` / `repo ls`

- Execute scripts:

 \quad  - `python run.py`
 
 \quad  - `python core/smoke.py`  (example; use `repo tree` to discover actual paths)

Other commands (e.g., `--upgrade`) are not supported.

\vspace{1em}

\# How to Interpret Errors (Important)

Errors are meant as clues without directly stating version ranges:

\quad - `ModuleNotFoundError: No module named pkgX` usually means pkgX is missing.

\quad - `ImportError: cannot import name 'S' from pkgX.mod` often means pkgX version does not export that symbol.

\quad - `TypeError: ... got an unexpected keyword argument kw` indicates signature/API mismatch.
  If the message says "during project entry", adjust the provider package used by the project.
  If it says "while importing caller$\_$pkg", it indicates a caller-$>$provider incompatibility.

Because installations can trigger side effects, a later fix may break an earlier sub-script.
Use sub-scripts to localize failures, but always re-run `python run.py` to confirm global consistency.

\vspace{1em}

\# History Action and Feedback:

\{history$\_$trajectories\}

\vspace{1em}

\# Current Environment Feedback:

\{specific\_feed\_back\_of\_command\}

\vspace{1em}

Now think step by step and choose the next action.
Output exactly ONE action inside $<$action$>$$<$/action$>$, e.g. $<$action$>$pip install pkg0==2.1$<$/action$>$.
\end{tcolorbox}
\end{figure}

\begin{figure}[H]
\begin{tcolorbox}
[title={Turn On Lights Environment User Instruction}]
Assume there are 3 bulbs (indices 0, 1, 2), all initially off.

\textbf{Example Logic (only shown in examples, these rules are hidden in actual tasks, users need to infer)}:
\begin{itemize}
    \item B0: True  \# Represents B0 can be turned on under any circumstances
    \item B1: B0  \# Represents B1 can only be turned on when B0 is on
    \item B2: not B1 and B0  \# Represents B2 can only be turned on when B1 is off and B0 is on
\end{itemize}

\textbf{Example Steps}:
\begin{enumerate}
    \item \textbf{Step 1}: Input action \texttt{1}, click ``Execute Action''
    \begin{itemize}
        \item Environment state after execution: $\circ$ $\circ$ $\circ$
        \item Environment feedback: B1 remains inactive... remaining bulbs should be in specific mode.
        \item Reason: B1 can only be turned on when B0 is on, but B0 is off, so B1 cannot be turned on.
    \end{itemize}

    \item \textbf{Step 2}: Input action \texttt{0}, click ``Execute Action''
    \begin{itemize}
        \item Environment state after execution: $\bullet$ $\circ$ $\circ$
        \item Environment feedback: Toggled B1 to True
        \item Reason: B0 can be turned on at any time.
    \end{itemize}

    \item \textbf{Step 3}: Input action \texttt{2}, click ``Execute Action''
    \begin{itemize}
        \item Environment state after execution: $\bullet$ $\circ$ $\bullet$
        \item Environment feedback: Toggled B2 to True
        \item Reason: B2 can only be turned on when B1 is off and B0 is on, so B2 was turned on.
    \end{itemize}

    \item \textbf{Step 4}: Input action \texttt{1}, click ``Execute Action''
    \begin{itemize}
        \item Environment state after execution: $\bullet$ $\bullet$ $\bullet$ \text{(Task completed)}
        \item Environment feedback: Toggled B1 to True
        \item Reason: B1 can only be turned on when B0 is on, so B1 was turned on.
    \end{itemize}
\end{enumerate}

\textbf{Tips}:
\begin{itemize}
    \item $\bullet$ indicates bulb is lit
    \item $\circ$ indicates bulb is not lit
    \item Each bulb's availability may depend on other bulbs' states
    \item You need to discover hidden rules through experimentation
    \item Maximum 200 steps allowed
\end{itemize}

\textbf{Goal}: Light all bulbs (all bulbs display as $\bullet$)
\end{tcolorbox}
\end{figure}

\begin{figure}[H]
\begin{tcolorbox}[title={AI Trading Environment User Instruction (Part I)}]
\textbf{Scenario Description} \\
You are a stock trader who needs to perform buy and sell operations across multiple trading days, achieving maximum returns within 120 days.

\vspace{1em}

\textbf{Important Concepts}:
\begin{itemize}
    \item \textbf{S0, S1}: Stock codes (Stocks), representing 2 different stocks that can be bought and sold
    \item \textbf{F0, F1}: Market factors (Factors), representing market factors that affect stock prices
    \begin{itemize}
        \item News will report changes in these factors (e.g., ``F0 rose slightly (+0.03)'')
        \item Factor changes affect stock prices through dependency matrix
        \item You need to predict stock price changes based on news, then trade
    \end{itemize}
    \item Please check news, e.g., ``F0 rose slightly (+0.03) | F1 decreased significantly (-0.10)'' predict which stocks will rise/fall based on factor changes
    \item Buying is limited by cash
    \item Selling is limited by holdings
\end{itemize}

\textbf{Available Actions}:
\begin{itemize}
    \item \textbf{Buy Stock}: Input positive number to buy (e.g., S0 input 100 means buy 100 shares of S0)
    \item \textbf{Sell Stock}: Input negative number to sell (e.g., S0 input -50 means sell 50 shares of S0)
    \item Buying is limited by cash, selling is limited by holdings
\end{itemize}

\textbf{Example}:

\vspace{1em}

\textbf{Example Logic (only shown in examples, these rules are hidden in actual tasks, users need to infer)}:
\begin{itemize}
    \item Matrix corresponding to S0, S1, F0, F1 is: 
    \[
    \begin{bmatrix} 
    0.1 & 0.2 \\ 
    -0.3 & 0.4 
    \end{bmatrix}
    \]
    \item Represents: 
    \begin{itemize}
        \item F0 rises 1 point, S0 rises 0.1 points; F0 rises 1 point, S1 falls 0.3 points
        \item F1 rises 1 point, S0 rises 0.2 points; F1 rises 1 point, S1 rises 0.4 points
    \end{itemize}
\end{itemize}

\textbf{Initial Environment in This Example}:
\begin{itemize}
    \item You have 100 cash
    \item S0 initial price is 1, S1 initial price is 2
    \item This example is a simple demonstration, only keeps 2 days (actual task is 120 days)
\end{itemize}

\end{tcolorbox}
\end{figure}

\begin{figure}
\begin{tcolorbox}
[title={AI Trading Environment User Instruction (Part II)}]
\textbf{Example Steps}:

\textbf{Step 1 (Day 1)}:  

\begin{itemize}
    \item Environment state before execution: Tomorrow F0 rose significantly (+0.10) | F1 rose slightly (+0.05)
    \item Stock prices before execution: S0 1.00, S1 2.00, Cash 100
    \item Action: Buy 100 shares of S0
    \item Reason: S0 tomorrow's price = 1.00 + (0.1×0.10) + (0.2×0.05) = 1.00 + 0.01 + 0.01 = 1.02 (up 2\%), while S1 tomorrow's price is S1 = 2.00 + ((-0.3)×0.10) + (0.4×0.05) = 2.00 - 0.03 + 0.02 = 1.99 (down 0.5\%). S0 rises while S1 falls, so buy S0. Buying 100 shares of S0 costs 100, cash becomes 0.
\end{itemize}

\textbf{Step 2 (Day 2)}:  
\begin{itemize}
    \item Environment state before execution: Tomorrow F0 decreased significantly (-0.15) | F1 rose significantly (+0.10)
    \item Stock prices before execution: S0 1.02, S1 1.99, Cash 0, Holdings 100 shares of S0
    \item Sell 100 shares of S0, buy about 51 shares of S1
    \item Reason: S0 tomorrow's price = 1.02 + (0.1×(-0.15)) + (0.2×0.10) = 1.02 - 0.015 + 0.02 = 1.025 (slight rise 0.5\%), while S1 tomorrow's price is S1 = 1.99 + ((-0.3)×(-0.15)) + (0.4×0.10) = 1.99 + 0.045 + 0.04 = 2.075 (up 4.3\%). S1 rise is much greater than S0, so sell S0 and buy S1. Selling 100 shares of S0 gets 102, can buy about 51 shares of S1 (102/1.99 = 51.26, rounded to 51 shares, costs about 101.49).
\end{itemize}

\textbf{Step 3 (Day 3)}:  
\begin{itemize}
    \item Environment state before execution: Tomorrow F0 stable (0.00) | F1 rose significantly (+0.20)
    \item Stock prices before execution: S0 1.025, S1 2.075, Cash 0.51, Holding 51 shares of S1
    \item Action: No operation (or use remaining cash to buy small amount of S1)
    \item Reason: S0 tomorrow's price = 1.025 + (0.1×0) + (0.2×0.20) = 1.025 + 0.04 = 1.065 (up 3.9\%), while S1 tomorrow's price is S1 = 2.075 + ((-0.3)×0) + (0.4×0.20) = 2.075 + 0.08 = 2.155 (up 3.9\%). Both stocks have similar rises, but S1 absolute rise is larger (0.08 vs 0.04), and already holding S1, so maintain position.
\end{itemize}

\vspace{1em}

\textbf{Final State}: 51 shares of S1, price 2.155 per share, total value about 109.91 (51×2.155), plus remaining cash about 0.51, total value about 110.42, return rate about 10.42\%
\end{tcolorbox}
\end{figure}

\begin{figure}[H]
\begin{tcolorbox}[title={Energy Dispatch Environment User Instruction (Part I)}]
\textbf{Scenario Description} \\
You need to manage an energy grid, balancing generation, demand, and budget while meeting stability and carbon emission targets, completing at least 120 days of tasks. If demand violations or budget violations occur for three consecutive days, the task will fail immediately.

\textbf{Task Objectives}:
\begin{itemize}
    \item \textbf{Completion Days}: Complete at least 120 days
    \item \textbf{Stability Target}: Final average stability must be $\ge$ target value (shown in status)
    \item \textbf{Carbon Emission Target}: Final carbon emission ratio must be $\le$ target value (shown in status)
    \item \textbf{Violation Limit}: 3 consecutive days of demand violations or budget violations will cause task failure
\end{itemize}

\textbf{Available Actions}:
\begin{itemize}
    \item \textbf{Thermal}: Input thermal power generation ($\ge$0)
    \item \textbf{Wind}: Input wind power generation ($\ge$0)
    \item \textbf{Solar}: Input solar power generation ($\ge$0)
    \item \textbf{Battery}: Input battery operation
    \begin{itemize}
        \item Negative value = charging (e.g., -20)
        \item Positive value = discharging (e.g., 20)
        \item 0 = no battery usage
        \item Battery has maximum capacity limit of 80
    \end{itemize}
\end{itemize}

\textbf{Actual Generation Calculation}:
\begin{itemize}
    \item Actual generation = Input generation $\times$ efficiency coefficient
    \item After actual generation, store to battery, this stage has no loss
    \item For example, input thermal 10, wind 20, solar 30, battery storage 10. Thermal efficiency 0.9, wind efficiency 1.1, solar efficiency 1
    \item Then actual generation: $10 \times 0.9 + 20 \times 1.1 + 30 \times 1 = 61$
    \item Applied to grid (subtract battery storage): $61 - 10 = 51$
\end{itemize}

\textbf{Stability Requirements}:
\begin{itemize}
    \item Daily generation configuration changes cannot be too large, otherwise it will cause grid instability
    \item Stability calculation considers: generation configuration change magnitude (ramping), budget violations, demand violations
    \item If budget violation or demand violation occurs, stability will decrease significantly
    \item \textbf{Important}: Insufficient stability will not directly terminate the task, but will be used to judge task success after completion. So you need to adjust strategy timely to improve stability
\end{itemize}

\end{tcolorbox}
\end{figure}

\begin{figure}
\begin{tcolorbox}
[title={Energy Dispatch Environment User Instruction (Part II)}]
\textbf{Carbon Emission Requirements}:
\begin{itemize}
    \item Carbon emission ratio = Historical cumulative thermal actual generation / Historical cumulative total actual generation
    \item When task is completed, carbon emission ratio must be $\le$ target value
    \item Need to control thermal power proportion of total generation throughout the task
    \item \textbf{Important}: Excessive carbon emissions will not directly terminate the task, but will be used to judge task success after completion. So you need to adjust strategy timely to reduce carbon emissions
\end{itemize}

\textbf{Violation Explanation}:
\begin{itemize}
    \item \textbf{Demand Violation}: Actual supply $<$ demand
    \item \textbf{Budget Violation}: Actual cost $>$ budget
    \item Insufficient stability or excessive carbon emissions do not count as violations
    \item Three consecutive days of violations will directly terminate and fail the task
    \item \textbf{Important}: Only demand violations and budget violations will increase consecutive violation days. Insufficient stability and excessive carbon emissions are not violations but affect final results
\end{itemize}

\textbf{Example}:

\textbf{Scenario Description}:
\begin{itemize}
    \item Thermal, wind, solar unit prices are 2, 4, 6 per unit respectively, battery operation cost 0.1 yuan/unit
    \item Carbon emission ratio target $\le$ 0.81 (i.e., thermal proportion $\le$ 0.19)
    \item Stability target $\ge$ 0.5
    \item This example demonstrates 6 days, actual task requires completing 120 days
\end{itemize}

\textbf{Example Logic (only shown in examples, these rules are hidden in actual tasks, users need to infer)}:
\begin{itemize}
    \item Thermal efficiency sequence: [1.0, 1.0, 1.0, 0.9, 1.1, 1.0] (randomly fluctuates around 1)
    \item Wind efficiency sequence: [1.1, 1.0, 1.1, 1.0, 1.1, 1.0] (cycle every 2 days)
    \item Solar efficiency sequence: [0.9, 1.0, 1.1, 0.9, 1.0, 1.1] (cycle every 3 days)
\end{itemize}

\textbf{Important Notes}:
\begin{itemize}
    \item In actual tasks, efficiency coefficients are hidden and need to be inferred from historical data
    \item Need to balance cost, stability, carbon emissions, and demand satisfaction
    \item Insufficient stability and excessive carbon emissions will not directly terminate the task but will affect final task completion conditions
    \item Only demand violations and budget violations will increase consecutive violation days, 3 consecutive days of violations will cause task failure
    \item When violations occur, need to adjust strategy timely to avoid consecutive violations
    \item In actual problems, you cannot see the specific calculation process of stability coefficient, you only see a result, please adjust strategy based on this result
\end{itemize}

\end{tcolorbox}
\end{figure}

\begin{figure}[H]
\begin{tcolorbox}[title={Repo System Environment User Instruction (Part I)}]
\textbf{Scenario Description} \\
You need to configure Python environment and install correct package versions so that the project can run successfully: \texttt{python run.py}

\textbf{Available Commands}:
\begin{itemize}
    \item \texttt{pip install python==3.10} - Install Python version
    \item \texttt{pip install pkg0==1.2} - Install package (supports version constraints)
    \item \texttt{pip uninstall pkg0} - Uninstall package
    \item \texttt{pip list} - View current environment status
    \item \texttt{repo tree} - View repository structure
    \item \texttt{python run.py} - Run project (success means task completed)
\end{itemize}

\textbf{Example Hidden Rules (users need to discover in actual tasks)}:
\begin{itemize}
    \item Requires \texttt{python>=3.10}
    \item Requires \texttt{pkg1==1.0}
    \item Requires \texttt{pkg2>=1.2,<=2.0}
    \item Requires \texttt{pkg3<=1.0}
    \item All version numbers of \texttt{pkg3} must match \texttt{pkg1} (including integer and decimal parts)
    \item Major version number of \texttt{pkg2} must match \texttt{pkg1} (integer part)
\end{itemize}

\textbf{Example Steps}:

\textbf{Step 1}: Input \texttt{pip install python==3.10}, click ``Execute Action'' \\
Environment feedback: Successfully installed \texttt{python==3.10} \\
Reason: Successfully installed

\vspace{1em}

\textbf{Step 2}: Input \texttt{python run.py}, click ``Execute Action'' \\
Environment feedback: \texttt{ModuleNotFoundError: No module named 'pkg1'.} \\
Reason: \texttt{pkg1} not installed

\vspace{1em}

\textbf{Step 3}: Input \texttt{pip install pkg1==1.0}, click ``Execute Action'' \\
Environment feedback: Successfully installed \texttt{pkg1==1.0} \\
Reason: Successfully installed \texttt{pkg1==1.0}

\vspace{1em}

\textbf{Step 4}: Input \texttt{python run.py}, click ``Execute Action'' \\
Environment feedback: \texttt{ModuleNotFoundError: No module named 'pkg2'.} \\
Reason: \texttt{pkg2} not installed

\vspace{1em}

\textbf{Step 5}: Input \texttt{pip install pkg2==2.0}, click ``Execute Action'' \\
Environment feedback: Successfully installed \texttt{pkg2==2.0} \\
Reason: Successfully installed \texttt{pkg2==2.0}

\vspace{1em}

\textbf{Step 6}: Input \texttt{python run.py}, click ``Execute Action'' \\
Environment feedback: \texttt{RuntimeError: ABI mismatch detected between 'pkg6' and dependent packages.} \\
Reason: Major version number of \texttt{pkg2} does not match \texttt{pkg1}

\end{tcolorbox}
\end{figure}

\begin{figure}
\begin{tcolorbox}[title={Repo Syetem Environment User Instruction (Part II)}]

\textbf{Step 7}: Input \texttt{pip install pkg2==1.0}, click ``Execute Action'' \\
Environment feedback: Successfully installed \texttt{pkg3==1.0} \\
Reason: Successfully installed \texttt{pkg3==1.2}

\vspace{1em}

\textbf{Step 8}: Input \texttt{python run.py}, click ``Execute Action'' \\
Environment feedback: \texttt{ModuleNotFoundError: No module named 'pkg3'.} \\
Reason: \texttt{pkg2} not installed

\vspace{1em}

\textbf{Step 9}: Input \texttt{pip install pkg3==1.0}, click ``Execute Action'' \\
Environment feedback: Successfully installed \texttt{pkg3==1.0} \\
Reason: Successfully installed \texttt{pkg3==0.1}

\vspace{1em}

\textbf{Step 10}: Input \texttt{python run.py}, click ``Execute Action'' \\
Environment feedback: \texttt{RuntimeError: tightly-coupled components are out of sync with 'pkg1'.} \\
Reason: All version numbers of \texttt{pkg3} must match \texttt{pkg1} (including integer and decimal parts)

\vspace{1em}

\textbf{Step 11}: Input \texttt{pip install pkg3==1.0}, click ``Execute Action'' \\
Environment feedback: Successfully installed \texttt{pkg3==1.0} \\
Reason: Successfully installed \texttt{pkg3==1.0}

\vspace{1em}

\textbf{Step 12}: Input \texttt{python run.py}, click ``Execute Action'' \\
Environment feedback:  Task completed! Project ran successfully! \\
Reason: All conditions met

\textbf{Tips}:
\begin{itemize}
    \item Packages may have dependencies and version conflicts
    \item Need to carefully handle version constraints
    \item Maximum 120 steps allowed
\end{itemize}

\textbf{Goal}:
Successfully run \texttt{python run.py} so that the project can execute normally
\end{tcolorbox}
\end{figure}

\subsection{Recruitment and Compensation}
% 对应 D2 & D4: Recruitment, Payment, and IRB status
% The annotation task was conducted internally by the authors of this paper. No external annotators are recruited. 
This approach ensures a high level of expertise, as all recruited annotators are students with a background in Artificial Intelligence and possess the necessary domain knowledge to accurately assess model performance against the specific criteria defined for \odyssey. Annotators are compensated at a rate of about \$ 15 per hour.

% \paragraph{Compensation and Ethics.}
% Since the annotation was performed by the research team as part of their contribution to the project, no compensation was provided.

\subsection{Consent}
% 对应 D3: Consent
Informed consent was obtained from all annotators, with the mutual understanding that the collected data would be utilized for research purposes and released as a public dataset.

\subsection{Annotation Process}
To mitigate individual bias, each example in our dataset was annotated by four independent annotators. The annotation workflow consisted of two stages:
\begin{enumerate}
    \item \textbf{One-shot Demonstration:} Participants were first presented with simplified examples for each environment. These examples served as tutorials and did not overlap with the test dataset.
    \item \textbf{Main Annotation:} Annotators labeled the generated outputs strictly adhering to the specific rules defined within each environment.
\end{enumerate}

\begin{table}[t]
\centering
\caption{Inter-rater reliability analysis across four experimental environments. For tasks with discrete success metrics (\textit{Turn On Lights}, \textit{Energy Dispatch}, and \textit{Repo System}), we report Fleiss' Kappa ($\kappa$). For the continuous reasoning task (\textit{AI Trading}), the Intraclass Correlation Coefficient (ICC) is employed.}
\label{tab:agreement_analysis}
\medskip
\resizebox{\linewidth}{!}{
\begin{tabular}{lccc} % 列格式从 lcc 改为 lccc（增加一列）
\toprule
Environment & Success Rate Type & Metric & Inner-Annotation Score \\ % 插入 Reward Type 表头
\midrule
Turn On Lights & Discrete & Fleiss' $\kappa$ & 0.42 \\
Energy Dispatch & Discrete & Fleiss' $\kappa$ & 0.40 \\
AI Trading & Continuous & Intraclass Correlation Coefficient & 0.12 \\
Repo System & Discrete & Fleiss' $\kappa$ & 0.18 \\
\bottomrule
\end{tabular}
}

\medskip

\end{table}

\subsection{Inter-Annotator Agreement}
% 对应 D5 & Stats:
% To ensure data quality, we monitored the agreement among the four annotators. For \textit{Turn On Lights}, \textit{Energy Dispatch} and \textit{Repo System}, the task can be explicitly evaluated by the success or not, so we apply Fleiss' Kappa coefficient to them. And for \textit{Repo System}, we use the Intraclass Correlation Coefficient (ICC) to assess the agreement aomong annotators. The Fleiss' Kappa coefficients of success or not in \textit{Turn On Lights}, \textit{Energy Dispatch}, \textit{Repo System} are 0.42, 0.40, 0.18, respectively. These moderate coefficients validate our data quality. Additionally,  ICC in \textit{AI Trading} is 0.12, demonstrating that annotators have explicitly different inductive reasoning strategies in this environment. 

To ensure the reliability of the human annotated data, we conducted inner agreement analysis across four experimental environments. For tasks with discrete outcomes (\textit{Turn On Lights}, \textit{Energy Dispatch}, and \textit{Repo System}), where performance can be explicitly judged as a binary success metric, Fleiss' Kappa was employed to account for multiple annotators. The resulting coefficients for these environments indicate a moderate level of consensus consistent. Furthermore, for the \textit{AI Trading} environment, which yields continuous performance metrics (profit rate), we use the Intraclass Correlation Coefficient (ICC). And the ICC score quantitatively confirm that annotators employed significantly divergent inductive reasoning strategies in this environment.

%% file: neurips_2026.bbl
\begin{thebibliography}{40}
\providecommand{\natexlab}[1]{#1}
\providecommand{\url}[1]{\texttt{#1}}
\expandafter\ifx\csname urlstyle\endcsname\relax
  \providecommand{\doi}[1]{doi: #1}\else
  \providecommand{\doi}{doi: \begingroup \urlstyle{rm}\Url}\fi

\bibitem[Akata et~al.(2025)Akata, Schulz, Coda-Forno, Oh, Bethge, and
  Schulz]{Akata_2025}
Elif Akata, Lion Schulz, Julian Coda-Forno, Seong~Joon Oh, Matthias Bethge, and
  Eric Schulz.
\newblock Playing repeated games with large language models.
\newblock \emph{Nature Human Behaviour}, 9\penalty0 (7):\penalty0 1380–1390,
  May 2025.
\newblock ISSN 2397-3374.
\newblock \doi{10.1038/s41562-025-02172-y}.
\newblock URL \url{http://dx.doi.org/10.1038/s41562-025-02172-y}.

\bibitem[Anthropic\;AI(2024)]{anthropic2024claude}
Anthropic\;AI.
\newblock The claude 3 model family: Opus, sonnet, haiku.
\newblock \emph{Claude-3 Model Card}, 1:\penalty0 1, 2024.
\newblock URL
  \url{https://assets.anthropic.com/m/61e7d27f8c8f5919/original/Claude-3-Model-Card.pdf}.

\bibitem[Chevalier-Boisvert et~al.(2019)Chevalier-Boisvert, Bahdanau, Lahlou,
  Willems, Saharia, Nguyen, and Bengio]{chevalierbabyai}
Maxime Chevalier-Boisvert, Dzmitry Bahdanau, Salem Lahlou, Lucas Willems,
  Chitwan Saharia, Thien~Huu Nguyen, and Yoshua Bengio.
\newblock Babyai: A platform to study the sample efficiency of grounded
  language learning.
\newblock In \emph{International Conference on Learning Representations}, 2019.
\newblock URL \url{https://iclr.cc/virtual/2019/poster/733}.

\bibitem[Chollet(2019)]{chollet2019measure}
Fran{\c{c}}ois Chollet.
\newblock On the measure of intelligence.
\newblock \emph{arXiv preprint arXiv:1911.01547}, 2019.
\newblock URL \url{https://arxiv.org/abs/1911.01547}.

\bibitem[Chung et~al.(2025)Chung, Zhang, Lin, Rawal, Gao, and
  Chai]{chung2025evaluatinglongcontextreasoningllmbased}
Andy Chung, Yichi Zhang, Kaixiang Lin, Aditya Rawal, Qiaozi Gao, and Joyce
  Chai.
\newblock Evaluating long-context reasoning in llm-based webagents.
\newblock In \emph{NeurIPS 2025 Workshop on Bridging Language, Agent, and World
  Models for Reasoning and Planning}, 2025.
\newblock URL \url{https://openreview.net/forum?id=oxj422wRvO}.

\bibitem[Clark(1997)]{clark1997dynamical}
Andy Clark.
\newblock The dynamical challenge.
\newblock \emph{Cognitive science}, 21\penalty0 (4):\penalty0 461--481, 1997.
\newblock URL
  \url{https://www.sciencedirect.com/science/article/abs/pii/S0364021399800305}.

\bibitem[Deng et~al.(2023)Deng, Gu, Zheng, Chen, Stevens, Wang, Sun, and
  Su]{deng2023mind2web}
Xiang Deng, Yu~Gu, Boyuan Zheng, Shijie Chen, Sam Stevens, Boshi Wang, Huan
  Sun, and Yu~Su.
\newblock Mind2web: Towards a generalist agent for the web.
\newblock \emph{Advances in Neural Information Processing Systems},
  36:\penalty0 28091--28114, 2023.
\newblock URL \url{https://openreview.net/forum?id=kiYqbO3wqw}.

\bibitem[{Gemini Team}(2025)]{comanici2025gemini}
{Gemini Team}.
\newblock Gemini 2.5: Pushing the frontier with advanced reasoning,
  multimodality, long context, and next generation agentic capabilities.
\newblock \emph{arXiv preprint arXiv:2507.06261}, 2025.
\newblock URL \url{https://arxiv.org/abs/2507.06261}.

\bibitem[GLM et~al.(2024)GLM, Zeng, Xu, Wang, Zhang, Yin, Zhang, Rojas, Feng,
  Zhao, et~al.]{glm2024chatglm}
Team GLM, Aohan Zeng, Bin Xu, Bowen Wang, Chenhui Zhang, Da~Yin, Dan Zhang,
  Diego Rojas, Guanyu Feng, Hanlin Zhao, et~al.
\newblock Chatglm: A family of large language models from glm-130b to glm-4 all
  tools.
\newblock \emph{arXiv preprint arXiv:2406.12793}, 2024.
\newblock URL \url{https://arxiv.org/abs/2406.12793}.

\bibitem[Grattafiori et~al.(2024)Grattafiori, Dubey, Jauhri, Pandey, Kadian,
  Al-Dahle, Letman, Mathur, Schelten, Vaughan, et~al.]{grattafiori2024llama}
Aaron Grattafiori, Abhimanyu Dubey, Abhinav Jauhri, Abhinav Pandey, Abhishek
  Kadian, Ahmad Al-Dahle, Aiesha Letman, Akhil Mathur, Alan Schelten, Alex
  Vaughan, et~al.
\newblock The llama 3 herd of models.
\newblock \emph{arXiv preprint arXiv:2407.21783}, 2024.
\newblock URL \url{https://arxiv.org/abs/2407.21783}.

\bibitem[Ha and Schmidhuber(2018)]{ha2018recurrent}
David Ha and J{\"u}rgen Schmidhuber.
\newblock Recurrent world models facilitate policy evolution.
\newblock \emph{Advances in neural information processing systems}, 31, 2018.
\newblock URL
  \url{https://proceedings.neurips.cc/paper_files/paper/2018/file/2de5d16682c3c35007e4e92982f1a2ba-Paper.pdf}.

\bibitem[Kwon et~al.(2023)Kwon, Li, Zhuang, Sheng, Zheng, Yu, Gonzalez, Zhang,
  and Stoica]{kwon2023efficient}
Woosuk Kwon, Zhuohan Li, Siyuan Zhuang, Ying Sheng, Lianmin Zheng, Cody~Hao Yu,
  Joseph~E. Gonzalez, Hao Zhang, and Ion Stoica.
\newblock Efficient memory management for large language model serving with
  pagedattention.
\newblock In \emph{Proceedings of the ACM SIGOPS 29th Symposium on Operating
  Systems Principles}, 2023.
\newblock URL \url{https://dl.acm.org/doi/abs/10.1145/3600006.3613165}.

\bibitem[Lake et~al.(2017)Lake, Ullman, Tenenbaum, and
  Gershman]{lake2017building}
Brenden~M Lake, Tomer~D Ullman, Joshua~B Tenenbaum, and Samuel~J Gershman.
\newblock Building machines that learn and think like people.
\newblock \emph{Behavioral and brain sciences}, 40:\penalty0 e253, 2017.
\newblock URL \url{https://pubmed.ncbi.nlm.nih.gov/27881212/}.

\bibitem[Li et~al.(2024)Li, Wu, Jin, Ma, Chen, and Zheng]{li2024state}
Jintang Li, Ruofan Wu, Xinzhou Jin, Boqun Ma, Liang Chen, and Zibin Zheng.
\newblock State space models on temporal graphs: A first-principles study.
\newblock \emph{Advances in Neural Information Processing Systems},
  37:\penalty0 127030--127058, 2024.
\newblock URL \url{https://openreview.net/forum?id=UaJErAOssN}.

\bibitem[Lin et~al.(2025)Lin, Le~Bras, Richardson, Sabharwal, Poovendran,
  Clark, and Choi]{linzebralogic}
Bill~Yuchen Lin, Ronan Le~Bras, Kyle Richardson, Ashish Sabharwal, Radha
  Poovendran, Peter Clark, and Yejin Choi.
\newblock Zebralogic: On the scaling limits of llms for logical reasoning.
\newblock In \emph{Forty-second International Conference on Machine Learning},
  2025.
\newblock URL \url{https://openreview.net/forum?id=sTAJ9QyA6l}.

\bibitem[Liu et~al.(2025)Liu, Mei, Lin, Xue, Wang, Xu, Wu, Zhang, Lin, Dong,
  et~al.]{liu2025deepseek}
Aixin Liu, Aoxue Mei, Bangcai Lin, Bing Xue, Bingxuan Wang, Bingzheng Xu,
  Bochao Wu, Bowei Zhang, Chaofan Lin, Chen Dong, et~al.
\newblock Deepseek-v3. 2: Pushing the frontier of open large language models.
\newblock \emph{arXiv preprint arXiv:2512.02556}, 2025.
\newblock URL \url{https://arxiv.org/abs/2512.02556}.

\bibitem[Liu et~al.(2024)Liu, Yu, Zhang, Xu, Lei, Lai, Gu, Ding, Men, Yang,
  et~al.]{liu2024agentbench}
Xiao Liu, Hao Yu, Hanchen Zhang, Yifan Xu, Xuanyu Lei, Hanyu Lai, Yu~Gu,
  Hangliang Ding, Kaiwen Men, Kejuan Yang, et~al.
\newblock Agentbench: Evaluating llms as agents.
\newblock In \emph{ICLR}, 2024.
\newblock URL \url{https://openreview.net/forum?id=zAdUB0aCTQ}.

\bibitem[Mialon et~al.(2023)Mialon, Fourrier, Wolf, LeCun, and
  Scialom]{mialon2023gaia}
Gr{\'e}goire Mialon, Cl{\'e}mentine Fourrier, Thomas Wolf, Yann LeCun, and
  Thomas Scialom.
\newblock Gaia: a benchmark for general ai assistants.
\newblock In \emph{The Twelfth International Conference on Learning
  Representations}, 2023.

\bibitem[OpenAI(2025)]{openai2025gptoss}
OpenAI.
\newblock gpt-oss-120b \& gpt-oss-20b model card.
\newblock \emph{gpt-oss model card}, 1:\penalty0 1, 2025.
\newblock URL \url{https://arxiv.org/abs/2508.10925}.

\bibitem[Patil et~al.(2025)Patil, Mao, Yan, Ji, Suresh, Stoica, and
  Gonzalez]{patilberkeley}
Shishir~G Patil, Huanzhi Mao, Fanjia Yan, Charlie Cheng-Jie Ji, Vishnu Suresh,
  Ion Stoica, and Joseph~E Gonzalez.
\newblock The berkeley function calling leaderboard (bfcl): From tool use to
  agentic evaluation of large language models.
\newblock In \emph{Forty-second International Conference on Machine Learning},
  2025.
\newblock URL \url{https://openreview.net/pdf?id=2GmDdhBdDk}.

\bibitem[Rawles et~al.(2025)Rawles, Clinckemaillie, Chang, Waltz, Lau, Fair,
  Li, Bishop, Li, Campbell-Ajala, Toyama, Berry, Tyamagundlu, Lillicrap, and
  Riva]{rawlesandroidworld}
Christopher Rawles, Sarah Clinckemaillie, Yifan Chang, Jonathan Waltz,
  Gabrielle Lau, Marybeth Fair, Alice Li, William~E Bishop, Wei Li, Folawiyo
  Campbell-Ajala, Daniel~Kenji Toyama, Robert~James Berry, Divya Tyamagundlu,
  Timothy~P Lillicrap, and Oriana Riva.
\newblock Androidworld: A dynamic benchmarking environment for autonomous
  agents.
\newblock In \emph{The Thirteenth International Conference on Learning
  Representations}, 2025.
\newblock URL \url{https://openreview.net/forum?id=il5yUQsrjC}.

\bibitem[Shinn et~al.(2023)Shinn, Cassano, Gopinath, Narasimhan, and
  Yao]{shinn2023reflexion}
Noah Shinn, Federico Cassano, Ashwin Gopinath, Karthik Narasimhan, and Shunyu
  Yao.
\newblock Reflexion: Language agents with verbal reinforcement learning.
\newblock \emph{Advances in Neural Information Processing Systems},
  36:\penalty0 8634--8652, 2023.
\newblock URL \url{https://openreview.net/pdf?id=vAElhFcKW6}.

\bibitem[Shridhar et~al.(2021)Shridhar, Yuan, Cote, Bisk, Trischler, and
  Hausknecht]{shridharalfworld}
Mohit Shridhar, Xingdi Yuan, Marc-Alexandre Cote, Yonatan Bisk, Adam Trischler,
  and Matthew Hausknecht.
\newblock Alfworld: Aligning text and embodied environments for interactive
  learning.
\newblock In \emph{International Conference on Learning Representations}, 2021.
\newblock URL \url{https://openreview.net/pdf?id=0IOX0YcCdTn}.

\bibitem[Sun et~al.(2023)Sun, Yin, Li, Wu, Qiu, and Kong]{sun2024pushing}
Qiushi Sun, Zhangyue Yin, Xiang Li, Zhiyong Wu, Xipeng Qiu, and Lingpeng Kong.
\newblock Corex: Pushing the boundaries of complex reasoning through
  multi-model collaboration.
\newblock \emph{arXiv preprint arXiv:2310.00280}, 2023.

\bibitem[Sun et~al.(2025{\natexlab{a}})Sun, Cheng, Ding, Jin, Wang, Xu, Wu,
  Jia, Chen, Liu, et~al.]{sun2025genesis}
Qiushi Sun, Kanzhi Cheng, Zichen Ding, Chuanyang Jin, Yian Wang, Fangzhi Xu,
  Zhenyu Wu, Chengyou Jia, Liheng Chen, Zhoumianze Liu, et~al.
\newblock Os-genesis: Automating gui agent trajectory construction via reverse
  task synthesis.
\newblock In \emph{Proceedings of the 63rd Annual Meeting of the Association
  for Computational Linguistics (Volume 1: Long Papers)}, pages 5555--5579,
  2025{\natexlab{a}}.

\bibitem[Sun et~al.(2025{\natexlab{b}})Sun, Liu, Ma, Ding, Xu, Yin, Zhao, Wu,
  Cheng, Liu, et~al.]{sun2025scienceboard}
Qiushi Sun, Zhoumianze Liu, Chang Ma, Zichen Ding, Fangzhi Xu, Zhangyue Yin,
  Haiteng Zhao, Zhenyu Wu, Kanzhi Cheng, Zhaoyang Liu, et~al.
\newblock Scienceboard: Evaluating multimodal autonomous agents in realistic
  scientific workflows.
\newblock \emph{arXiv preprint arXiv:2505.19897}, 2025{\natexlab{b}}.

\bibitem[Tang et~al.(2024)Tang, Li, Liang, Zhu, Zhang, and Zheng]{tang2024mars}
Xiaojuan Tang, Jiaqi Li, Yitao Liang, Song-Chun Zhu, Muhan Zhang, and Zilong
  Zheng.
\newblock Mars: Situated inductive reasoning in an open-world environment.
\newblock In \emph{The Thirty-eight Conference on Neural Information Processing
  Systems Datasets and Benchmarks Track}, 2024.
\newblock URL \url{https://openreview.net/forum?id=3qoQ6AolAz}.

\bibitem[Vodrahalli et~al.(2024)Vodrahalli, Ontanon, Tripuraneni, Xu, Jain,
  Shivanna, Hui, Dikkala, Kazemi, Fatemi,
  et~al.]{vodrahalli2024michelangelolongcontextevaluations}
Kiran Vodrahalli, Santiago Ontanon, Nilesh Tripuraneni, Kelvin Xu, Sanil Jain,
  Rakesh Shivanna, Jeffrey Hui, Nishanth Dikkala, Mehran Kazemi, Bahare Fatemi,
  et~al.
\newblock Michelangelo: Long context evaluations beyond haystacks via latent
  structure queries.
\newblock \emph{CoRR}, 2024.
\newblock URL \url{https://openreview.net/forum?id=jdc57bqY3u}.

\bibitem[Wang et~al.(2025{\natexlab{a}})Wang, Shi, Hu, Ma, Liu, Wang, Yao, Liu,
  Ge, and Zhang]{wang2024largelanguagemodelsrobotics}
Jiaqi Wang, Enze Shi, Huawen Hu, Chong Ma, Yiheng Liu, Xuhui Wang, Yincheng
  Yao, Xuan Liu, Bao Ge, and Shu Zhang.
\newblock Large language models for robotics: Opportunities, challenges, and
  perspectives.
\newblock \emph{Journal of Automation and Intelligence}, 4\penalty0
  (1):\penalty0 52--64, 2025{\natexlab{a}}.
\newblock URL
  \url{https://www.sciencedirect.com/science/article/pii/S2949855424000613}.

\bibitem[Wang et~al.(2025{\natexlab{b}})Wang, Han, Diaz, Xu, R{\"u}hle, and
  Rajmohan]{wang2025odysseybench}
Weixuan Wang, Dongge Han, Daniel~Madrigal Diaz, Jin Xu, Victor R{\"u}hle, and
  Saravan Rajmohan.
\newblock Odysseybench: Evaluating llm agents on long-horizon complex office
  application workflows.
\newblock \emph{arXiv preprint arXiv:2508.09124}, 2025{\natexlab{b}}.
\newblock URL \url{https://arxiv.org/abs/2508.09124}.

\bibitem[Wei et~al.(2025)Wei, Sun, Papay, McKinney, Han, Fulford, Chung,
  Passos, Fedus, and Glaese]{wei2025browsecomp}
Jason Wei, Zhiqing Sun, Spencer Papay, Scott McKinney, Jeffrey Han, Isa
  Fulford, Hyung~Won Chung, Alex~Tachard Passos, William Fedus, and Amelia
  Glaese.
\newblock Browsecomp: A simple yet challenging benchmark for browsing agents.
\newblock \emph{arXiv preprint arXiv:2504.12516}, 2025.
\newblock URL \url{https://arxiv.org/pdf/2504.12516}.

\bibitem[Xie et~al.(2024)Xie, Zhang, Chen, Li, Zhao, Cao, Hua, Cheng, Shin,
  Lei, et~al.]{xie2024osworld}
Tianbao Xie, Danyang Zhang, Jixuan Chen, Xiaochuan Li, Siheng Zhao, Ruisheng
  Cao, Toh~J Hua, Zhoujun Cheng, Dongchan Shin, Fangyu Lei, et~al.
\newblock Osworld: Benchmarking multimodal agents for open-ended tasks in real
  computer environments.
\newblock \emph{Advances in Neural Information Processing Systems},
  37:\penalty0 52040--52094, 2024.
\newblock URL \url{https://openreview.net/forum?id=tN61DTr4Ed#discussion}.

\bibitem[Xu et~al.(2025{\natexlab{a}})Xu, Yan, Ma, Zhao, Liu, Lin, and
  Wu]{xu2025ph}
Fangzhi Xu, Hang Yan, Chang Ma, Haiteng Zhao, Jun Liu, Qika Lin, and Zhiyong
  Wu.
\newblock $\phi$-decoding: Adaptive foresight sampling for balanced
  inference-time exploration and exploitation.
\newblock In \emph{Proceedings of the 63rd Annual Meeting of the Association
  for Computational Linguistics (Volume 1: Long Papers)}, pages 13214--13227,
  Vienna, Austria, July 2025{\natexlab{a}}. Association for Computational
  Linguistics.
\newblock ISBN 979-8-89176-251-0.
\newblock \doi{10.18653/v1/2025.acl-long.647}.
\newblock URL \url{https://aclanthology.org/2025.acl-long.647/}.

\bibitem[Xu et~al.(2025{\natexlab{b}})Xu, Yan, Ma, Zhao, Sun, Cheng, He, Liu,
  and Wu]{xu2025genius}
Fangzhi Xu, Hang Yan, Chang Ma, Haiteng Zhao, Qiushi Sun, Kanzhi Cheng, Junxian
  He, Jun Liu, and Zhiyong Wu.
\newblock Genius: A generalizable and purely unsupervised self-training
  framework for advanced reasoning.
\newblock In \emph{Proceedings of the 63rd Annual Meeting of the Association
  for Computational Linguistics (Volume 1: Long Papers)}, pages 13153--13167,
  Vienna, Austria, July 2025{\natexlab{b}}. Association for Computational
  Linguistics.
\newblock ISBN 979-8-89176-251-0.
\newblock \doi{10.18653/v1/2025.acl-long.644}.
\newblock URL \url{https://aclanthology.org/2025.acl-long.644/}.

\bibitem[Xu et~al.(2025{\natexlab{c}})Xu, Song, Li, Tang, Jain, Bao, Wang,
  Zhou, Guo, Cao, Yang, Lu, Martin, Su, Maben, Mehta, Chi, Jang, Xie, Zhou, and
  Neubig]{xu2025theagentcompany}
Frank~F. Xu, Yufan Song, Boxuan Li, Yuxuan Tang, Kritanjali Jain, Mengxue Bao,
  Zora~Zhiruo Wang, Xuhui Zhou, Zhitong Guo, Murong Cao, Mingyang Yang,
  Hao~Yang Lu, Amaad Martin, Zhe Su, Leander~Melroy Maben, Raj Mehta, Wayne
  Chi, Lawrence~Keunho Jang, Yiqing Xie, Shuyan Zhou, and Graham Neubig.
\newblock Theagentcompany: Benchmarking {LLM} agents on consequential real
  world tasks.
\newblock In \emph{The Thirty-ninth Annual Conference on Neural Information
  Processing Systems Datasets and Benchmarks Track}, 2025{\natexlab{c}}.
\newblock URL \url{https://openreview.net/forum?id=LZnKNApvhG}.

\bibitem[Yan et~al.(2025)Yan, Che, Xu, Sun, Ding, Cheng, Zhang, Qin, Liu, and
  Lin]{yan2025tide}
Hang Yan, Xinyu Che, Fangzhi Xu, Qiushi Sun, Zichen Ding, Kanzhi Cheng, Jian
  Zhang, Tao Qin, Jun Liu, and Qika Lin.
\newblock Tide: Trajectory-based diagnostic evaluation of test-time improvement
  in llm agents.
\newblock \emph{arXiv preprint arXiv:2602.02196}, 2025.
\newblock URL \url{https://arxiv.org/abs/2602.02196}.

\bibitem[Yang et~al.(2025)Yang, Li, Yang, Zhang, Hui, Zheng, Yu, Gao, Huang,
  Lv, et~al.]{yang2025qwen3}
An~Yang, Anfeng Li, Baosong Yang, Beichen Zhang, Binyuan Hui, Bo~Zheng, Bowen
  Yu, Chang Gao, Chengen Huang, Chenxu Lv, et~al.
\newblock Qwen3 technical report.
\newblock \emph{arXiv preprint arXiv:2505.09388}, 2025.
\newblock URL \url{https://arxiv.org/abs/2505.09388}.

\bibitem[Yao et~al.(2023)Yao, Zhao, Yu, Du, Shafran, Narasimhan, and
  Cao]{yao2022react}
Shunyu Yao, Jeffrey Zhao, Dian Yu, Nan Du, Izhak Shafran, Karthik~R Narasimhan,
  and Yuan Cao.
\newblock React: Synergizing reasoning and acting in language models.
\newblock In \emph{The eleventh international conference on learning
  representations}, 2023.
\newblock URL \url{https://openreview.net/forum?id=WE_vluYUL-X}.

\bibitem[Yao et~al.(2025)Yao, Shinn, Razavi, and Narasimhan]{yao2025taubench}
Shunyu Yao, Noah Shinn, Pedram Razavi, and Karthik~R Narasimhan.
\newblock \{\${\textbackslash}tau\$\}-bench: A benchmark for
  {\textbackslash}underline\{T\}ool-{\textbackslash}underline\{A\}gent-{\textbackslash}underline\{U\}ser
  interaction in real-world domains.
\newblock In \emph{The Thirteenth International Conference on Learning
  Representations}, 2025.
\newblock URL \url{https://openreview.net/forum?id=roNSXZpUDN}.

\bibitem[Zhou et~al.(2024)Zhou, Xu, Zhu, Zhou, Lo, Sridhar, Cheng, Ou, Bisk,
  Fried, et~al.]{zhouwebarena}
Shuyan Zhou, Frank~F Xu, Hao Zhu, Xuhui Zhou, Robert Lo, Abishek Sridhar,
  Xianyi Cheng, Tianyue Ou, Yonatan Bisk, Daniel Fried, et~al.
\newblock Webarena: A realistic web environment for building autonomous agents.
\newblock In \emph{The Twelfth International Conference on Learning
  Representations}, 2024.
\newblock URL \url{https://openreview.net/forum?id=oKn9c6ytLx}.

\end{thebibliography}
